\documentclass[11pt,fleqn,twoside]{article}

\newif\ifdraft\drafttrue
\newif\ifinlineref\inlinereffalse
\newif\iffinal\finalfalse
\newif\ifdotikz\dotikzfalse
\newif\iftechreport\techreportfalse
\newif\ifmakeallproofsinline\makeallproofsinlinefalse
\newif\ifshowotherappendix\showotherappendixtrue

\inlinereftrue
\draftfalse
\dotikztrue
\techreporttrue
\usepackage[german,english]{babel}
\usepackage{lcreport}
\usepackage[T1]{fontenc}

\foreignreport

\usepackage[table]{xcolor}

\usepackage{mycite}

\newcommand{\nop}[1]{}

\usepackage{times}
\usepackage[scaled]{helvet}
\usepackage{courier}

\usepackage{latexsym}
\usepackage{amsmath,wasysym}
\usepackage{amssymb}
\usepackage{harpoon}
\usepackage{multirow}
\usepackage{paralist}
\usepackage{calc}
\usepackage{booktabs}

\usepackage[vlined,nofillcomment,algoruled]{algorithm2e}
\usepackage[outsourcebydefault]{mytheorem}

\newMyTheoremType{theorem}{Theorem}
\newMyTheoremType{claim}{Claim}
\newMyTheoremType{corollary}{Corollary}
\newMyTheoremType{proposition}{Proposition}
\newMyTheoremType{lemma}{Lemma}

\SetAlFnt{\sffamily\tiny}

\usepackage{url}
\makeatletter
\let\UrlSpecialsOld\UrlSpecials
\def\UrlSpecials{\UrlSpecialsOld\do\/{\Url@slash}\do\_{\Url@underscore}}%
\def\Url@slash{\@ifnextchar/{\kern-.11em\mathchar47\kern-.2em}%
    {\kern-.0em\mathchar47\kern-.08em\penalty\UrlBigBreakPenalty}}
\def\Url@underscore{\nfss@text{\leavevmode \kern.06em\vbox{\hrule\@width.3em}}}
\makeatother

\usepackage{tikz}
\pgfrealjobname{inconsistency}
\usetikzlibrary{shapes,arrows,backgrounds,chains,%
matrix,patterns,arrows,decorations.pathmorphing,decorations.pathreplacing,%
positioning,fit,calc,decorations.text,shadows,plotmarks%
}

\newcommand{\leanparagraph}[1]{\smallskip\noindent\textbf{#1}}

\newcounter{myenumeratecounter}

\renewcommand{\vec}[1]{\ensuremath{\mb{#1}}}

\newcommand{\mi}[1]{\ensuremath{\mathit{#1}}}
\newcommand{\mb}[1]{\ensuremath{\mathbf{#1}}}
\def\lif{\ensuremath{\leftarrow}}

\def\naf{\ensuremath{\mathop{not}}}

\def\cI{\ensuremath{{\mathcal{I}}}}
\def\cA{\ensuremath{{\mathcal{A}}}}
\def\cE{\ensuremath{{\mathcal{E}}}}

\def\scI{\ensuremath{{\textsc{i}}}}
\def\scO{\ensuremath{{\textsc{o}}}}

\newcommand\bi{\begin{itemize}}
\newcommand\ei{\end{itemize}}
\newcommand\quo[1]{`#1'}

\newcommand{\eqs}{\,{=}\,}
\newcommand{\neqs}{\,{\neq}\,}

\newcommand{\cups}{\,{\cup}\,}

\newcommand{\lors}{\,{\vee}\,}

\newcommand\hex{{\sc hex}\xspace}
\newcommand\dlv{{\small\sffamily dlv}\xspace}
\newcommand\dlvhex{{\small\sffamily dlvhex}\xspace}

\newcommand{\AS}{\mathcal{A\!S}}
\newcommand{\lifs}{\,{\lif}\,}

\newcommand\heurold{\textsl{H1}\xspace}
\newcommand\heurnew{\textsl{H2}\xspace}

\newcommand{\mycuri}[0]{\ensuremath{\mi{cur\scI}}}
\newcommand{\mycuro}[0]{\ensuremath{\mi{cur\scO}}}
\newcommand{\myrefcounto}[0]{\ensuremath{\mi{refs\scO}}}

\newcommand\clasp{{\small\sffamily clasp}\xspace}

\newcommand\gringo{{\small\sffamily gringo}\xspace}
\newcommand\clingo{{\small\sffamily clingo}\xspace}

\newcommand{\amp}[1]{\ensuremath{\text{\textsl{{\&}}}\!\mathit{#1}}}
\newcommand{\ext}[3]{\ensuremath{\amp{#1}[#2](#3)}}
\newcommand{\extfun}[1]{\ensuremath{f_{\text{\sl\&}#1}}}
\newcommand{\extFun}[1]{\ensuremath{F_{\text{\sl\&}#1}}}
\newcommand{\extsem}[4]{\ensuremath{f_{\text{\sl\&}#1}(#2,#3,#4)}}

\newcommand{\rat}[1]{\ensuremath{d_{\text{\it\&}#1}}}

\newcommand{\p}{\ensuremath{\mathsf{P}}\xspace}

\newcommand{\np}{\ensuremath{\mathsf{NP}}\xspace}
\newcommand{\conp}{co-\ensuremath{\mathsf{NP}}\xspace}

\newcommand{\GroundLiberallyDomainExpansionSafeProgram}{\ensuremath{\textsc{GroundHEX}}}
\newcommand{\EvaluateGroundHEX}{\ensuremath{\textsc{EvaluateGroundHEX}}}
\newcommand{\EvaluateLDESafe}{\ensuremath{\textsc{EvaluateLDESafe}}}

\newcommand{\grnd}{\ensuremath{\mathit{grnd}}}

\newcommand{\ipar}{\textsc{i}}
\newcommand{\opar}{\textsc{o}}

\def\dependsext{\rightarrow^e}

\def\dependsmon{\rightarrow_m}
\def\dependsnmon{\rightarrow_n}

\newcommand{\T}{\mathbf{T}}
\newcommand{\F}{\mathbf{F}}

\newcommand{\Assignment}{\ensuremath{\mathbf{A}}}
\newcommand{\AssignmentP}{\ensuremath{\hat{\mathbf{A}}}}
\newcommand{\Program}{\ensuremath{P}}
 
\newcommand{\decisionlevel}{\ensuremath{\mathit{dl}}}

\newcommand{\ProgramP}{\ensuremath{\hat{P}}}

\newcommand{\toFacts}[1]{\mathit{facts}(#1)}

\newcommand{\Propagation}{\textsf{Propagation}}

\newcommand{\CDNLHEX}{\textsf{\hex-CDNL-PA}}

\def\join{{\:\bowtie\:}}

\def\ufinal{u_{\mi{final}}}

\def\myiint{i-inter\-pre\-ta\-tion\xspace}

\def\myoint{o-inter\-pre\-ta\-tion\xspace}

\def\myimodels{\mi{i\text{-}ints}}
\def\myomodels{\mi{o\text{-}ints}}

\def\myunit{\mi{unit}}
\def\mytype{\mi{type}}
\def\myint{\mi{int}}
\newcommand{\myinputsE}[1]{\mi{preds}_{#1}}
\newcommand{\myinputs}{\myinputsE{\cE}}

\def\myevaluatePregroundable{\ensuremath{\textsc{evaluate}\-\textsc{Pre}\-\textsc{Groundable}}\xspace}

\def\myBuildAnswerSets{\ensuremath{\textsc{Build}\-\textsc{Answer}\-\textsc{Sets}}\xspace}
\def\myGetNextUnitModel{\ensuremath{\textsc{Get}\-\textsc{Next}\-\textsc{Unit}\-\textsc{Model}}\xspace}
\def\myGetNextIModel{\ensuremath{\textsc{Get}\-\textsc{Next}\-\textsc{Input}\-\textsc{Model}}\xspace}
\def\myGetNextOModel{\ensuremath{\textsc{Get}\-\textsc{Next}\-\textsc{Output}\-\textsc{Model}}\xspace}
\def\myEnsureModelIncrement{\ensuremath{\textsc{Ensure}\-\textsc{Model}\-\textsc{Increment}}\xspace}
\def\myOnDemandAS{\ensuremath{\textsc{Answer}\-\textsc{Sets}\-\textsc{On}\-\textsc{Demand}}\xspace}
\def\myNextAnswerSet{\ensuremath{\mathit{NextAnswerSet}}}
\def\undef{\ensuremath{\textsc{undef}}}

\def\myCAUtext{FAI\xspace}
\def\myCAUstext{FAIs\xspace}
\def\mycau{\mi{fai}}

\newcommand\facts[0]{\ensuremath{\mi{facts}}}

\usepackage{ntheorem}

\newtheorem{theorem}{Theorem}

\newtheorem{definition}{Definition}
\newtheorem{example}{Example}

{\qed
\medskip}

\newcommand{\qed}[0]{\ensuremath{\Box}}

\newcommand{\comment}[1]{}

\newcommand{\Lightning}{\lightning}

\ifmakeallproofsinline
\newcommand{\myinlineproof}[1]{#1}
\newcommand{\mylocatedproof}[1]{}
\else
\newcommand{\myinlineproof}[1]{}
\newcommand{\mylocatedproof}[1]{#1}
\fi

\input{figuremacros.tex}

\infsysrr{18-04}

\authornames{
{\sc Christoph Redl}
}

\title{Inconsistency in Answer Set Programs and Extensions}

\author{
Christoph Redl\affiliation{
Institut f\"ur Logic and Computation, Technische Universit\"at Wien,
Favoritenstra\ss{}e\ 9-11, A-1040 Vienna, Austria;
email: redl@kr.tuwien.ac.at.}
}

\abstract{
		Answer Set Programming (ASP) is a well-known problem solving approach based on nonmonotonic logic programs.
		\hex-programs extend ASP with \emph{external atoms} for accessing arbitrary external information, which
		can introduce values that do not appear in the input program.
		In this work we consider \emph{inconsistent} ASP- and \hex-programs, i.e., programs without answer sets.
		We study characterizations of inconsistency, introduce a novel notion for explaining inconsistencies in terms of input facts,
		analyze the complexity of reasoning tasks in context of inconsistency analysis, and present techniques for computing inconsistency reasons.
		This theoretical work is motivated by two concrete applications, which we also present.
		The first one is the new modeling technique of \emph{query answering over subprograms}
		as a convenient alternative to the well-known \emph{saturation technique}.
		The second application is a new evaluation algorithm for \hex-programs
		based on \emph{conflict-driven learning for programs with multiple components}:
		while for certain program classes previous techniques suffer an evaluation bottleneck,
		the new approach shows significant, potentially exponential speedup in our experiments.
		Since well-known ASP extensions such as constraint ASP and DL-programs correspond to special cases of \hex,
		all presented results are interesting beyond the specific formalism.
}

\acknowledgement{
	This research has been supported by the Austrian Science
    Fund (FWF) project P27730.
}

\begin{document}

\maketitle
\tableofcontents
\newpage

	\section{Introduction}
	\label{sec:introduction}

		Answer-Set Programming (ASP) is a declarative programming paradigm
		based on nonmonotonic programs and a multi-model semantics~\cite{gelf-lifs-91}.
		The problem at hand is encoded as an ASP-program, which consists of rules,
		such that its models, called \emph{answer sets}, correspond to the solutions to the original problem.
		\hex{}-programs are an extension of ASP with external sources
		such as description logic ontologies and Web resources~\cite{DBLP:conf/birthday/EiterRS16}.
		So-called external atoms pass information from the logic program,
		given by predicate extensions and constants, to an external source, which in turn
		returns values to the program.
		Notably, \emph{value invention} allows for
		domain expansion, i.e., external sources might return values which do not appear
		in the program.
		For instance, the external atom $\ext{\mathit{synonym}}{\mathit{car}}{X}$
		might be used to find the synonyms $X$ of $\mathit{car}$, e.g.~$\mathit{automobile}$.
		Also recursive data exchange between the program and external sources
		is supported, which leads to high expressiveness of the formalism.

		Inconsistent programs are programs which do not possess any answer sets.
		In this paper we study inconsistent programs:
		we present a technique for deciding inconsistency of programs by other logic programs,
		introduce a novel notion for explaining inconsistencies in terms of input facts,
		analyze the complexity of reasoning tasks in this context, and present techniques for computing inconsistency explanations.
		The work is motivated by two concrete applications.
		
		The first one is
		\textbf{meta-reasoning} about the answer sets of a \emph{(sub-)program} within another \emph{(meta-)program},
		such as aggregation of results.
		Traditionally, such meta-reasoning requires postprocessing, i.e., the answer sets are inspected after the reasoner terminates.
		Some simple reasoning tasks, such as brave or cautious query answering, are directly supported by some systems.
		However, even then the answer to a brave or cautious query is not represented \emph{within} the program but appears only as output on the command-line,
		which prohibits the direct continuation of reasoning based on the query answer.
		An existing approach, which allows for meta-reasoning within a program over the answer sets of another program, are \emph{manifold programs}.
		They compile the calling and the called program into a single one~\cite{DBLP:conf/birthday/FaberW11}.
		The answer sets of the called program are then represented within each answer set of the calling program.
		However, this approach uses weak constraints, which are not supported by all systems.
		Moreover, the encoding requires a separate copy of the subprogram for each atom occurring in it, which appears to be impractical.
		Another approach are \emph{nested \hex-programs}. Here, dedicated atoms access answer sets of a subprograms and their literals explicitly as accessible objects~\cite{ekr2013-inap11}.
		However, this approach is based on \hex-programs -- an extension of ASP -- and not applicable if an ordinary ASP solver is used.
		Our work on inconsistency can be exploited to realize \textbf{query answering over subprograms}.
		We propose this possibility as alternative modeling technique to \emph{saturation} (cf.~e.g.~\citeN{eik2009-rw}), which
		exploits the minimality of answer sets for solving \conp-hard problems within disjunctive ASP, such as checking if a property holds for \emph{all} objects in a domain.
		However, the technique is advanced and not easily applicable by average ASP users
		as using default-negation for checking properties within saturation encodings is restricted.

		Later, we extend the study on inconsistency by
		introducing the \textbf{concept of inconsistency reasons (IRs)} for
		identifying \emph{classes of inconsistent program instances} in terms of
		\emph{input facts}.
		The concept is driven by the typical usage of ASP,
		where the proper rules (IDB; intensional database) are fixed and encode the general problem,
		while the current instance is specified as facts (EDB; extensional database).
		It is then interesting to identify sets of instances which lead to inconsistency.
		The extension is motivated by another application: \textbf{optimizing the evaluation algorithm for \hex-programs}.
		Grounding a \hex-program is expensive for certain program classes since in general,
		already the generation of a single ground instance of a rule requires
		external sources to be evaluated under up to exponentially many inputs to ensure that all relevant constants are respected~\cite{efkr2016-aij}.
		The situation is relieved by
		a model-building framework based on \emph{program splitting},
		where program components are arranged in a directed acyclic graph~\cite{efikrs2016-tplp}.
		Then, at the time a component is grounded, its predecessors
		have already been evaluated and their answer sets can be exploited to skip evaluations.
		However, splitting deteriorates the performance of the conflict-driven solver
		since splits act as barriers for propagation.
		Therefore, for certain program classes, current approaches suffer a bottleneck, which is either due to expensive grounding or due to splitting the guessing part from the checking part.
		To overcome this bottleneck we propose a novel learning technique for programs with multiple components, which is based on inconsistency reasons.

		Previous related work was in context of answer set program debugging and
		focused on explaining why a \emph{particular interpretation} fails to be an answer set,
		while we aim at explaining why the overall program is inconsistent.
		Moreover, debugging approaches also focus on explanations
		which support the \emph{human user} to find errors in the program.
		Such reasons can be, for instance, in terms of minimal sets of constraints which need to be removed in order to regain consistency,
		in terms of odd loops (i.e., cycles of mutually depending atoms which involve negation and are of odd length), or in terms of unsupported atoms, see e.g.~\citeN{lpnmr07a}
		and~\citeN{TUW-167810} for some approaches.
		To this end, one usually assumes that a single fixed program is given whose undesired behavior needs to be explained.
		In contrast, we consider a program whose input facts are subject to change
		and identify classes of instances which lead to inconsistency.

		The organization and contributions of this paper are as follows:
		\begin{itemize}
			\item In Section~\ref{sec:prelim} we present the necessary preliminaries on ASP- and \hex-programs.
			\item In Section~\ref{sec:embedding} we first discuss restrictions of the saturation technique and point out that using default-negation within
				saturation encodings would be convenient but is not easily possible.
				We then show how inconsistency of a normal logic program can be decided within another (disjunctive) program.
				To this end, we present a saturation encoding which simulates the computation of answer sets of the subprogram and represents the existence of
				an answer set by a single atom of the meta-program.
			\item In Section~\ref{sec:queryanswering} we present our first application, which is meta-reasoning over the answer sets of a subprogram.
				To this end, we first realize brave and cautious query answering over a subprogram,
				which can be used as black box such that the user does not need to have deep knowledge about the underlying ideas.
				Checking a \conp-complete property can then be expressed naturally by a cautious query.
			\item In Section~\ref{sec:inconsistencyanalysis} we define the novel notion of \emph{inconsistency reasons (IRs)} for \hex-programs
				wrt.~a set of input facts.
				In contrast to Section~\ref{sec:embedding}, which focuses on deciding inconsistency of a single fixed program, we identify classes of program instances
				which are inconsistent.
				We then analyze the complexity of reasoning tasks related to the computation of inconsistency reasons.
			\item In Section~\ref{sec:computing} we present a meta-programming technique as well as a procedural algorithms for computing IRs
				for various classes of programs.
			\item In Section~\ref{sec:hexprogramevaluation} we present our second application, which is a novel evaluation algorithm for \hex-programs based on IRs.
				To this end, we develop a technique for
				conflict-driven program solving in presence of multiple program components. We implement this approach in our prototype system
				and perform an experimental analysis, which shows a significant (potentially exponential) speedup.
			\item In Section~\ref{sec:related} we discuss related work and differences to our approach in more detail.
			\item In Section~\ref{sec:conclusion} we conclude the paper and give an outlook on future work.
			\item Proofs are outsourced to Appendix~\ref{sec:proofs}.
		\end{itemize}
		
		A preliminary version of this work has been presented at LPNMR~2017 and IJCAI~2017~\cite{r2017a-lpnmr,r2017b-lpnmr,r2017-ijcai}.
		The extensions in this paper comprise of more extensive formalizations and discussions of the theoretical contributions,
		additional complexity results, and formal proofs of the results.

	\section{Preliminaries}
	\label{sec:prelim}

		In this section we recapitulate the syntax and semantics of \hex-programs, which
		generalize (disjunctive) logic programs under the answer set
		semantics~\cite{gelf-lifs-91}; for an exhaustive discussion of the background and possible applications we refer to~\citeN{efikrs2016-tplp}.

		We use as our alphabet the sets
		$\mathcal{P}$ of predicates, $\mathcal{F}$ of function symbols, $\mathcal{X}$ of external predicates, $\mathcal{C}$ of constants, and $\mathcal{V}$ of variables.
		We assume that the sets of predicates $\mathcal{P}$ and function symbols $\mathcal{F}$
		can share symbols, while the sets are mutually disjoint otherwise.
		This is by intend and will be needed for our encoding in Section~\ref{sec:embedding}.
		However, it is clear from its position if a symbol is currently used as a predicate or a function symbol.
		We further let the set of terms $\mathcal{T}$ be the least set such that
		$\mathcal{C} \subseteq \mathcal{T}$,
		$\mathcal{V} \subseteq \mathcal{T}$,
		and whenever $f \in \mathcal{F}$ and $T_1, \ldots, T_\ell \in \mathcal{T}$ then $f(T_1, \ldots, T_\ell) \in \mathcal{T}$.

		Each predicate $p \in \mathcal{P}$ has a fixed arity $\mathit{ar}(p) \in \mathbb{N}$.
		An (ordinary) atom $a$ is of form $p(t_1, \dotsc, t_\ell)$,
		where $p \in \mathcal{P}$ is a predicate symbol with arity $\mathit{ar}(p) = \ell$ and
		$t_1, \dotsc, t_\ell \in \mathcal{T}$ are terms,
		abbreviated as $p(\vec{t})$;
		for $\mathit{ar}(p) = 0$ we drop parentheses and write $p()$ simply as $p$.
		For a list of terms $\vec{t} = t_1, \ldots, t_\ell$
		we write $t \in \vec{t}$ if $t = t_i$ for some $1 \le i \le \ell$; akin for lists of other objects.
		We call an atom \emph{ground} if it does not contain variables; similarly for other objects.
		A (signed) literal is of type $\T a$ or $\F a$, where $a$ is an atom.
		We let $\overline{\sigma}$ denote the negation of a
		literal $\sigma$, i.e.\ $\overline{\T a} = \F a$ and $\overline{\F a} = \T a$.
		
		An {\em assignment} $\Assignment$ over a (finite) set $A$ of ground atoms 
		is a set of literals over $A$,
		where $\T a \in \Assignment$ expresses that~$a$ is true, also denoted $\Assignment \models a$, and $\F a \in \Assignment$ that $a$ is false, also denoted $\Assignment \not\models a$.
		Assignments $\Assignment$ are called \emph{complete} wrt.~$A$, if for every $a \in A$ either $\T a$ or $\F a$ is contained in $\Assignment$,
		and they are called \emph{partial} otherwise; partial assignments allow for distinguishing false from unassigned atoms, as needed while the reasoner is traversing the search space.
		Complete assignments $\Assignment$ are also called \emph{interpretations} and, for simplicity, they are also denoted as the set $I = \{ a \mid \T a \in \Assignment \}$ of true atoms,
		while all other atoms are implicitly false (there are no unassigned ones).
		In Sections~\ref{sec:embedding}--\ref{sec:inconsistencyanalysis}
		we will make use of the simplified notation as only complete assignments are relevant,
		while beginning from Section~\ref{sec:computing:ground} we need partial assignments and use the notation as set of signed literals;
		we will notify the reader again when we switch the notation.

		\leanparagraph{\hex-Program Syntax}
		\hex-programs extend ordinary ASP-programs by \emph{external atoms},
		which enable a bidirectional interaction between a program
		and external sources of computation.
		An \emph{external atom} is of the form 
		$\ext{g}{\vec{Y}}{\vec{X}}$,
		where $\amp{g} \in \mathcal{X}$ is an external predicate,
		$\vec{Y} = p_1, \dotsc, p_k$ is a list of input parameters
		(predicate parameters from $\mathcal{P}$ or constant parameters\footnote{Also variables become constants after the grounding step.}
		from $\mathcal{C} \cup \mathcal{V}$), called \emph{input list},
		and $\vec{X} = t_1, \dotsc, t_l$ are output terms.

		\begin{definition}
			\label{def:rule}
			A \hex-program $\Program$ is a set of rules
			\begin{equation*}
				\label{eq:rule}
			  a_1\lor\cdots\lor a_k \leftarrow b_1,\dotsc, b_m, \naf\, b_{m+1}, \dotsc, \naf\, b_n \ ,
			\end{equation*}
			where each $a_i$ is an ordinary atom and each~$b_j$
			is either an ordinary atom or an external atom.
		\end{definition}
	
		\noindent The \emph{head} of a rule $r$ is $H(r) = \{ a_1, \ldots, a_k \}$,
		its \emph{body} is $B(r) = \{b_1, \dotsc, b_m,\naf\, b_{m+1}, \dotsc, \naf\, b_n\}$,
		and its \emph{positive resp.~negative body} is $B^{+}(r) = \{ b_1, \ldots, b_m \}$ resp.~$B^{-}(r) = \{ b_{m+1}, \ldots, b_n \}$.
		We let $B^{+}_o(r)$ resp.~$B^{-}_o(r)$ be the set of ordinary atoms in $B^{+}(r)$ resp.~$B^{-}(r)$.
		For a program $\Program$ we let $X(P) = \bigcup_{r \in P} X(r)$ for $X \in \{ H, B, B^{+}, B^{-} \}$.

		In the following, we call a program \emph{ordinary} if it does not contain external atoms,
		i.e., if it is a standard ASP-program.
		Moreover, a rule as by Definition~\ref{def:rule} is called \emph{normal} if $k = 1$
		and a program is called \emph{normal} if all its rules are normal.
		A rule $\leftarrow b_1,\dotsc, b_m, \naf\, b_{m+1}, \dotsc, \naf\, b_n$ (i.e., with $k = 0$)
		is called a \emph{constraint} and can be seen as normal rule $f \leftarrow b_1,\dotsc, b_m, \naf\, b_{m+1}, \dotsc, \naf\, b_n, \naf f$
		where $f$ is a new atom which does not appear elsewhere in the program.

		We will further use \emph{builtin atoms} in rule bodies,
		which are of kind $o_1 \circ o_2$, where $o_1$ and $o_2$ are constants and $\circ$ is an arithmetic operator (e.g., $<$, $\le$, $!=$, etc.).
		More language features (in particular conditional literals) are introduced on-the-fly as needed.
		
		\leanparagraph{\hex-Program Semantics}
		We first discuss the semantics of ground programs $\Program$.
		In the following, assignments are always over the set $A(P)$ of ordinary atoms 
		that occur in the program $\Program$ at hand.
		The semantics of a ground external atom $\ext{g}{\vec{p}}{\vec{c}}$
		wrt.~an assignment $\Assignment$ is given by the value of a $1{+}k{+}l$-ary decidable
		\emph{two-valued (Boolean) oracle function} $\extfun{g}$ that is defined for all possible values
		of $\Assignment$, $\vec{p}$ and $\vec{c}$, where $k$ and $l$ are the lengths of $\vec{p}$ and $\vec{c}$,
		respectively, and $\Assignment$ must be complete over the ordinary atoms $A(P)$ in the program at hand.
		We make the restriction that for a fixed complete assignment $\Assignment$ and input $\vec{p}$,
		we have that $\extsem{g}{\Assignment}{\vec{p}}{\vec{c}} = \T$ only for finitely many different vectors $\vec{c}$.
		Thus,
		$\ext{g}{\vec{p}}{\vec{c}}$ is true relative
		to $\Assignment$, denoted $\Assignment \models \ext{g}{\vec{p}}{\vec{c}}$, if
		$\extsem{g}{\Assignment}{\vec{p}}{\vec{c}} = \T$ and false,
		denoted $\Assignment \not\models \ext{g}{\vec{p}}{\vec{c}}$,
		otherwise.
		
		Satisfaction of ordinary rules and ASP-programs~\cite{gelf-lifs-91}
		is then extended to 
		\hex-rules and -programs as follows.
		An assignment $\Assignment$ satisfies an atom $a$, denoted $\Assignment \models a$, if $\T a \in \Assignment$, and it does not satisfy it, denoted $\Assignment \not\models a$, otherwise.
		It satisfies a default-negated atom $\naf a$, denoted $\Assignment \models \naf a$, if $\Assignment \not\models a$, and it does not satisfy it, denoted $\Assignment \not\models \naf a$, otherwise.
		The truth value of a builtin atom $o_1 \circ o_2$ under $\Assignment$ depends only on $o_1$, $o_2$ and $\circ$ but not on the elements in $\Assignment$
		and is defined according to the standard semantics of operators $\circ$; for non-numeric operators this is according to an arbitrary but fixed ordering of constants.
		A rule $r$ is satisfied under assignment $\Assignment$, denoted $\Assignment \models r$, if $\Assignment \models a$ for some $a \in H(r)$ or $\Assignment \not\models a$ for some $a \in B(r)$.
		A \hex-program $\Program$ is satisfied under assignment $\Assignment$, denoted $\Assignment \models P$, if $\Assignment \models r$ for all $r \in P$.

		The answer sets of a \hex-program $\Program$ are defined as follows.
		Let the \emph{FLP-reduct} of $\Program$ wrt.~an assignment $\Assignment$
		be the set $f P^{\Assignment} = \{ r \in P \mid \Assignment \models b \text{ for all } b \in B(r) \}$
		of all rules whose body is satisfied by $\Assignment$.
		Then:
		\begin{definition}
			\label{def:answerset}
			A complete assignment $\Assignment$ is an answer set of a \hex-program $\Program$,
			if $\Assignment$ is a subset-minimal model of  
			$f P^{\Assignment}$ wrt.~positive literals in $\Assignment$ (i.e., $\{ \T a \in \Assignment \}$).
		\end{definition}

		\begin{example}
			\label{ex:id}
			Consider the \hex-program 
			$\Program = \{ p \leftarrow \ext{\mathit{id}}{p}{} \}$,
			where $\ext{\mathit{id}}{p}{}$ is true iff $p$ is true. Then 
			$\Program$ has the answer set $\Assignment_1 = \emptyset$; indeed it is
			a subset-minimal model of~$f P^{\Assignment_1} = \emptyset$.
		\end{example}
		
		For a given ground \hex-program $\Program$ we let $\mathcal{AS}(P)$ denote the set of all answer sets of $\Program$.

		For ordinary ASP-programs (i.e., \hex-programs without external atoms), the above definition of answer sets based on the FLP-reduct $f P^{\Assignment}$
		is equivalent to the original definition of answer sets by~\citeN{gelf-lifs-91}
		based on the \emph{GL-reduct} $P^{\Assignment} = \{ H(r) \leftarrow B^{+}(r) \mid r \in P, \Assignment \not\models b \text{ for all } b \in B^{-}(r) \}$.
		Further note that for a normal ordinary ASP-program $P$ and a complete assignment $\Assignment$, the reduct $P^{\Assignment}$ is always a positive program.
		This allows for an alternative characterization of answer sets of normal ASP-programs based on fixpoint iteration.
		For such a program $P$, we let $T_P(S) = \{ a \in H(r) \mid r \in P, B^{+}(r) \subseteq S \}$
		be the monotonic \emph{immediate consequence operator}, which derives the consequences of a set $S$ of atoms when applying the positive rules in $P$.
		Then the least fixpoint of $T_P$ over the empty set, denoted $\mathit{lfp}(T_P)$, is the \emph{unique} least model of $P$.
		Hence, an interpretation $I$, denoted as set of true atoms as by our simplified notation, is an answer set of a normal ASP-program $P$ if $I = \mathit{lfp}(T_{P^I})$.
		
		The answer sets of a general, possibly non-ground program $\Program$ are given by the answer sets
		$\mathcal{AS}(P) = \mathcal{AS}(\grnd_{\mathcal{C}}(P))$ of the \emph{grounding} $\grnd_{\mathcal{C}}(P)$ of $\Program$,
		which results from $\Program$ if all variables $\mathcal{V}$ in $\Program$ are replaced by all constants in $\mathcal{C}$ in all possible ways.
		In practice, suitable safety criteria guarantee that a finite subset of $\mathcal{C}$ suffices to compute the answer sets~\cite{efkr2016-aij}.
		For ordinary programs, one can use the (finite) Herbrand universe $\mathit{HU}(\Program)$ of all constants in $\Program$ for grounding.

		\leanparagraph{Saturation Technique}
		The saturation technique dates back to the $\Sigma^{\p}_2$-hardness proof of disjunctive ASP~\cite{DBLP:journals/amai/EiterG95},
		but was later exploited as a modeling technique, cf.~e.g.~\citeN{eik2009-rw}.
		It is applied for solving \conp-hard problems, which typically involve checking a condition \emph{for all} objects in a certain domain.

		To this end, the search space is defined in a program component $P_{\mathit{guess}}$ using disjunctions.
		Another program component $P_{\mathit{check}}$ checks if the current guess satisfies the property (e.g., being \emph{not} a valid 3-coloring) and derives a dedicated so-called saturation atom $\mathit{sat}$ in this case.
		A third program component $P_{\mathit{sat}}$ derives all atoms in $P_{\mathit{guess}}$ whenever $\mathit{sat}$ is true, i.e., it \emph{saturates the model}.
		This has the following effect: if all guesses fulfill the property, all atoms in $P_{\mathit{guess}}$ are derived for all guesses
		and the so-called \emph{saturation model} $I_{\mathit{sat}} = A(P_{\mathit{guess}} \cup P_{\mathit{check}})$ is an answer set of $P_{\mathit{guess}} \cup P_{\mathit{check}} \cup P_{\mathit{sat}}$;
		as said above, the complete assignment (interpretation) $I_{\mathit{sat}}$ is denoted as set of true atoms.
		On the other hand, if there is at least one guess which does not fulfill it, then $\mathit{sat}$ -- and possibly further atoms -- are not derived. Then, by minimality of answer sets, $I_{\mathit{sat}}$ is not an answer set.
		
		\begin{example}
			\label{ex:non3col}
			The program $P_{\mathit{non3col}} = F \cup P_{\mathit{guess}} \cup P_{\mathit{check}} \cup P_{\mathit{sat}}$ where
			{
			\begin{align*}
				P_{\mathit{guess}} = & \ \{ r(X) \vee g(X) \vee b(X) \leftarrow \mathit{node}(X) \} \\
				P_{\mathit{check}} = & \ \{ \mathit{sat} \leftarrow c(X), c(Y), \mathit{edge}(X,Y) \mid c \in \{ \mathit{r}, \mathit{g}, \mathit{b} \} \} \\
				P_{\mathit{sat}} = & \ \{ c(X) \leftarrow \mathit{node}(X), \mathit{sat} \mid c \in \{ \mathit{r}, \mathit{g}, \mathit{b} \} \}
			\end{align*}
			}
			has the answer set $I_{\mathit{sat}} = A(P_{\mathit{non3col}})$ iff the graph specified by facts $F$ is not 3-colorable.
			Otherwise its answer sets are proper subsets of $I_{\mathit{sat}}$ which represent valid 3-colorings.
		\end{example}

		Importantly, such a check \emph{cannot} be encoded as a \emph{normal} logic program in such a way that the program has an answer set iff the condition holds for all guesses (unless $\mathit{NP}=\mathit{coNP}$).
		Instead, one can only write a normal program which has \emph{no} answer set if the property holds for all guesses
		and non-existence of answer sets needs to be determined in the postprocessing.
		For instance, reconsider non-3-colorability and the following program
		{
		\begin{align*}
			P_{\mathit{3col}} = F \ \cup \{ & \mathit{c_1}(X) \leftarrow \mathit{node}(X), \naf \mathit{c_2}(X), \naf \mathit{c_3}(X) \mid \{ c_1, c_2, c_3 \} = \{ r, g, b \} \\
								& \leftarrow c(X), c(Y), \mathit{edge}(X,Y) \mid c \in \{ \mathit{r}, \mathit{g}, \mathit{b} \} \}\text{,}
		\end{align*}
		}
		where the graph is supposed to be defined by facts $F$ over predicates $\mathit{node}(\cdot)$ and $\mathit{edge}(\cdot, \cdot)$.
		Its answer sets correspond one-to-one to valid 3-colorings. However, in contrast to the program based on the saturation technique from Example~\ref{ex:non3col},
		it does \emph{not} have an answer set if there is no valid 3-coloring.
		For complexity reasons, it is not possible to define a normal program with an answer set that represents that there is \emph{no} such coloring (under the usual assumptions about complexity).
		This limitation inhibits that reasoning continues \emph{within} the program after checking the property.

	\section{Deciding Inconsistency of Normal Programs in Disjunctive ASP}
	\label{sec:embedding}

		In this section we first discuss restrictions of the saturation technique concerning the usage of default-negation within the checking part.
		These restrictions make it sometimes difficult to apply the technique even if complexity considerations imply that it is applicable in principle.
		In such cases, the technique appears to be inconvenient, especially for standard ASP users who are less experienced with saturation encodings.
		After discussing the restrictions using some examples, we present an encoding for deciding inconsistency of a normal ASP-program
		within disjunctive ASP.
		Different from the one presented by~\citeN{DBLP:journals/tplp/EiterP06}, ours uses conditional literals, which make it conceptually simpler.
		Moreover, it can also handle non-ground programs.
	
		\subsection{Restrictions of the Saturation Technique}
		\label{sec:restrictions}

			For complexity reasons, any problem in \conp can be polynomially reduced to brave reasoning over disjunctive ASP (the latter is $\Sigma^{\p}_2$-complete~\cite{flp2011-ai}),
			but the reduction is not always obvious. In particular, the saturation technique is difficult to apply if the property to check cannot be easily expressed without default-negation.
			Intuitively, this is because saturation works only if $I_{\mathit{sat}}$ is guaranteed to be an answer set of $P_{\mathit{guess}} \cup P_{\mathit{check}} \cup P_{\mathit{sat}}$ whenever no proper subset is one.
			While this is guaranteed for positive $P_{\mathit{check}}$, interpretation $I_{\mathit{sat}}$ might be unstable otherwise.
			
			\begin{example}
				\label{ex:vertexcover}
				A \emph{vertex cover} of a graph $\langle V, E \rangle$ is a subset $S \subseteq V$ of its nodes s.t.~each edge in $E$ is incident with at least one node in $S$.
				Deciding if a graph has \emph{no} vertex cover $S$ with size $|S| \leq k$ for some integer $k$ is \conp-complete.
				Consider $P_{\mathit{vc}}$ consisting of facts $F$ over $\mathit{node}$ and $\mathit{edge}$ and the following parts:
				{
				\begin{align*}
					P_{\mathit{guess}} = & \ \{ \mathit{in}(X) \vee \mathit{out}(X) \leftarrow \mathit{node}(X) \} \\
					P_{\mathit{check}} = & \ \{ \mathit{sat} \leftarrow \mathit{edge}(X,Y), \naf \mathit{in}(X), \naf \mathit{in}(Y); \\
										 & \ \phantom{\{}
										\mathit{sat} \leftarrow \mathit{in}(X_1), \ldots, \mathit{in}(X_{k+1}), X_1 \not= X_2, \ldots, X_k \not= X_{k+1} \} \\
					P_{\mathit{sat}} = & \ \{ \mathit{in}(X) \leftarrow \mathit{node}(X), \mathit{sat}; \ \mathit{out}(X) \leftarrow \mathit{node}(X), \mathit{sat} \}
				\end{align*}			
				}
				Program $P_{\mathit{guess}}$ guesses a candidate vertex cover $S$, $P_{\mathit{check}}$ derives $\mathit{sat}$ whenever for some edge $(u,v) \in E$ neither $u$ nor $v$ is in $S$ (thus $S$ is invalid),
				and $P_{\mathit{sat}}$ saturates in this case.
			\end{example}

			Observe that
			for inconsistent instances $F$ (e.g.~$\langle \{a,b,c,d\}, \{(a,b),(b,c),(c,d)\} \rangle$ with $k=1$), this encoding does not work as desired
			because model $I_{\mathit{sat}} = A(P_{\mathit{vc}})$ is unstable.
			More specifically, the instances of the first rule of $P_{\mathit{check}}$ are eliminated from the reduct $f P_{\mathit{vc}}^{I_\mathit{sat}}$ of $P_{\mathit{vc}}$ wrt.~$I_\mathit{sat}$ due to default-negation.
			But then, the least model of the reduct does not contain $\mathit{sat}$ or any atom $\mathit{in}(\cdot)$.
			Then, $I_{\mathit{<}} = I_{\mathit{sat}} \setminus ( \{ \mathit{sat} \} \cup \{ \mathit{in}(x) \mid x \in V \} )$ is a smaller model of the reduct
			and $I_{\mathit{sat}}$ is not an answer set of $P_{\mathit{vc}} \cup F$.
			
			In this example, the problem may be fixed by replacing literals $\naf \mathit{in}(X)$ and $\naf \mathit{in}(Y)$ by $\mathit{out}(X)$ and $\mathit{out}(Y)$, respectively.
			That is, instead of checking if a node is not in the vertex cover, one explicitly checks if it is out.
			However, the situation is more cumbersome if default-negation is not directly applied to the guessed atoms but to derived ones,
			as the following example demonstrates.
			
			
			

			\begin{example}
				\label{ex:hamiltoniancycle}
				A \emph{Hamiltonian cycle} in a directed graph $\langle V, E \rangle$ is a cycle that visits each node in $V$ exactly once.
				Deciding if a given graph has a Hamiltonian cycle is a well-known \np-complete problem; deciding if a graph does not have such a cycle is therefore \conp-complete.
				A natural attempt to solve the problem using saturation is as follows:
				{
				\begin{align}
					P_{\mathit{guess}} = & \ \{ \mathit{in}(X,Y) \vee \mathit{out}(X,Y) \leftarrow \mathit{arc}(X,Y) \} \label{ex:hamiltoniancycle:1} \\
					P_{\mathit{check}} = & \ \{ \mathit{sat} \leftarrow \mathit{in}(Y_1,X), \mathit{in}(Y_2,X), Y1 \not= Y2; \ \mathit{sat} \leftarrow \mathit{in}(X,Y_1), \mathit{in}(X,Y_2), Y1 \not= Y2 \label{ex:hamiltoniancycle:4} \\
										& \ \phantom{\{} \mathit{sat} \leftarrow \mathit{node}(X), \naf \mathit{hasIn}(X); \ \mathit{sat} \leftarrow \mathit{node}(X), \naf \mathit{hasOut}(X) \label{ex:hamiltoniancycle:5} \\
										& \ \phantom{\{} \mathit{hasIn}(X) \leftarrow \mathit{node}(X), \mathit{in}(Y,X); \ \mathit{hasOut}(X) \leftarrow \mathit{node}(X), \mathit{in}(X,Y) \} \label{ex:hamiltoniancycle:2} \\
					P_{\mathit{sat}} = & \ \{ \mathit{in}(X,Y) \leftarrow \mathit{arc}(X,Y), \mathit{sat}; \ \mathit{out}(X,Y) \leftarrow \mathit{arc}(X,Y), \mathit{sat} \label{ex:hamiltoniancycle:6} \}
				\end{align}
				}
				Program $P_{\mathit{guess}}$ guesses a candidate Hamiltonian cycle as a set of arcs.
				Program $P_{\mathit{check}}$ derives $\mathit{sat}$ whenever some node in $V$ does not have exactly one incoming and exactly one outgoing arc,
				and $P_{\mathit{sat}}$ saturates in this case.
				The check is split into two checks for at most (rules~(\ref{ex:hamiltoniancycle:4})) and at least (rules~(\ref{ex:hamiltoniancycle:5})) one incoming/outgoing arc.
				While the check if a node has at most one incoming/outgoing arc is possible using the positive rules~(\ref{ex:hamiltoniancycle:4}),
				the check if a node has at least one incoming/outgoing edge is more involved. In contrast to the check in Example~\ref{ex:vertexcover},
				one cannot perform it based on the atoms from $P_{\mathit{guess}}$ alone. Instead, auxiliary predicates $\mathit{hasIn}$ and $\mathit{hasOut}$ are defined by rules~(\ref{ex:hamiltoniancycle:2}).
				Unlike $\mathit{in}(\cdot,\cdot)$, the negation of $\mathit{hasIn}(\cdot)$ and $\mathit{hasOut}(\cdot)$ is not explicitly represented,
				thus default-negation is used in rules~(\ref{ex:hamiltoniancycle:5}) of $P_{\mathit{check}}$.
				However, this harms stability of $I_{\mathit{sat}}$: the graph $\langle \{ a,b,c \}, \{ (a,b), (b,a), (b,c), (c,b) \} \rangle$,
				which does not have a Hamiltonian cycle,
				causes $P_{\mathit{guess}} \cup P_{\mathit{check}} \cup P_{\mathit{sat}}$ to be inconsistent.
				This is due to default-negation in $P_{\mathit{check}}$, which eliminates rules~(\ref{ex:hamiltoniancycle:5}) from the reduct wrt.~$I_{\mathit{sat}}$,
				which has in turn a smaller model.
			\end{example}

			Note that in the previous example, for a fixed node $X$, the literal $\naf \mathit{hasOut}(X)$ is used to determine if all atoms $\mathit{in}(X,Y)$ are false (or equivalently:
			if all atoms $\mathit{out}(X,Y)$ are true). Here, default-negation can be eliminated on the ground level by replacing rule
				$\mathit{sat} \leftarrow \mathit{node}(X), \naf \mathit{hasOut}(X)$
			by
				$\mathit{sat} \leftarrow \mathit{node}(x), \mathit{out}(x,y_1), \ldots, \mathit{out}(x,y_n)$
			for all nodes $x \in V$ and all nodes $y_i$ for $1 \le i \le n$ such that $(x,y_i) \in E$.\footnote{On the non-ground level, this might be simulated using
				\emph{conditional literals} as supported by some reasoners,
			cf.~\citeN{gekakasc12a} and below.}
			But even this is not always possible, as shown with the next example.
			
			\begin{example}
				\label{ex:aspinconsistency}
				Deciding if a ground normal ASP-program $\Program$ is inconsistent is \conp-complete.
				An attempt to apply the saturation technique is as follows:
				{
				\begin{align}
					P' = & \ \{ \mathit{true}(a) \vee \mathit{false}(a) \mid a \in A(P) \} \label{def:arewriting:1} \\
						\cup & \ \{ \mathit{inReduct}(r) \leftarrow \{ \mathit{false}(b) \mid b \in B^{-}(r) \} \mid r \in P \} \label{def:arewriting:2} \\
						\cup & \ \{ \mathit{leastModel}(a) \leftarrow \mathit{inReduct}(r), \{ \mathit{leastModel}(b) \mid b \in B^{+}(r) \} \mid r \in P, a \in H(r) \} \label{def:arewriting:3} \\
						\cup & \ \{ \mathit{noAS} \leftarrow \mathit{false}(a), \mathit{leastModel}(a) \mid a \in A(P) \} \label{def:arewriting:4} \\
						\cup & \ \{ \mathit{noAS} \leftarrow \mathit{true}(a), \naf \mathit{leastModel}(a) \mid a \in A(P) \} \label{def:arewriting:5} \\
						\cup & \ \{ \mathit{true}(a) \leftarrow \mathit{noAS}; \ \mathit{false}(a) \leftarrow \mathit{noAS} \mid a \in A(P) \} \label{def:arewriting:6} \\
						\cup & \ \{ \mathit{inReduct}(r) \leftarrow \mathit{noAS} \} \label{def:arewriting:7}
				\end{align}
				}
				The idea is to guess all possible interpretations $I$ over the atoms $A(P)$ in $\Program$ (rules~(\ref{def:arewriting:1})).
				Next, rules~(\ref{def:arewriting:2}) identify the rules $r \in P$ which are in $P^I$ (modulo $B^{-}(r)$)\footnote{We can use the GL-reduct here as $P$ is an ordinary program.}; these are all rules $r \in P$ whose atoms $B^{-}(r)$ are all false.
				Rules~(\ref{def:arewriting:3}) compute the least model of the reduct by simulating fixpoint iteration under operator $T_P$ as shown in Section~\ref{sec:prelim}.
				Rules~(\ref{def:arewriting:4}) and~(\ref{def:arewriting:5}) compare the least model of the reduct to $I$:
				if this comparison fails, then $I$ is not an answer set and rules~(\ref{def:arewriting:6}) and~(\ref{def:arewriting:7}) saturate.
				While the comparison of the least model of the reduct to the original guess in rule~(\ref{def:arewriting:5}) is natural, it uses default-negation
				and destroys stability of $I_{\mathit{sat}}$ in general.
				In contrast to Example~\ref{ex:hamiltoniancycle}, eliminating the negation it is not straightforward, not even on the ground level.
			\end{example}
			
			We conclude that some problems involve checks which can easily be expressed using negation,
			but such a check within a saturation encoding may harm stability of the saturation model.
			In the next subsection, we present a valid encoding for checking inconsistency of normal programs (cf.~Example~\ref{ex:aspinconsistency}) within disjunctive ASP.
			Subsequently, we exploit this encoding for query answering over normal logic programs within disjunctive ASP in the next section.
		
		\subsection{A Meta-Program for Propositional Programs}

			We reduce the check for inconsistency of a normal logic program $\Program$ to brave reasoning over a disjunctive meta-program.
			The major part $M$ of the meta-program is static and consists of proper rules which are independent of $\Program$. The concrete program $\Program$ is then specified
			by facts $M^P$ which are added to the static part. The overall program $M \cup M^P$ is constructed such that it is consistent for all $\Program$ and
			its answer sets either represent the answer sets of $\Program$, or a distinguished answer set represents that $\Program$ is inconsistent.
			
			In this subsection we restrict the discussion to ground programs $\Program$.
			Moreover, we assume that all predicates in $\Program$ are of arity $0$. This is w.l.o.g.~because any atom $p(t_1, \ldots, t_\ell)$ can be replaced by an atom consisting of a new predicate $p'$ without any parameters.
			In the meta-program defined in the following, we let all atoms be new atoms which do not occur in $\Program$. We further use each rule $r \in P$ also as a constant in the meta-program.

			The static part consists of component $M_{\mathit{extract}}$ for the extraction of various information from
			the program encoding $M^P$, which we call $M^P_{\mathit{gr}}$ in this subsection to stress that $\Program$ must be ground,
			and a saturation encoding $M_{\mathit{guess}} \cup M_{\mathit{check}} \cup M_{\mathit{sat}}$
			for the actual inconsistency check.
			We first show the complete encoding and discuss its components subsequently.

\nop{
			\begin{definition}
				\label{def:grewriting}
				We define the \emph{meta-program} $M = M_{\mathit{extract}} \cup M_{\mathit{guess}} \cup M_{\mathit{check}} \cup M_{\mathit{sat}}$, where:
				\allowdisplaybreaks
				\begin{align}
					M_{\mathit{extract}} =& \ \{ \mathit{atom}(X) \leftarrow \mathit{head}(R,X); \ \mathit{atom}(X) \leftarrow \mathit{bodyP}(R,X); \ \mathit{atom}(X) \leftarrow \mathit{bodyN}(R,X) \} \label{def:grewriting:0} \\
						\cup & \ \{ \mathit{rule}(R) \leftarrow \mathit{head}(R,X); \ \mathit{rule}(R) \leftarrow \mathit{bodyP}(R,X); \ \mathit{rule}(R) \leftarrow \mathit{bodyN}(R,X) \} \label{def:grewriting:0a} \\[0.9ex]
					M_{\mathit{guess}} =& \ \{ \mathit{true}(X) \vee \mathit{false}(X) \leftarrow \mathit{atom}(X) \} \label{def:grewriting:1} \\[0.9ex]
					M_{\mathit{check}} =& \ \{ \mathit{inReduct}(R) \leftarrow \mathit{rule}(R), (\mathit{false}(X) : \mathit{bodyN}(R,X))  \label{def:grewriting:2} \\
						& \phantom{\ \{} \mathit{outReduct}(R) \leftarrow \mathit{rule}(R), \mathit{bodyN}(R,X), \mathit{true}(X)  \label{def:grewriting:3} \\[0.6ex]
						& \phantom{\ \{} \mathit{iter}(X, I) \vee \mathit{niter}(X, I) \leftarrow \mathit{true}(X), \mathit{int}(I) \label{def:grewriting:4} \\
						& \phantom{\ \{} \mathit{niter}(X, I) \leftarrow \mathit{false}(X), \mathit{int}(I)  \label{def:grewriting:4a} \\[0.6ex]
						& \phantom{\ \{} \mathit{notApp}(R) \leftarrow \mathit{outReduct}(R)  \label{def:grewriting:7} \\
						& \phantom{\ \{} \mathit{notApp}(R) \leftarrow \mathit{inReduct}(R), \mathit{bodyP}(R,X), \mathit{false}(X)  \label{def:grewriting:8} \\
						& \phantom{\ \{} \mathit{notApp}(R) \leftarrow \mathit{head}(R, X_1), \mathit{bodyP}(R, X_2), \mathit{iter}(X_1,I_1), \mathit{iter}(X_2,I_2), I_2 \ge I_1  \label{def:grewriting:9} \\[0.6ex]
						& \phantom{\ \{} \mathit{noAS} \leftarrow \mathit{true}(X), (\mathit{notApp}(R) : \mathit{head}(R,X))  \label{def:grewriting:10} \\
						& \phantom{\ \{} \mathit{noAS} \leftarrow \mathit{inReduct}(R), \mathit{head}(R,X), \mathit{false}(X), (\mathit{true}(Y) : \mathit{bodyP}(R,Y))  \label{def:grewriting:11} \\
						& \phantom{\ \{} \mathit{noAS} \leftarrow \mathit{true}(X), (\mathit{niter}(X, I) : \mathit{int}(I))  \label{def:grewriting:5} \\
						& \phantom{\ \{} \mathit{noAS} \leftarrow \mathit{iter}(X, I_1), \mathit{iter}(X, I_2), I_1 \not= I_2  \label{def:grewriting:6} \\[0.6ex]
						& \phantom{\ \{} \mathit{iter}_{<}(X, I) \leftarrow \mathit{false}(X), \mathit{int}(I) \label{def:grewriting:6a} \\
						& \phantom{\ \{} \mathit{iter}_{<}(X, I_2) \leftarrow \mathit{true}(X), \mathit{iter}(X, I_1), \mathit{int}(I_2), I_2 > I_1 \label{def:grewriting:6aa} \\[0.6ex]
						& \phantom{\ \{} \mathit{notApp}(R) \leftarrow \mathit{head}(R, X_1), \mathit{iter}(X_1, I_1), I_1 > 0, I_2 = I_1 - 1, \label{def:grewriting:6b} \\
						& \phantom{\{ \mathit{notApp}(R) \leftarrow\ \ } (\mathit{iter}_{<}(X_2, I_2) : \mathit{bodyP}(R, X_2)) \} \nonumber \\[0.9ex]
					M_{\mathit{sat}} =& \ \{ \mathit{true}(X) \leftarrow \mathit{atom}(X), \mathit{noAS}; \ \mathit{false}(X) \leftarrow \mathit{atom}(X), \mathit{noAS}  \label{def:grewriting:12} \\
						& \phantom{\ \{} \mathit{iter}(X, I) \leftarrow \mathit{atom}(X), \mathit{int}(I), \mathit{noAS}; \ \mathit{niter}(X, I) \leftarrow \mathit{atom}(X), \mathit{int}(I), \mathit{noAS}  \label{def:grewriting:13} \\
						& \phantom{\ \{} \mathit{inReduct}(R) \leftarrow \mathit{rule}(R), \mathit{noAS}; \ \mathit{outReduct}(R) \leftarrow \mathit{rule}(R), \mathit{noAS} \} \label{def:grewriting:14}
				\end{align}
			\end{definition}
}

			\begin{definition}
				\label{def:grewriting}
				We define the \emph{meta-program} $M = M_{\mathit{extract}} \cup M_{\mathit{guess}} \cup M_{\mathit{check}} \cup M_{\mathit{sat}}$, where:
				\allowdisplaybreaks
				\begin{align}
					M_{\mathit{extract}} =& \ \{ \mathit{atom}(X) \leftarrow \mathit{head}(R,X); \ \mathit{atom}(X) \leftarrow \mathit{bodyP}(R,X); \ \mathit{atom}(X) \leftarrow \mathit{bodyN}(R,X) \} \label{def:grewriting:0} \\
						\cup & \ \{ \mathit{rule}(R) \leftarrow \mathit{head}(R,X); \ \mathit{rule}(R) \leftarrow \mathit{bodyP}(R,X); \ \mathit{rule}(R) \leftarrow \mathit{bodyN}(R,X) \} \label{def:grewriting:0a} \\[0.9ex]
					M_{\mathit{guess}} =& \ \{ \mathit{true}(X) \vee \mathit{false}(X) \leftarrow \mathit{atom}(X) \} \label{def:grewriting:1} \\[0.9ex]
					M_{\mathit{check}} =& \ \{ \mathit{inReduct}(R) \leftarrow \mathit{rule}(R), (\mathit{false}(X) : \mathit{bodyN}(R,X))  \label{def:grewriting:2} \\
						& \phantom{\ \{} \mathit{outReduct}(R) \leftarrow \mathit{rule}(R), \mathit{bodyN}(R,X), \mathit{true}(X)  \label{def:grewriting:3} \\[0.6ex]						
						& \phantom{\ \{} \mathit{derivationSeq}(X, Y) \vee \mathit{derivationSeq}(Y, X) \leftarrow \mathit{true}(X), \mathit{true}(Y), X \not= Y \label{def:grewriting:3a} \\
						& \phantom{\ \{} \mathit{derivationSeq}(X, Z) \leftarrow \mathit{derivationSeq}(X, Y), \mathit{derivationSeq}(Y, Z) \label{def:grewriting:3b} \\[0.6ex]
						& \phantom{\ \{} \mathit{derivationSeq}(X_1, X_2) \leftarrow \mathit{head}(R, X_1), (\mathit{derivationSeq}(Y, X_1) : \mathit{bodyP}(R, Y)), \label{def:grewriting:3c} \\
						& \phantom{\ \{ \mathit{derivationSeq}(X_1, X_2) \leftarrow } \ \mathit{atom}(X_2), (\mathit{derivationSeq}(Y, X_2) : \mathit{bodyP}(R, Y)), X_2 > X_1 \nonumber \\[0.6ex]
						& \phantom{\ \{} \mathit{notApp}(R) \leftarrow \mathit{outReduct}(R)  \label{def:grewriting:7} \\
						& \phantom{\ \{} \mathit{notApp}(R) \leftarrow \mathit{inReduct}(R), \mathit{bodyP}(R,X), \mathit{false}(X)  \label{def:grewriting:8} \\
						& \phantom{\ \{} \mathit{notApp}(R) \leftarrow \mathit{head}(R, X_1), \mathit{bodyP}(R, X_2), \mathit{derivationSeq}(X_1, X_2)  \label{def:grewriting:9} \\[0.6ex]
						& \phantom{\ \{} \mathit{noAS} \leftarrow \mathit{true}(X), (\mathit{notApp}(R) : \mathit{head}(R,X))  \label{def:grewriting:10} \\
						& \phantom{\ \{} \mathit{noAS} \leftarrow \mathit{inReduct}(R), \mathit{head}(R,X), \mathit{false}(X), (\mathit{true}(Y) : \mathit{bodyP}(R,Y)) \}  \label{def:grewriting:11} \\[0.9ex]
					M_{\mathit{sat}} =& \ \{ \mathit{true}(X) \leftarrow \mathit{atom}(X), \mathit{noAS}; \ \mathit{false}(X) \leftarrow \mathit{atom}(X), \mathit{noAS}  \label{def:grewriting:12} \\
						& \phantom{\ \{} \mathit{derivationSeq}(X, Y) \leftarrow \mathit{atom}(X), \mathit{atom}(Y), \mathit{noAS} \label{def:grewriting:13} \\
						& \phantom{\ \{} \mathit{inReduct}(R) \leftarrow \mathit{rule}(R), \mathit{noAS}; \ \mathit{outReduct}(R) \leftarrow \mathit{rule}(R), \mathit{noAS} \} \label{def:grewriting:14}
				\end{align}
			\end{definition}

			This encoding is to be extended by the program-dependent part $M^P_{\mathit{gr}}$.
			Each rule of $\Program$ is represented by atoms of form
			$\mathit{head}(\mathit{r},a)$, $\mathit{bodyP}(\mathit{r},a)$, and $\mathit{bodyN}(\mathit{r},a)$,
			where $\mathit{r}$ is a rule from $\Program$ (used as new constant representing the respective rule),
			and $\mathit{head}(\mathit{r},a)$, $\mathit{bodyP}(\mathit{r},a)$ and $\mathit{bodyN}(\mathit{r},a)$
			denote that $a$ is an atom that occurs in the head, positive and negative body of rule $\mathit{r}$, respectively.
			For the following formalization we assume that in $P$,
			constraints of form
			$\leftarrow b_1,\dotsc, b_m, \naf\, b_{m+1}, \dotsc, \naf\, b_n$
			have already written to normal rules of form
			$f \leftarrow b_1,\dotsc, b_m, \naf\, b_{m+1}, \dotsc, \naf\, b_n, \naf f$,
			where $f$ is a new ground atom which does not appear elsewhere in the program,
			as discussed in Section~\ref{sec:prelim}.
			
			\begin{definition}
				\label{def:gProgramEncoding}
				For a ground normal logic program $\Program$ we let:
				\begin{align*}
					M^P_{\mathit{gr}} =& \ \{ \mathit{head}(r,h) \mid r \in P, h \in H(r) \} \\
						\cup & \ \{ \mathit{bodyP}(r,b) \mid r \in P, h \in B^{+}(r) \} \cup \{ \mathit{bodyN}(r,b) \mid r \in P, h \in B^{-}(r) \}
				\end{align*}
			\end{definition}

			We come now to the explanation of the static part $M$ of the meta-program.
			The component $M_{\mathit{extract}}$ uses rules~(\ref{def:grewriting:0}) and~(\ref{def:grewriting:0a}) to extract from $M^P_{\mathit{gr}}$ the sets of rules and atoms in $\Program$.
			
			The structure of the components $M_{\mathit{guess}}$, $M_{\mathit{check}}$ and $M_{\mathit{sat}}$ follows
			the basic architecture of saturation encodings presented in Section~\ref{sec:embedding}.
			Program $M_{\mathit{guess}}$ uses rule~(\ref{def:grewriting:1}) to guess an answer set candidate $I$ of program $\Program$,
			$M_{\mathit{check}}$ simulates the computation of the GL-reduct $P^I$ and checks
			if its least model coincides with $I$, and $M_{\mathit{sat}}$ saturates the model whenever this is \emph{not} the
			case.\footnote{Recall that the GL-reduct is equivalent to the FLP-reduct since $P$ is an ordinary ASP-program.}
			If all guesses fail to be answer sets, then every guess leads to saturation and the saturation model is an answer set.
			On the other hand, if at least one guess represents a valid answer set of $\Program$,
			then the saturation model is not an answer set due to subset-minimality.
			Hence, $M \cup M^P$ has exactly one (saturated) answer set if $\Program$ is inconsistent,
			and it has answer sets which are not saturated if $\Program$ is consistent, but none of them contains $\mathit{noAS}$.

\nop{
			We turn to the checking part $M_{\mathit{check}}$.
			Rules~(\ref{def:grewriting:2}) and~(\ref{def:grewriting:3}) compute for the current candidate $I$ the rules in $P^I$:
			a rule $r$ is in the reduct iff all atoms from $B^{-}(r)$ are false in $I$.
			Here, $(\mathit{false}(X) : \mathit{bodyN}(R,X))$ is a \emph{conditional literal} which extends the basic language from Section~\ref{sec:prelim}
			and evaluates to true iff $\mathit{false}(X)$ holds for all $X$ such that $\mathit{bodyN}(R,X)$ is true,
			i.e., all atoms in the negative body are false~\cite{gekakasc12a}.
			Rules~(\ref{def:grewriting:4}) and~(\ref{def:grewriting:4a}) simulate the computation of the least model $\mathit{lfp}(T_{P^I})$ of $P^I$ using fixpoint iteration.
			To this end, each atom $a \in I$ is assigned a guessed integer to represent an ordering of derivations during fixpoint iteration under $T_P$.
			More precisely, an atom $\mathit{iter}(a,i)$ denotes that atom $a$ is derived in the $i$-th iteration of $T_P$ (index origin: $0$).
			The number of required iterations is limited by the number of ground atoms in $\Program$ because the least model can only contain atoms
			which appear in $P$ and the fixpoint iteration stops if no new atoms are derivable.
			
			Rules~(\ref{def:grewriting:7})--(\ref{def:grewriting:6}) check if the current interpretation is \emph{not}
			an answer set of $\Program$ which can be justified by the guessed derivation sequence, and derive $\mathit{noAS}$ in this case.
			Importantly, $\mathit{noAS}$ is derived both if (i) $I$ is not an answer set at all, and if (ii) $I$ \emph{is} an answer set,
			but one that cannot be reproduced using the guessed derivation sequence.
			As a preparation for both checks (i) and (ii),
			rules~(\ref{def:grewriting:7})--(\ref{def:grewriting:9}) determine the rules $r \in P$ which
			are \emph{not} applicable in the fixpoint iteration (wrt.~the current derivation sequence) to justify their head atom $H(r)$ being true.
			A rule is not applicable if it is not in the reduct (rule~(\ref{def:grewriting:7})),
			if at least one positive body atom is false (rule~(\ref{def:grewriting:8})),
			or if it has a positive body atom which is derived in a
			later iteration (rule~(\ref{def:grewriting:9})) because then the rule cannot fire (yet)
			in the iteration the head atom was guessed to be derived.

			We can then perform the actual checks as follows.
			(i) For checking if $I$ is an answer set, rule~(\ref{def:grewriting:10}) checks if all atoms in $I$ are derived
			by some rule in $P^I$ (i.e., $I \subseteq \mathit{lfp}(T_{P^I})$). Conversely, rule~(\ref{def:grewriting:11})
			checks if all atoms derived by some rule in $P^I$ are also in $I$ (i.e., $I \supseteq \mathit{lfp}(T_{P^I})$).
			Overall, the rules~(\ref{def:grewriting:10})--(\ref{def:grewriting:11}) check if $I = \mathit{lfp}(T_{P^I})$.
			This check compares $I$ and $\mathit{lfp}(T_{P^I})$ only under the assumption that the guessed derivation sequence is valid.
			(ii)
			This validity remains to be checked. To this end,
			rule~(\ref{def:grewriting:5}) ensures that an iteration number is specified for all atoms which are true in $I$;
			in order to avoid default-negation we explicitly check if all atoms $\mathit{niter}(a, i)$ for all integers $i$ are true using a conditional literal.
			Rule~(\ref{def:grewriting:6}) guarantees that this number is unique for each atom.
			If one of these conditions does not apply, then the \emph{currently}
			guessed derivation order does not justify that $I$ is accepted as an answer set,
			hence it is dismissed by deriving $\mathit{noAS}$.
			However, it could still be an answer set justified by another (valid) derivation sequence.

			However, not all instances need the maximum of $|A(P)|$ iterations. Thus, without further means, there could be gaps in this sequence,
			which can lead to redundant solutions (e.g.~a fact could be derived immediately or in a later iteration).
			In order to eliminate them, rules~(\ref{def:grewriting:6a})--(\ref{def:grewriting:6b}) ensure that true atoms are derived in the earliest possible iteration.
			To this end, an atom of kind $\mathit{iter}_{<}(a, i)$ encodes that either $a$ is false (rule~(\ref{def:grewriting:6a})) or $a$ is derived in an iteration smaller than $i$ (rule~(\ref{def:grewriting:6aa})).
			Rule~(\ref{def:grewriting:6b}) defines then that a program rule is not applicable to justify its head atom being derived in iteration $i$,
			if all positive body atoms have been derived in an iteration smaller than $i-1$ (because then also the head atom could already be derived earlier).
}

			We turn to the checking part $M_{\mathit{check}}$.
			Rules~(\ref{def:grewriting:2}) and~(\ref{def:grewriting:3}) compute for the current candidate $I$ the rules in $P^I$:
			a rule $r$ is in the reduct iff all atoms from $B^{-}(r)$ are false in $I$.
			Here, $(\mathit{false}(X) : \mathit{bodyN}(R,X))$ is a \emph{conditional literal} which extends the basic language from Section~\ref{sec:prelim}
			and evaluates to true iff $\mathit{false}(X)$ holds for all $X$ such that $\mathit{bodyN}(R,X)$ is true,
			i.e., all atoms in the negative body are false~\cite{gekakasc12a}.
			Rules~(\ref{def:grewriting:3a})--(\ref{def:grewriting:3c}) simulate the computation of the least model $\mathit{lfp}(T_{P^I})$ of $P^I$ using fixpoint iteration.
			To this end, rule~(\ref{def:grewriting:3a}) guesses a derivation sequence over the true atoms $a \in I$ during fixpoint iteration under $T_P$.
			More precisely, an atom $\mathit{derivationSequence}(a,b)$ denotes that atom $a$ is derived before atom $b$; since this is a transitive property,
			rule~(\ref{def:grewriting:3b}) compute the closure.
			Although $T_P$ may derive multiple atoms in the same iteration, we guess a strict sequence here.
			Since this may lead to repetitive solutions as all permutations of atoms derived in the same iteration
			are valid, rule~(\ref{def:grewriting:3c}) enforces atoms to be derived in lexicographical order whenever possible.
			More precisely, after satisfaction of the positive body $B^{+}(r)$ of a rule $r$, the head atom $H(r)$
			must be derived before any lexicographically larger atom can be derived.
			
			Next, we check if the current interpretation $I$ is \emph{not}
			an answer set of $\Program$ which can be justified by the guessed derivation sequence, and derive $\mathit{noAS}$ in this case.
			Importantly, $\mathit{noAS}$ is derived both if (i) $I$ is not an answer set at all, and if (ii) $I$ \emph{is} an answer set,
			but one that cannot be justified using the guessed derivation sequence (i.e., the derivation sequence is invalid).
			As a preparation for this check,
			rules~(\ref{def:grewriting:7})--(\ref{def:grewriting:9}) determine the rules $r \in P$ which
			are \emph{not} applicable in the fixpoint iteration (wrt.~the current derivation sequence) to justify their head atom $H(r)$ being true.
			A rule is not applicable if it is not even in the reduct (rule~(\ref{def:grewriting:7})),
			if at least one positive body atom is false (rule~(\ref{def:grewriting:8})),
			or if it has a positive body atom which is derived in a
			later iteration (rule~(\ref{def:grewriting:9})) because then the rule cannot fire (yet)
			in the iteration the head atom was guessed to be derived.
			We can then perform the actual check as follows.
			Rule~(\ref{def:grewriting:10}) checks if all atoms in $I$ are derived
			by some rule in $P^I$ (i.e., $I \subseteq \mathit{lfp}(T_{P^I})$).
			Conversely, rule~(\ref{def:grewriting:11})
			checks if all atoms derived by some rule in $P^I$ are also in $I$ (i.e., $I \supseteq \mathit{lfp}(T_{P^I})$).
			Overall, the rules~(\ref{def:grewriting:10})--(\ref{def:grewriting:11}) check if $I = \mathit{lfp}(T_{P^I})$,
			and derive $\mathit{noAS}$ if this is not the case.

			Finally, the saturation part in rules~(\ref{def:grewriting:12})--(\ref{def:grewriting:14})
			derive all atoms in the program whenever the guess does not represent a valid answer set of $P$.
			
			\smallskip
			\noindent One can show that atom $\mathit{noAS}$ correctly represents the (in)consistency of $\Program$,
			as formalized by the next proposition.
			
			\addProposition{prop:genericRewriting}{
				For any ground normal logic program $\Program$, we have that
				\begin{compactenum}[(1)]
					\item if $P$ is inconsistent, $M \cup M^P_{\mathit{gr}}$ has exactly one answer set which contains $\mathit{noAS}$; and
					\item if $P$ is consistent, $M \cup M^P_{\mathit{gr}}$ has at least one answer set and none of them contains $\mathit{noAS}$.
				\end{compactenum}
			}
			
			\addProof{prop:genericRewriting}{
				(1) If $P$ is inconsistent, then each guess by rules~(\ref{def:grewriting:1}) represents an interpretation $I$ which is not an answer set of $P$.
					For each guess, rules~(\ref{def:grewriting:3a})--(\ref{def:grewriting:3b}) consider all possible derivation sequences under $T_{P^I}$;
					acyclicity is guaranteed by minimality of answer sets.
					Since $I$ is not an answer set of $P$, $I$ is not equivalent to the least model of $P^I$.
					Hence, we either have (i) $I \not= \mathit{lfp}(T_{P^I})$, or (ii) none of the guessed derivation sequences describes the fixpoint iteration under $T_{P^I}$.
					Rules~(\ref{def:grewriting:7})--(\ref{def:grewriting:9})
					identify rules which are not applicable for justifying their head atom being true, either because they are not in the reduct, their positive body is unsatisfied,
					or the head atom is derived in an earlier iteration than one of the positive body atoms.
					derive $\mathit{noAS}$ in such cases, as described in the text
					Based on this information, rule~(\ref{def:grewriting:10}) derives $\mathit{noAS}$ if at least one atom in $I$ cannot be derived as no rule which could derive it is applicable
					(i.e., $I \not\subseteq \mathit{lfp}(T_{P^I})$),
					and rule~(\ref{def:grewriting:11}) derives $\mathit{noAS}$ if a rule has a satisfied body but its head atom is false in $I$
					(i.e., $I \not\supseteq \mathit{lfp}(T_{P^I})$).
					Then, rules~(\ref{def:grewriting:10}) and~(\ref{def:grewriting:11}) together derive $\mathit{noAS}$ iff $I \not= \mathit{lfp}(T_{P^I})$.
					
					Rules~(\ref{def:grewriting:12}) to~(\ref{def:grewriting:14}) saturate whenever $\mathit{noAS}$ is true, i.e., whenever $I \not= \mathit{lfp}(T_{P^I})$.
					Clearly, if $P$ is inconsistent, then all guesses by rules~(\ref{def:grewriting:1}) fail to be answer sets, and thus $\mathit{noAS}$ and all atoms in $M^P$ are derived for all gusses.
					Then, the saturation model $I_{\mathit{sat}} = A(M^P)$ containing all atoms from $M^P$ is the unique answer set of $M^P$.

				(2) If $P$ is consistent, then at least one guess by rules rules~(\ref{def:grewriting:1}) represents a valid answer set of $P$.
					Then $\mathit{noAS}$ is not derived as shown in case~(1). By minimality of answer sets, $I_{\mathit{sat}}$ is excluded from being an answer set.
			}

		\subsection{A Meta-Program for Non-Ground Programs}

			We extend the encoding of a ground normal logic program as facts as by Definition~\ref{def:gProgramEncoding} to non-ground programs.
			The program-specific part is called $M^P_{\mathit{ng}}$ to stress that $P$ can now be non-ground (but also ground, which we consider a special case of non-ground programs).
			In the following, for a rule $r$ let $\vec{V}_r$ be the vector of unique variables occurring in $r$ in the order of appearance.

			The main idea of the following encoding is to interpret atoms as function terms.
			That is, for an atom $p(t_1, \ldots, t_\ell)$ we see $p$ as function symbol rather than predicate (recall that Section~\ref{sec:prelim} allows that $\mathcal{P}$ and $\mathcal{F}$ to share elements).
			Then, atoms, interpreted as function terms, can occur as parameters of other atoms.
			As before, we assume that constraints in $\Program$ have been rewritten to normal rules:

			\begin{definition}
				\label{def:ngProgramEncoding}
				For a (ground or non-ground) normal logic program $P$ we let:
				\begin{align*}
					M^P_{\mathit{ng}} =& \ \{ \mathit{head}(r(\vec{V}_r), h) \leftarrow \{ \mathit{head}(R_d,d) \mid d \in B^{+}(r) \} \mid r \in P, h \in H(r) \} \} \\
						\cup & \ \{ \mathit{bodyP}(r(\vec{V}_r), b) \leftarrow \{ \mathit{head}(R_d,d) \mid d \in B^{+}(r) \} \mid r \in P, b \in B^{+}(r) \} \\
						\cup & \ \{ \mathit{bodyN}(r(\vec{V}_r), b) \leftarrow \{ \mathit{head}(R_d,d) \mid d \in B^{+}(r) \} \mid r \in P, b \in B^{-}(r) \}
				\end{align*}
			\end{definition}

			For each possibly non-ground rule $r \in P$, we construct a unique identifier $r(\vec{V}_r)$ for each ground instance of $r$.
			It consists of $r$, used as a unique function symbol to identify the rule, and all variables in $r$ as parameters.
			As for the ground case, the head, the positive and the negative body are extracted from $r$.
			For ensuring safety, we add a \emph{domain atom} $\mathit{head}(R_d, d)$ for all positive body atoms $d \in B^{+}(r)$ to the body of the rule in the meta-program
			in order to instantiate it with all derivable ground instances.
			More precisely, for each positive body atom $d$ of the current rule,
			we use $R_d$ as a variable and add atom $\mathit{head}(R_d,d)$, to represents all (other) program rules of $P$ that instantiate the variables in atom $d$.
			This creates an instance of $r$ for all variable substitutions such that all body atoms of the instance are potentially derivable in the meta-program.
			
			\begin{example}
				\label{ex:nonground}
				Let $P \hspace{-0.3mm}=\hspace{-0.3mm} \{ f\colon\hspace{-0.15mm} d(a); \ r_1\colon\hspace{-0.15mm} q(X) \leftarrow d(X), \naf p(X); \ r_2\colon\hspace{-0.15mm} p(X) \leftarrow d(X), \naf q(X) \}$.
				We have:
				{
				\begin{align}
					M^P_{\mathit{ng}} =& \ \{ \mathit{head}(f, d(a)) \leftarrow \label{ex:nonground:1} \\
						& \phantom{\ \{} \mathit{head}(r_1(X), q(X)) \leftarrow \mathit{head}(R_{d(X)},d(X)) \label{ex:nonground:2} \\
						& \phantom{\ \{} \mathit{bodyP}(r_1(X), d(X)) \leftarrow \mathit{head}(R_{d(X)},d(X)) \label{ex:nonground:3} \\
						& \phantom{\ \{} \mathit{bodyN}(r_1(X), p(X)) \leftarrow \mathit{head}(R_{d(X)},d(X)) \label{ex:nonground:4} \\
						& \phantom{\ \{} \mathit{head}(r_2(X), p(X)) \leftarrow \mathit{head}(R_{d(X)},d(X)) \label{ex:nonground:5} \\
						& \phantom{\ \{} \mathit{bodyP}(r_2(X), d(X)) \leftarrow \mathit{head}(R_{d(X)},d(X)) \label{ex:nonground:6} \\
						& \phantom{\ \{} \mathit{bodyN}(r_2(X), q(X)) \leftarrow \mathit{head}(R_{d(X)},d(X)) \label{ex:nonground:7} \}
				\end{align}%
				}%
				We explain the encoding with the example of $r_1$. Since $r_1$ is non-ground, it may represent multiple ground instances, which are determined by the substitutions of $X$.
				We use $r_1(X)$ as identifier and define that, for any substitution of $X$, atom $q(X)$ appears in the head (cf.~rule~(\ref{ex:nonground:2})),
				$d(X)$ in the positive body (cf.~rule~(\ref{ex:nonground:3})) and $p(X)$ in the negative body (cf.~rule~(\ref{ex:nonground:4})).
				The domain of $X$ is defined by all atoms $d(X)$ which are potentially derivable by any rule (identified by variable $R_{d(X)}$),
				i.e., which occur in the head of a rule. We encode this by atom $\mathit{head}(R_{d(X)},d(X))$ in the bodies of the rules~(\ref{ex:nonground:2})--(\ref{ex:nonground:4}).
				The encoding works similar for $f$, cf.~rule~(\ref{ex:nonground:1}) and $r_2$, cf.~rules~(\ref{ex:nonground:5})--(\ref{ex:nonground:7}).
			\end{example}

			One can show that the encoding is still sound and complete for non-ground programs:

			\addProposition{prop:genericRewritingNonground}{
				For any normal logic program $P$, we have that
				\begin{compactenum}[(1)]
					\item if $P$ is inconsistent, $M \cup M^P_{\mathit{ng}}$ has exactly one answer set which contains $\mathit{noAS}$; and
					\item if $P$ is consistent, $M \cup M^P_{\mathit{ng}}$ has at least one answer set and none of them contains $\mathit{noAS}$.
				\end{compactenum}
			}
			
			\addProof{prop:genericRewritingNonground}{
				The proof follows from Proposition~\ref{prop:genericRewriting} and the observation that $M^P_{\mathit{ng}}$ simulates the construction of $M^P_{\mathit{gr}}$ for the grounded version $P_{\mathit{gr}}$ of $P$.
			}
			
	\section{Application: Query Answering over Subprograms}
	\label{sec:queryanswering}

		We now present a language extension with dedicated \emph{query atoms} which allow for answering queries over a subprogram within another program
		and accessing their results.
		This is intended to be a more convenient alternative to the saturation technique.
		Afterwards we show how the language feature can be reduced to our encoding for deciding inconsistency from the previous section.
		Finally we demonstrate this language extension with an example.
		
		Note that our approach focuses on simplicity (from user's perspective) and easy usability for typical use cases rather than high expressibility.
		Other approaches may have higher expressibility, but this comes at the price of a more complex semantics as e.g.~in the approach by \citeN{DBLP:journals/tplp/BogaertsJT16};
		we will discuss related approaches in more detail in Section~\ref{sec:related}.
	
		\subsection{Programs with Query Atoms}
		\label{sec:queryansweringencoding}

			In the following, a ground query $q$ is a set of ground literals (atoms or default-negated atoms) interpreted as conjunction.
			For an atom or default-negated atom $l$, let $\bar{l}$ be its negation, i.e., $\bar{l} = a$ if $l = \naf a$ and $\bar{l} = \naf a$ if $l = a$.
			We say that an interpretation $I$ satisfies a query $q$, denoted $I \models q$, if $I \models l$ for all $l \in q$.
			A \hex-program $P$ \emph{bravely entails} a query $q$, denoted $P \models_b q$, if $I \models q$ for some answer set $I$ of $P$;
			it \emph{cautiously entails} a query $q$, denoted $P \models_c q$, if $I \models q$ for all answer sets $I$ of $P$.

			\newcommand{\queryatom}[4]{\ifthenelse{\equal{#4}{}}{#2}{#2(#4)} \vdash_{#1} #3}
			
			A ground query $q$ over a subprogram $S$, possibly extended with input from the calling program, is then formalized as follows;
			in implementations, $S$ may be specified by its filename.
			\begin{definition}
				A \emph{query atom} is of form $\queryatom{t}{S}{q}{\vec{p}}$, where $t \in \{ b, c \}$ determines the type of the query,
				$S$ is a normal ordinary ASP-(sub-)program, $\vec{p}$ is a vector of predicates which specify the input, and $q$ is a ground query.
			\end{definition}
			

			We allow such query atoms to occur in bodies of general \hex-programs in place of ordinary atoms.
			Similar to Definition~\ref{def:rule} we can then define:

			\begin{definition}
				\label{def:programswithqueryatoms}
				A \emph{\hex-program with query atoms} is a set of rules
				\begin{equation*}
					\label{eq:rule}
				  a_1\lor\cdots\lor a_k \leftarrow b_1,\dotsc, b_m, \naf\, b_{m+1}, \dotsc, \naf\, b_n \ ,
				\end{equation*}
				where each $a_i$ is an ordinary atom and each~$b_j$
				is either an ordinary atom, an external atom or a query atom.
			\end{definition}
			
			The semantics as introduced in Section~\ref{sec:prelim} is then extended to \hex-programs with query atoms as follows.
			An interpretation $I$ satisfies a query atom $\queryatom{t}{S}{q}{\vec{p}}$, denoted $I \models \queryatom{t}{S}{q}{\vec{p}}$, if $S \cup \{ p(\vec{c}) \leftarrow \mid p \in \vec{p}, p(\vec{c}) \in I \} \models_t q$.
			The definitions of modelhood and of answer sets are then as for ordinary ASP-programs.

		\subsection{Reducing Query Answering to Inconsistency Checking}
		\label{sec:queryansweringencoding}

			Next, we show how programs with query atoms can be reduced to standard ASP-programs
			using our encoding for inconsistency checking.
			The reduction is based on the following observation:

			\addProposition{prop:qabrave}{
				For a normal ASP-program $P$ and a query $q$ we have that
				\begin{inparaenum}[(1)]
					\item $P \models_b q$ iff $P \cup \{ \leftarrow \bar{l} \mid l \in q \}$ is consistent; and
					\item $P \models_c q$ iff $P \cup \{ \leftarrow q \}$ is inconsistent.
				\end{inparaenum}
			}

			\addProof{prop:qabrave}{
				(1) By definition, $P \models_b q$ holds iff $q \subseteq I$ for some answer set $I$ of $P$.
					The constraints $\{ \leftarrow \naf a \mid a \in q \}$ eliminate exactly those answer sets $I$ of $P$, for which $I \models q$ does \emph{not} holds,
					while answer sets $I$ of $P$, for which $I \models q$ holds, are still answer sets of $P \cup \{ \leftarrow \naf a \mid a \in q \}$.
					Moreover, the addition of constraints cannot yield new answer compared to $P$.
					Hence, $P$ has an answer set $I$ with $I \models q$ iff $P \cup \{ \leftarrow \naf a \mid a \in q \}$ is consistent.
				(2) By definition, $P \models_c q$ holds iff $I \models q$ for all answer sets $I$ of $P$.
					The constraint $\{ \leftarrow q \}$ eliminates exactly those answer sets $I$ of $P$, for which $I \models q$ holds,
					while answer sets $I$ of $P$, for which $I \models q$ does \emph{not} hold, are still answer sets of $P \cup \{ \leftarrow q \}$.
					Moreover, the addition of constraints cannot yield new answer compared to $P$.
					Hence, $P$ has \emph{no} answer set iff every answer set $I$ of $P$ satisfies $q$.
			}

			This observation together with Proposition~\ref{prop:genericRewriting} implies the following result.
			
			\addProposition{prop:qaEncoding}{
				For a normal ASP-program $P$ be a predicate and query $q$ we have that
				\begin{compactenum}[(1)]
					\item $M \cup M^{P \cup \{ \leftarrow \bar{l} \mid l \in q \}}_{\mathit{ng}}$ is consistent and each answer set contains $\mathit{noAS}$ iff $P \not\models_b q$; and
					\item $M \cup M^{P \cup \{ \leftarrow q \}}_{\mathit{ng}}$ is consistent and each answer set contains $\mathit{noAS}$ iff $P \models_c q$.
				\end{compactenum}
			}
			
			\addProof{prop:qaEncoding}{
				(1) By Proposition~\ref{prop:qabrave} we have that $P \models_b q$ iff $P \cup \{ \leftarrow \bar{l} \mid l \in q \}$ is consistent (because then at least one answer set of $P$ does not satisfy all negations of the query literals, i.e., satisfies the query).
					If the program is consistent (and thus $P \models_b q$), then by Proposition~\ref{prop:genericRewriting} it has at least one answer set and none of its answer sets contains atom $\mathit{noAS}$.
					Conversely, if the program is inconsistent (and thus $P \not\models_b q$), then by Proposition~\ref{prop:genericRewriting} it has exactly (and thus at least) one answer set which contains atom $\mathit{noAS}$.

				(2) By Proposition~\ref{prop:qabrave} we have that $P \models_c q$ iff $P \cup \{ \leftarrow q \}$ is inconsistent (because then every answer set of $P$ violates the added constraint, i.e., satisfies the query).
					If the program is consistent (and thus $P \not\models_c q$), then by Proposition~\ref{prop:genericRewriting} it has at least one answer set which does not contain atom $\mathit{noAS}$.
					Conversely, if the program is inconsistent (and thus $P \models_c q$), then by Proposition~\ref{prop:genericRewriting} it has exactly (and thus at least) one answer set which contains atom $\mathit{noAS}$.
			}

			The idea of evaluation based reduction is now to apply Proposition~\ref{prop:qaEncoding} to each query atom in the program at hand.
			That is, for each query atom $\queryatom{t}{S}{q}{\vec{p}}$ we create a copy of
			$M \cup M^{S \cup \{ \leftarrow \bar{l} \mid l \in q \}}_{\mathit{ng}}$ if $t = b$
			and of $M \cup M^{S \cup \{ \leftarrow q \}}_{\mathit{ng}}$ if $t=c$.
			Importantly, each copy must use a different name space of atoms, which is realized by adding $\queryatom{t}{S}{q}{\vec{p}}$ as an index to every atom.
			By Proposition~\ref{prop:genericRewritingNonground} we can then use the atom $\mathit{noAS}_{\queryatom{t}{S}{q}{\vec{p}}}$ to check consistency of
			the respective copy of $S \cup \{ \leftarrow \bar{l} \mid l \in q \}$ resp.~$M \cup M^{S \cup \{ \leftarrow q \}}_{\mathit{ng}}$,
			which corresponds by Proposition~\ref{prop:qaEncoding} to the answer to the query $S \models_t q$.

			Formally we define the following translation, where $\vec{X}^{\mathit{ar}(p)}$ denotes a vector of variables of length $\mathit{ar}(p)$:
			\smallskip
			\newcommand{\replacement}{/}
			\begin{align}
				[P] = & P|_{\queryatom{c}{S}{q}{\vec{p}} \replacement \mathit{noAS}_{\queryatom{c}{S}{q}{\vec{p}}}, \queryatom{b}{S}{q}{\vec{p}} \replacement \naf \mathit{noAS}_{\queryatom{b}{S}{q}{\vec{p}}}}|_{\naf \naf a \replacement a} \label{translation:1} \\
					\cup & \bigcup\nolimits_{\queryatom{b}{S}{q}{\vec{p}} \text{ in } P} \big(M \cup M^{S \cup \{ \leftarrow \bar{l} \mid l \in q \}}_{\mathit{ng}}\big)\big|_{a \replacement a_{\queryatom{b}{S}{q}{\vec{p}}}}
							\cup \bigcup\nolimits_{\queryatom{c}{S}{q}{\vec{p}} \text{ in } P} \big(M \cup M^{S \cup \{ \leftarrow q \}}_{\mathit{ng}}\big)\big|_{a \replacement a_{\queryatom{c}{S}{q}{\vec{p}}}} \label{translation:2} \\
					\cup & \bigcup\nolimits_{\queryatom{t}{S}{q}{\vec{p}} \text{ in } P,\ p \in \vec{p}} \big(\{ \mathit{head}(r_{\mathit{p}}(\vec{X}^{\mathit{ar}(p)}), p(\vec{X}^{\mathit{ar}(p)})) \leftarrow p(\vec{X}^{\mathit{ar}(p)}) \} \big)|_{a \replacement a_{\queryatom{t}{S}{q}{\vec{p}}}} \label{translation:3}
			\end{align}
			\smallskip
			
			Here, the program part in line~(\ref{translation:1}) denotes program $P$ after first replacing
			every query atom of kind $\queryatom{c}{S}{q}{\vec{p}}$ (for some $S$, $\vec{p}$ and $q$) by the new ordinary atom $\mathit{noAS}_{\queryatom{c}{S}{q}{\vec{p}}}$
			and every query atom of kind $\queryatom{b}{S}{q}{\vec{p}}$ (for some $S$, $\vec{p}$ and $q$) by the new ordinary literal $\naf \mathit{noAS}_{\queryatom{b}{S}{q}{\vec{p}}}$,
			and then eliminating double negation (which may be introduced by the previous replacement if in the program $P$ a query atom with a brave query appears in the default-negated part of a rule).
			This part of the rewriting makes sure that the $\mathit{noAS}$ atoms of our encoding from the previous section is accessed in place of the original query atoms following Proposition~\ref{prop:qaEncoding}.
			
			Next, the program part in line~(\ref{translation:2}) defines these $\mathit{noAS}$ atoms using the encoding from the previous section.
			Here, the expression $|_{a \replacement a_{\queryatom{b}{S}{q}{\vec{p}}}}$ denotes the program after replacing each atom $a$ by $a_{\queryatom{b}{S}{q}{\vec{p}}}$ (likewise for $\queryatom{c}{S}{q}{\vec{p}}$).
			This realizes different namespaces, i.e., for every query atom $\queryatom{b}{S}{q}{\vec{p}}$ resp.~$\queryatom{c}{S}{q}{\vec{p}}$ in $P$,
			a separate copy of $M$ and $M^{S \cup \{ \leftarrow \bar{l} \mid l \in q \}}_{\mathit{ng}}$ resp.~$M^{S \cup \{ \leftarrow q \}}_{\mathit{ng}}$
			with disjoint vocabularies is generated. Then each such copy uses also a separate atom $\mathit{noAS}_{\queryatom{b}{S}{q}{\vec{p}}}$ resp.~$\mathit{noAS}_{\queryatom{c}{S}{q}{\vec{p}}}$
			which represents by Proposition~\ref{prop:qaEncoding} the answer to query $q$ as expected by the part in line~(\ref{translation:1}).

			Finally, we must add the input facts to $S$.
			To this end, sets $M^{S \cup \{ \leftarrow q \}}_{\mathit{ng}}$ resp.~$M^{S \cup \{ \leftarrow \bar{l} \mid l \in q \}}_{\mathit{ng}}$
			must be extended. For each input predicate $p \in \vec{p}$, each atom of kind $p(\ \cdots)$ must be added as additional rule $p(\ \cdots) \leftarrow$, which is encoded akin to Definition~\ref{def:ngProgramEncoding}.
			This is realized in line~(\ref{translation:3}), which translates the values of $p$-atoms of the calling program to facts in the subprogram.

			One can formally show that $[P]$ resembles the semantics of programs with query atoms, as by Definition~\ref{def:programswithqueryatoms}:

			\addProposition{prop:queryAtoms}{
				For a logic program $P$ with query atoms
				we have that $\mathcal{AS}(P)$ and $\mathcal{AS}([P])$, projected to the atoms in $P$, coincide.
			}
			
			\addProof{prop:queryAtoms}{
				%
				By Proposition~\ref{prop:qaEncoding}, we have that for $\queryatom{b}{S}{q}{\vec{p}}$, $\big(M \cup M^{S \cup \{ \leftarrow \bar{l} \mid l \in q \}}_{\mathit{ng}}\big)\big|_{a \replacement a_{\queryatom{b}{S}{q}{\vec{p}}}}$
				has at least one answer set and in each answer set the atom $\mathit{noAS}_{\queryatom{b}{S}{q}{\vec{p}}}$ represents satisfaction of the query $S \models_b q$.
				Similarly, for $\queryatom{c}{S}{q}{\vec{p}}$, $\bigcup_{\queryatom{c}{S}{q}{\vec{p}} \text{ in } P} \big(M \cup M^{S \cup \{ \leftarrow q \}}_{\mathit{ng}}\big)\big|_{a \replacement a_{\queryatom{c}{S}{q}{\vec{p}}}}$
				has at least one answer set and in each answer set the atom $\mathit{noAS}_{\queryatom{c}{S}{q}{\vec{p}}}$ represents satisfaction of the query $S \models_c q$.
				Since all copies of $M \cup M^{S \cup \{ \leftarrow \bar{l} \mid l \in q \}}_{\mathit{ng}}$ resp.~$M \cup M^{S \cup \{ \leftarrow q \}}_{\mathit{ng}}$
				use disjoint vocabularies, the program line~(\ref{translation:3}) still has at least one answer set and in each answer set
				atom $\mathit{noAS}_{\queryatom{t}{S}{q}{\vec{p}}}$ represents satisfaction of query $q$ over $S$ for all query atoms $\queryatom{t}{S}{q}{\vec{p}}$ in $P$.
				
				Next, line~(\ref{translation:2}) translates the values of $p$-atoms of the calling program to facts in the subprogram using the same encoding as by Definition~\ref{def:ngProgramEncoding}
				as for the facts which are directly in $S$. Thus, for given values of the $p$-atoms under the current interpretation $I$, the program in lines~(\ref{translation:2})--(\ref{translation:3}) behaves
				as if $\{ p(\vec{c}) \leftarrow \mid p(\vec{c}) \in I \}$ was already in $S$.
				
				Finally, line~(\ref{translation:1}) uses atoms $\queryatom{t}{S}{q}{\vec{p}}$ in place of $\queryatom{t}{S}{q}{\vec{p}}$, which, however, correspond to each other one-by-one.
				Thus, program $[P]$ behaves as desired.
			}

			Note that while the construction of $[P]$ may not be trivial, this does not harm usability from user's perspective.
			This is because the rewriting concerns only the implementer of a reasoner for programs with queries over subprograms, while the user can simply use query atoms.

		\subsection{Using Query Atoms}
		\label{sec:defaultNegation}

			Query answering over subprograms can now be exploited as a modeling technique to check a criterion for all objects in a domain.
			As observed in Section~\ref{sec:embedding}, saturation may fail in cases where the check involves default-negation.
			Moreover, saturation is an advanced technique which might be not intuitive for less experienced ASP users
			(it was previously called `hardly usable by ASP laymen'~\cite{DBLP:journals/corr/abs-1107-5742}).
			Thus, even for problems whose conditions can be expressed by positive rules, an encoding based on query answering might be easier to understand.
			To this end, one starts with a program $P_{\mathit{guess}}$ which spans a search space of all objects to check.
			As with saturation, $P_{\mathit{check}}$ checks if the current guess satisfies the criteria and derives a dedicated atom $\mathit{ok}$ in this case.
			However, instead of saturating the interpretation whenever $\mathit{ok}$ is true, one now checks if $\mathit{ok}$ is cautiously entailed by $P_{\mathit{guess}} \cup P_{\mathit{check}}$
			using a rule of kind $\mathit{allOk} \leftarrow P_{\mathit{guess}} \cup P_{\mathit{check}} \vdash_c \mathit{ok}$.
			Importantly, the user does not need to deal with saturation and can use default-negation in the checking part $P_{\mathit{check}}$.

			\nop{
			\begin{example}
				For the program from Example~\ref{ex:non3col},
				$[\{ \mathit{not3col} \leftarrow P_{\mathit{guess}} \cup P_{\mathit{check}} \vdash_c \mathit{sat} \}]$
				always has at least one answer set
				and in each answer set, $\mathit{not3col}$ represents that the graph is not 3-colorable.
			\end{example}
			}
			
			\begin{example}
				\label{ex:hamiltoniancycleqa}
				Recall Example~\ref{ex:hamiltoniancycle} where we want to check if a graph does not possess a Hamiltonian cycle.
				Consider the following programs:
				\begin{align}
					P_{\mathit{guess}} = & \ \{ \mathit{in}(X,Y) \vee \mathit{out}(X,Y) \leftarrow \mathit{arc}(X,Y) \} \label{ex:hamiltoniancycleqa:1} \\
					P_{\mathit{check}} = & \ \{ \mathit{hasIn}(X) \leftarrow \mathit{node}(X), \mathit{in}(Y,X); \ \mathit{hasOut}(X) \leftarrow \mathit{node}(X), \mathit{in}(X,Y) \} \label{ex:hamiltoniancycleqa:2} \\
										& \ \phantom{\{} \mathit{invalid} \leftarrow \mathit{node}(X), \naf \mathit{hasIn}(X); \ \mathit{invalid} \leftarrow \mathit{node}(X), \naf \mathit{hasOut}(X) \label{ex:hamiltoniancycleqa:5} \\
										& \ \phantom{\{} \mathit{invalid} \leftarrow \mathit{in}(Y_1,X), \mathit{in}(Y_2,X), Y_1 \not= Y_2; \ \mathit{invalid} \leftarrow \mathit{in}(X,Y_1), \mathit{in}(X,Y_2), Y1 \not= Y2 \label{ex:hamiltoniancycleqa:4} \\
					P = & \ \{ \mathit{noHamiltonian} \leftarrow \queryatom{c}{P_{\mathit{guess}} \cup P_{\mathit{check}} \cup F}{\mathit{invalid}}{} \}
				\end{align}

				Assume that $F$ represents a graph encoded by predicates $\mathit{node}(\cdot)$ and $\mathit{edge}(\cdot, \cdot)$.
				Then $P_{\mathit{guess}}$ guesses all potential Hamiltonian cycles and $P_{\mathit{check}}$ decides for each of them whether it is valid or invalid;
				in the latter cae, atom $\mathit{invalid}$ is derived.
				We use both programs together as a subprogram which we access from $P$.
				We have that $P$ has a single answer set which contains $\mathit{noHamiltonian}$ if and only if the graph at hand does not contain a Hamiltonian cycle.
				If there are Hamiltonian cycles, then $P$ has at least one answer set but none of the answer sets contains atom $\mathit{noHamiltonian}$.

				Recall that the same implementation of the checking part did not work in our saturation encoding from Example~\ref{ex:hamiltoniancycleqa} due to default-negation.
				Also note that even if the encoding is adopted such that saturation becomes applicable, the cyclic nature of a saturation encoding
				makes it more difficult to understand then the strictly acyclic encoding based on query atoms.
			\end{example}

			We adopt the example to show how to use input to a subprogram:

			\begin{example}[cont'd]
				Consider the following programs:
				\begin{align}
					P_{\mathit{check}} = & \ \{ \mathit{hasIn}(X) \leftarrow \mathit{node}(X), \mathit{in}(Y,X); \ \mathit{hasOut}(X) \leftarrow \mathit{node}(X), \mathit{in}(X,Y) \} \label{ex:hamiltoniancycleqa:2} \\
										& \ \phantom{\{} \mathit{invalid} \leftarrow \mathit{node}(X), \naf \mathit{hasIn}(X); \ \mathit{invalid} \leftarrow \mathit{node}(X), \naf \mathit{hasOut}(X) \label{ex:hamiltoniancycleqa:5} \\
										& \ \phantom{\{} \mathit{invalid} \leftarrow \mathit{in}(Y_1,X), \mathit{in}(Y_2,X), Y_1 \not= Y_2; \ \mathit{invalid} \leftarrow \mathit{in}(X,Y_1), \mathit{in}(X,Y_2), Y1 \not= Y2 \label{ex:hamiltoniancycleqa:4} \\
					P = & \ \{ \mathit{in}(X,Y) \vee \mathit{out}(X,Y) \leftarrow \mathit{arc}(X,Y) \\
						& \ \phantom{\{} \mathit{notHamiltonian} \leftarrow \queryatom{c}{P_{\mathit{check}}}{\mathit{invalid}}{\mathit{node}, \mathit{arc}, \mathit{in}} \} \cup F
				\end{align}

				Here, we make a guess in $P$ in send each candidate separately to a subprogram for verification.
				To this end, for each guess we add all true atoms over the predicates
				$\mathit{node}$, $\mathit{arc}$, and $\mathit{in}$ to the subprogram $P_{\mathit{check}}$
				and check if the guess is a valid Hamiltonian cycle.
				If this is not the case, $\mathit{invalid}$ is derived in the subprogram, which satisfies the query in $P$
				and derives $\mathit{notHamiltonian}$.
			\end{example}

			In summary, we have provided an encoding for a particular $\mathit{coNP}$-hard problem, namely deciding inconsistency of a normal ASP-program.
			Based on this encoding, have a developed a language extension towards queries over normal subprograms.
			While our encoding makes use of the saturation technique, the user of the language extension does not get in touch with it.
			That is, the saturation technique is hidden from the user and needs to be implemented only once, namely as part of a solver for the language extension,
			but not when solving a concrete problem on top of it.
			
	\section{Reasons for Inconsistency of \hex-Programs}
	\label{sec:inconsistencyanalysis}

		In this section we consider programs $P$ which are extended by a set of (input) atoms $F \subseteq D$ from a given domain $D$, which are added as facts.
		More precisely, for a given set $F \subseteq D$, we consider $P \cup \toFacts{F}$, where $\toFacts{F} = \{ a \leftarrow \mid a \in F \}$ is the representation of $F$ transformed to facts.
		We assume that the domain of atoms $D$, which can be added as facts, might only occur in bodies $B(P)$ of rules in $P$, but are not in their heads $H(P)$.
		This is in spirit of the typical usage of (ASP- and \hex-)programs, where proper rules (IDB; intensional database) encode the problem at hand and are fixed,
		while facts (EDB; extensional database) specify the concrete instance and are subject to change.
		
		We present a concept which expresses the reasons for the inconsistency of \hex-program $P \cup \toFacts{F}$ in terms of $F$.
		That is, we want to identify a sufficient criterion wrt.~the problem instance $F$ which guarantees that $P \cup \toFacts{F}$ does not possess any answer sets.
		This leads to a characterization of \emph{classes} of inconsistent instances.
		While identifying such classes is a natural task by itself, a concrete application can be found in context of \hex-program evaluation and will be presented in the next section.

		\subsection{Formalizing Inconsistency Reasons}
		\label{sec:inconsistencyanalysis:formalizing}

			Inspired by inconsistency explanations for multi-context systems~\cite{DBLP:journals/ai/EiterFSW14},
			we propose to make an inconsistency reason dependent on atoms from $D$
			which \emph{must occur} resp.~\emph{must not occur} in $F$ such that $P \cup \toFacts{F}$ is inconsistent, while the remaining atoms from $D$ might either occur or not occur in $F$ without influencing (in)consistency.
			
			We formalize this idea as follows:
			
			\begin{definition}[Inconsistency Reason (IR)]
				\label{def:inconsistencyReason}
				Let $P$ be a \hex-program and $D$ be a domain of atoms.
				An \emph{inconsistency reason (IR)} of $P$ wrt.~$D$
				is a pair $R = (R^{+}, R^{-})$ of sets of atoms $R^{+} \subseteq D$ and $R^{-} \subseteq D$ with $R^{+} \cap R^{-} = \emptyset$
				s.t.~$P \cup \toFacts{F}$ is inconsistent for all $F \subseteq D$ with $R^{+} \subseteq F$ and $R^{-} \cap F = \emptyset$.
			\end{definition}

			Here, $R^{+}$ resp.~$R^{-}$ define the sets of atoms which must be present resp.~absent in $F$ such that
			$P \cup \toFacts{F}$ is inconsistent, while atoms from $D$ which are neither in $R^{+}$ nor in $R^{-}$ might be arbitrarily added or not without affecting inconsistency.

			\begin{example}
				An IR of the program $P = \{ \leftarrow a, \naf c; \ d \leftarrow b. \}$ 
				wrt.~$D = \{ a, b, c \}$ is $R = (\{ a \}, \{ c \})$
				because $P \cup \toFacts{F}$ is inconsistent for all $F \subseteq D$ whenever $a \in F$ and $c \not\in F$,
				while $b$ can be in $F$ or not without affecting (in)consistency.
			\end{example}

			Note that in contrast to work on ASP debugging, which we discuss in more detail in Section~\ref{sec:related} and aims at finding (human-readable) explanations for inconsistency of
			single programs, we do \emph{not} focus on debugging a particular program instance.
			Instead, our notion rather aims at identifying \emph{classes of program instances} depending on the input facts, which are inconsistent.
			As a consequence, the techniques developed for ASP debugging cannot directly be used. 

			For a given program $\Program$ and a domain $D$,
			we say that an IR $R_1 = (R_1^{+}, R_1^{-})$ is \emph{smaller than} an IR $R_2 = (R_2^{+}, R_2^{-})$,
			if $R_1^{+} \subseteq R_2^{+}$ and $R_1^{-} \subseteq R_2^{-}$; it is \emph{strictly smaller} if at least one of the inclusions is proper.
			An IR is \emph{subset-minimal}, if there is no strictly smaller IR wrt.~$\Program$ and $D$.			
			Although small IRs are typically preferred, we do not formally require them to be minimal in general.

			In general there are multiple IRs, some of which might not be minimal.
			For instance, the program $P = \{ \leftarrow a; \leftarrow b \}$
			has inconsistency reasons $R_1 = (\{ a \}, \emptyset)$, $R_2 = (\{ b \}, \emptyset)$) and $R_3 = (\{ a, b \}, \emptyset)$ wrt.~$D = \{ a, b \}$,
			where $R_1$ and $R_2$ are minimal but $R_3$ is not.
			On the other hand, a program $P$ might not have any IR at all if $P \cup \toFacts{F}$ is consistent for all $F \subseteq D$. This is the case, for instance, for the empty \hex-program $P = \emptyset$.
			However, one can show that there is always at least one IR if $P \cup \toFacts{F}$ is inconsistent for some $F \subseteq D$.

			\addProposition{prop:inconsistencyReasonExistence}{
				For all \hex-programs $P$ and domains $D$ such that $P \cup \toFacts{F}$ is inconsistent for some set $F \subseteq D$ of atoms,
				then there is an IR of $P$ wrt.~$D$.
			}

			\addProof{prop:inconsistencyReasonExistence}{
				Take some $F$ such that $P \cup \toFacts{F}$ is inconsistent
				and consider $R = (R^{+}, R^{-})$ with $R^{+}=F$ and $R^{-}=D \setminus F$.
				Then the only $F'$ with $R^{+} \subseteq F'$ and $R^{-} \cap F' = \emptyset$
				is $F$ itself. But $P \cup \toFacts{F}$ is inconsistent by assumption.
			}

		\subsection{Computational Complexity}
		\label{sec:inconsistencyanalysis:complexity}

			We now discuss the computational complexity of reasoning problems in context of computing inconsistency reasons.
			Similarly as for the results by~\citeN{flp2011-ai} we assume in the following that oracle functions can be evaluated in polynomial time wrt.~the size of their input.

			We start with the problem of checking if a candidate IR of a program is a true IR.
			The complexities correspond to those of inconsistency checking over the respective program class,
			which is because the problems of candidate IR checking and inconsistency checking can be reduced to each other.
			\addProposition{prop:irComplexityIRChecking}{
				Given a \hex-program $P$, a domain $D$, and a pair $R = (R^{+}, R^{-})$ of sets of atoms
				with $R^{+} \subseteq D$, $R^{-} \subseteq D$ and $R^{+} \cap R^{-} = \emptyset$.
				Deciding if $R$ is an IR of $P$ wrt.~$D$ is
				\begin{enumerate}[(i)]
					\item $\Pi^P_2$-complete for general \hex-programs and for ordinary disjunctive ASP-programs, and
					\item $\mathit{coNP}$-complete if $P$ is an ordinary disjunction-free program.
				\end{enumerate}
			}

			\addProof{prop:irComplexityIRChecking}{
				\noindent\textbf{Hardness:}
					Checking if $P$ does not have any answer set is a well-known problem that is $\Pi^P_2$-complete for (i) and $\mathit{coNP}$-complete for (ii), see~\citeN{flp2011-ai}.
					
					We reduce the problem to checking if a given $R = (R^{+}, R^{-})$ is an IR of $P$ wrt.~a domain $D$,
					which shows that the latter cannot be easier.
					Consider $R = (\emptyset, \emptyset)$ and domain $D = \emptyset$. Then the only $F \subseteq D$
					with $\emptyset \subseteq F$ and $D \cap F = \emptyset$
					is $F = \emptyset$. Then $R$ is an IR iff $P \cup \toFacts{\emptyset} = P$ is inconsistent,
					which allows for reducing the inconsistency check to the check of $R$ for being an IR of $P$ wrt.~$D$.

				\noindent\textbf{Membership:}
					Consider
					$P' = P \cup \{ x \leftarrow \mid x \in R^{+} \} \cup \{ x \leftarrow \naf \mathit{nx}; \ \mathit{nx} \leftarrow \naf x \mid x \in D \setminus R^{-} \}$,
					where $\mathit{nx}$ is a new atom for all $x \in D$.
					Then $P'$ has an answer set iff for some $F \subseteq D$
					with $R^{+} \subseteq F$ and $R^{-} \cap F = \emptyset$ we have that $P \cup \toFacts{F}$ has an answer set.
					Conversely, $P'$ does not have an answer set, iff for all $F \subseteq D$
					with $R^{+} \subseteq F$ and $R^{-} \cap F = \emptyset$ we have that $P \cup \toFacts{F}$ has no answer set,
					which is by Definition~\ref{def:inconsistencyReason} exactly the case iff $R$ is an IR of $P$ wrt.~$D$.
					The observation that checking if $P'$ has no answer set -- which is equivalent to deciding if $R$ is an IR --
					is in $\Pi^P_2$ resp.~$\mathit{coNP}$ for (i) resp.~(ii),
					because the property of being ordinary and disjunction-free carries over from $P$ to $P'$, concludes the proof.
			}
			
			\appendToProofs{
				Before we analyze the complexity of deciding if a program has some IR, observe:
			}

			\addLemmaOutsourcedOnly{lem:irComplexityIRExistence}{
				There is an IR of a \hex-program $P$ wrt.~$D$ iff $P \cup \toFacts{F}$ is inconsistent for some $F \subseteq D$.
			}

			\addProofOutsourcedOnly{lem:irComplexityIRExistence}{
				($\Rightarrow$) If $R$ is an IR of $P$ wrt.~$D$, then
					by Definition~\ref{def:inconsistencyReason} $P \cup \toFacts{F'}$ is inconsistent for all $F' \subseteq D$
					with $R^{+} \subseteq F'$ and $R^{-} \cap F' = \emptyset$.
					This holds in particular for $F = R^{+}$.

				($\Leftarrow$) If $P \cup \toFacts{F}$ is inconsistent for some $F \subseteq D$,
					then choosing $R = (F, D \setminus F)$ is an IR.
					This is because the only $F' \subseteq D$ with $R^{+} \subseteq F'$ and $R^{-} \cap F' = \emptyset$
					is $F$ itself, and we already know that $P \cup \toFacts{F}$ is inconsistent.
			}

			Based on this result one can derive the complexities for checking the existence of IRs.
			For all program classes, they are one level higher than checking a single candidate IR.
			Intuitively this is because the consideration of all potential IRs introduces another level of nondeterminism.

			\addProposition{prop:irComplexityIRExistence}{
				Given a \hex-program $P$ and a domain $D$.
				Deciding if there is an IR of $P$ wrt.~$D$ is
				\begin{enumerate}[(i)]
					\item $\Sigma^P_3$-complete for general \hex-programs and for ordinary disjunctive ASP-programs, and
					\item $\Sigma^P_2$-complete if $P$ is an ordinary disjunction-free program.
				\end{enumerate}
			}

			\addProof{prop:irComplexityIRExistence}{
				\noindent\textbf{Hardness:}
					We first show $\Pi^P_3$-hardness (for (i)) resp.~$\Pi^P_2$-hardness (for (ii)) of the complementary problem,
					which is checking if there is \emph{no} IR of $P$ wrt.~$D$.
					To this end, we reduce satisfiability of an appropriate QBF to our reasoning problem.
					
					(i) We reduce checking satisfiability of a QBF
					of form $\forall \vec{X} \exists \vec{Y} \forall \vec{Z} \phi(\vec{X}, \vec{Y}, \vec{Z})$
					to the check for non-existence of an IR of $P$ wrt.~$D$.
					For a fixed vector of truth values $\vec{t}$ for the variables $\vec{X}$, we construct the following program disjunctive ordinary ASP-program:
					\begin{align*}
						P(\vec{t}) = \{
						y \vee \mathit{ny} \leftarrow& \text{ for all } y \in \vec{Y} \\
						z \vee \mathit{nz} \leftarrow& \text{ for all } z \in \vec{Z} \\
						\mathit{sat}_i \leftarrow& l_{i,j} \text{ for all literals } l_{i,j} \text{ in } \phi_i \text{and all clauses } \phi_i \text{ in } \phi \\
						\mathit{sat} \leftarrow& \mathit{sat}_1, \ldots, \mathit{sat}_\ell \\
						z \leftarrow& \mathit{sat} \text{ for all } z \in \vec{Z} \\
						\mathit{nz} \leftarrow& \mathit{sat} \text{ for all } z \in \vec{Z} \\
						\leftarrow& \naf \mathit{sat} \} \\
						\cup \ \{ x_i \leftarrow \mid x_i& \in \vec{X}, t_i = \top \}
					\end{align*}
					Then $P(\vec{t})$ has an answer set iff $\exists \vec{Y} \forall \vec{Z} \phi(\vec{t}, \vec{Y}, \vec{Z})$
					is satisfiable.
					This is because the values of $x_i$ are forced to the truth value $t_i$ in any answer set of
					$P(\vec{t})$: if $t_i = \top$ then the fact $x_i \leftarrow$ forces $x_i$ to be true,
					if $t_i = \bot$ then there is no support for $x_i$. Moreover, all $y \in \vec{Y}$ are guessed nondeterministically,
					which resembles existential quantification of $\vec{Y}$.
					Finally, the program employs a saturation encoding for checking
					if for the current guess $\vec{y}$ of $\vec{Y}$ we have that $\phi(\vec{t}, \vec{y}, \vec{z})$ holds for all guesses $\vec{z}$ of $\vec{Z}$.
					The atom $\mathit{sat}$ is derived whenever $\phi(\vec{t}, \vec{y}, \vec{z})$ holds for the current guesses $\vec{y}$ of $\vec{Y}$ and $\vec{z}$ of $\vec{Z}$.
					Minimality of answer sets guarantees that the saturated interpretation, containing atom $\mathit{sat}$,
					is an answer set of $P(\vec{t})$ iff some guess for $y \in \vec{Y}$ and all subsequent guesses of values for $z \in \vec{Z}$
					satisfy $\phi(\vec{t}, \vec{Y}, \vec{Z})$, which resembles the original quantifications of $\vec{Y}$ and $\vec{Z}$,
					respectively.
					Then, $\forall \vec{X} \exists \vec{Y} \forall \vec{Z} \phi(\vec{X}, \vec{Y}, \vec{Z})$
					is satisfiable iff $P(\vec{t})$ has an answer set for all values $\vec{t}$ for $\vec{X}$.
					By Lemma~\ref{lem:irComplexityIRExistence}, this is the case iff there is no IR of $P$ wrt.~$D$,
					Because satisfiability checking of a QBF of the given form is $\Pi^P_3$-hard, the same applies to checking the non-existence of an IR of $P$ wrt.~$D$.

					(ii) We reduce checking satisfiability of a QBF
					of form $\forall \vec{X} \exists \vec{Y} \phi(\vec{X}, \vec{Y})$
					to the check for non-existence of an IR of an ordinary \emph{disjunction-free} $P$ wrt.~$D$.
					To this end, one can apply the same program as for case (i) by using the empty vector for $\vec{Z}$.
					Observe that then the program is ordinary and disjunction-free.
					Then, $\forall \vec{X} \exists \vec{Y} \phi(\vec{X}, \vec{Y})$
					is satisfiable iff $P(\vec{t})$ has an answer set for all values $\vec{t}$ for $\vec{X}$.
					By Lemma~\ref{lem:irComplexityIRExistence}, this is the case iff there is no IR of $P$ wrt.~$D$,
					Because satisfiability checking of a QBF of the given form is $\Pi^P_2$-hard, the same applies to checking the non-existence of an IR of $P$ wrt.~$D$.

					Finally, since the complementary problem of checking the non-existence of an IR is $\Pi^P_3$-hard for (i) and $\Pi^P_2$-hard for (ii),
					checking existence of such an IR is therefore $\Sigma^P_3$-hard for (i) and $\Sigma^P_2$-hard for (ii).

				\noindent\textbf{Membership:}
					Let $P$ be a program and $D$ be a domain.
					In order to decide if $P$ has an IR, one can
					guess an IR $R = (R^{+}, R^{-})$ of $P$ wrt.~$D$
					and decide in $\Pi^P_2$ for (i) resp.~in $\mathit{coNP}$ for (ii) (cf.~Proposition~\ref{prop:irComplexityIRChecking})
					that $F$ is an IR.
			}

			A related natural question concerns subset-minimal IRs.
			We recall that for each integer $i \ge 1$, the complexity class $D^P_i$ contains decision problems
			that are a conjunction of independent $\Sigma^P_i$ and $\Pi^P_i$ problems.
			More precisely, we have that $D^P_i = \{ L_1 \times L_2 \mid L_1 \in \Sigma^P_i, L_2 \in \Pi^P_i \}$.

			One can then show the following complexity results.
			Intuitively, the computational efforts comprise of checking that $R$ is an IR at all
			(giving lower bounds for the complexities as by Proposition~\ref{prop:irComplexityIRChecking}),
			and a further check that there is no smaller IR.

			\addProposition{prop:irComplexityMIRChecking}{
				Given a \hex-program $P$, a domain $D$, and a pair $R = (R^{+}, R^{-})$ of sets of atoms
				with $R^{+} \subseteq D$, $R^{-} \subseteq D$ and $R^{+} \cap R^{-} = \emptyset$.
				Deciding if $R$ is a subset-minimal IR of $P$ wrt.~$D$ is
				\begin{enumerate}[(i)]
					\item $D^P_2$-complete for general \hex-programs and for ordinary disjunctive ASP-programs, and
					\item $D^P_1$-complete if $P$ is an ordinary disjunction-free program.
				\end{enumerate}
			}

			\addProof{prop:irComplexityMIRChecking}{
				\noindent\textbf{Hardness:}
					For (i), let $S$ be a QBF-formula of form $\exists \vec{X_S} \forall \vec{Y_S} \phi_S(\vec{X_S}, \vec{Y_S})$
					and let $U$ be a QBF-formula of form $\exists \vec{X_U} \forall \vec{Y_U} \phi_U(\vec{X_U}, \vec{Y_U})$;
					we assume that $\phi_S$ and $\phi_U$ are given in clause normal form.
					We further assume w.l.o.g.~that the sets of variables occurring in $S$ and $U$ are disjoint, i.e., they are standardized apart.
					Deciding if $S$ is satisfiable and $U$ is unsatisfiable are well-known $\Sigma^P_2$- and $\Pi^P_2$-complete problems, respectively.
					Thus, deciding both together is complete for $D^P_2$.

					We reduce the problem of deciding satisfiability of $S$ and unsatisfiability of $U$
					to checking a candidate IR $R = (R^{+}, R^{-})$ for being a subset-minimal IR of a program wrt.~a domain,
					which proves hardness of the latter problem for class $D^P_2$.
					To this end, we construct the following ordinary disjunctive program:
					\begin{align}
						P = \{\ \ \ \ \ 
						x \vee \mathit{nx} \leftarrow& \text{ for all } x \in \vec{X_S} \cup \vec{X_U} \label{def:irComplexityMIRChecking:1} \\
						y \vee \mathit{ny} \leftarrow& \text{ for all } y \in \vec{Y_S} \cup \vec{Y_U} \label{def:irComplexityMIRChecking:2} \\
						\nonumber \\
						\mathit{satU}_i \leftarrow& l_{i,j} \text{ for all literals } l_{i,j} \text{ in } \phi_i \text{ and all clauses } \phi_i \text{ in } \phi_U \label{def:irComplexityMIRChecking:3} \\					
						\mathit{modelU} \leftarrow& \mathit{satU}_1, \ldots, \mathit{satU}_\ell \text{ for all clauses } \phi_1, \ldots, \phi_\ell \text{ in } \phi_U \label{def:irComplexityMIRChecking:4} \\
						y \leftarrow& \mathit{modelU} \text{ for all } y \in \vec{Y_U} \label{def:irComplexityMIRChecking:5} \\
						\mathit{ny} \leftarrow& \mathit{modelU} \text{ for all } y \in \vec{Y_U} \label{def:irComplexityMIRChecking:6} \\
						\mathit{v_1} \leftarrow& \naf \mathit{modelU}, u \label{def:irComplexityMIRChecking:7} \\
						\nonumber \\
						\mathit{satS}_i \leftarrow& l_{i,j} \text{ for all literals } l_{i,j} \text{ in } \phi_i \text{ and all clauses } \phi_i \text{ in } \phi_S \label{def:irComplexityMIRChecking:8} \\
						\mathit{modelS} \leftarrow& \mathit{satS}_1, \ldots, \mathit{satS}_\ell \text{ for all clauses } \phi_1, \ldots, \phi_\ell \text{ in } \phi_S \label{def:irComplexityMIRChecking:9} \\
						y \leftarrow& \mathit{modelS} \text{ for all } y \in \vec{Y_S} \label{def:irComplexityMIRChecking:10} \\
						\mathit{ny} \leftarrow& \mathit{modelS} \text{ for all } y \in \vec{Y_S} \label{def:irComplexityMIRChecking:11} \\
						\mathit{v_2} \leftarrow& \naf \mathit{modelS} \label{def:irComplexityMIRChecking:12} \\
						\mathit{v_2} \leftarrow& s \label{def:irComplexityMIRChecking:13} \\
						\nonumber \\
						\leftarrow& v_1, v_2 \} \label{def:irComplexityMIRChecking:14}
					\end{align}

					The encoding works as follows.
					Rules~(\ref{def:irComplexityMIRChecking:1}) guess an assignment of the existential variables,
					and rules~(\ref{def:irComplexityMIRChecking:2}) guess an assignment of the universal variables.
					For the current assignment, rules~(\ref{def:irComplexityMIRChecking:3}) and~(\ref{def:irComplexityMIRChecking:8}) check for each clause $\phi_i$ in $U$ and $S$, respectively,
					if it is currently satisfied. Based on the results for individual clauses,
					rules~(\ref{def:irComplexityMIRChecking:4}) and~(\ref{def:irComplexityMIRChecking:9}) check for the current assignment of the existential variables $\vec{x_U}$ and $\vec{x_S}$ if they are models of
					$\forall \vec{Y_U} \phi_U(\vec{x_U}, \vec{Y_U})$ and $\forall \vec{Y_S} \phi_S(\vec{x_S}, \vec{Y_S})$, respectively.

					To this end,
					rules~(\ref{def:irComplexityMIRChecking:5})--(\ref{def:irComplexityMIRChecking:6}) and~(\ref{def:irComplexityMIRChecking:10})--(\ref{def:irComplexityMIRChecking:11})
					make use of a saturation encoding (see Section~\ref{sec:prelim}) to realize the universal quantification over the variables in $\vec{Y_S}$ and $\vec{Y_U}$, respectively.
					More precisely, the saturated interpretation including atoms $\mathit{modelU}$ resp.~$\mathit{modelS}$ is derived iff all clauses in $U$ resp.~$S$
					are satisfied under \emph{all} guesses of $\vec{Y_S}$ resp.~$\vec{Y_U}$;
					if there is at least one guess which does not satisfy at least one clause, then -- for this guess -- $\mathit{modelU}$ resp.~$\mathit{modelS}$ is not derived,
					which results in a smaller interpretation than the saturated one. Hence, the saturated interpretation is part of an answer set iff the universal quantification is fulfilled.

					Rule~(\ref{def:irComplexityMIRChecking:7}) derives $v_1$ iff, for the current guess $\vec{x_U}$ of the existential variables in $U$,
					$\forall \vec{Y_U} \phi_U(\vec{x_U}, \vec{Y_U})$ is unsatisfiable \emph{and} $u$ is in the facts added to $P$.
					Rules~(\ref{def:irComplexityMIRChecking:12})--(\ref{def:irComplexityMIRChecking:13}) derive $v_2$ iff, for the current guess $\vec{x_S}$ of the existential variables in $S$,
					$\forall \vec{Y_S} \phi_S(\vec{x_S}, \vec{Y_S})$ is unsatisfiable \emph{or} $s$ is in the facts added to $P$.

					Suppose we evaluate $P \cup \toFacts{F}$ for some $F \subseteq D$. Note that, since the remaining part of the program is positive, the
					only way to become inconsistent is a derivation of $v_1$ and $v_2$ and a violation of the constraint~(\ref{def:irComplexityMIRChecking:14}).
					We show now that $R = (\{u, s\}, \emptyset)$ is a subset-minimal IR of $P$ wrt.~$D = \{ u, s \}$ iff $S$ is satisfiable and $U$ is unsatisfiable.
					To this end we consider four cases for the combinations of $S$ and $U$ being satisfiable or unsatisfiable, respectively,
					and show that in each case, $R$ behaves as stated by the proposition, i.e., it is a subset-minimal IR of $P$ or not as claimed.
					\begin{itemize}
						\item Case 1: Both $S$ and $U$ are satisfiable. \\
							Then there is a guess of the rules~(\ref{def:irComplexityMIRChecking:1}) such that $\mathit{modelU}$ is derived by the rules~(\ref{def:irComplexityMIRChecking:4}), and thus $v_1$ is not derived
							by the rule~(\ref{def:irComplexityMIRChecking:7}),
							which is independent of whether $u \in F$ or not.
							Hence, the constraint~(\ref{def:irComplexityMIRChecking:14}) is not violated and $P \cup \toFacts{F}$ is consistent.
							Thus $R$ is not an IR of $P$ wrt.~$D$.
						\item Case 2: Both $S$ and $U$ are unsatisfiable. \\
							Then for any guess of the rules~(\ref{def:irComplexityMIRChecking:1}), we have that $\mathit{modelU}$ is not derived by the rules~(\ref{def:irComplexityMIRChecking:4})
							and $\mathit{modelS}$ is not derived by the rules~(\ref{def:irComplexityMIRChecking:9}),
							thus
							rule~(\ref{def:irComplexityMIRChecking:7}) derives $\mathit{v_1}$ whenever $u$ is a fact and
							rule~(\ref{def:irComplexityMIRChecking:11})
							always derives $\mathit{v_2}$. In particular, it is \emph{not} necessary that $s \in F$ in order to derive $v_1$ and $v_2$ and violate the constraint~(\ref{def:irComplexityMIRChecking:14}).
							Thus, $R' = (\{u\}, \emptyset)$ is an IR of $P$ wrt.~$D$, which proves that $R$ is not subset-minimal.
						\item Case 3: $S$ is unsatisfiable and $U$ is satisfiable. \\
							Since $U$ is satisfiable, there is some guess of the rules~(\ref{def:irComplexityMIRChecking:1}) which leads to derivation of $\mathit{modelU}$ by rules~(\ref{def:irComplexityMIRChecking:4}).
							In this case, $v_1$ is \emph{not} derived by rule~(\ref{def:irComplexityMIRChecking:7}) and the constraint~(\ref{def:irComplexityMIRChecking:14}) is not violated, independent of $F$.
							But then in any case $R$ is not an IR of $P$ wrt.~$D$.
						\item Case 4: $S$ is satisfiable and $U$ is unsatisfiable. \\
							We have that $\mathit{modelU}$ is not derived for any guess and thus rule~(\ref{def:irComplexityMIRChecking:7}) derives $v_1$ iff $u \in F$.
							Moreover, there is a guess of the rules~(\ref{def:irComplexityMIRChecking:1})
							such that $\mathit{modelS}$ is derived by the rules~(\ref{def:irComplexityMIRChecking:9}) and thus $v_2$ is not derived the rules~(\ref{def:irComplexityMIRChecking:12}),
							hence $s \in F$ is needed to make sure that $v_2$ is dervied by rule~(\ref{def:irComplexityMIRChecking:13}).
							Thus, both $u$ and $s$ must be in $F$ in order to make $P \cup \toFacts{F}$ inconsistent,
							i.e., $P \cup \toFacts{F}$ is inconsistent iff $s, u \in R^{+}$.
							But then $R = (\{u, s\}, \emptyset)$ is a subset-minimal IR of $P$ wrt.~$D$.
					\end{itemize}

					For (ii), let $S$ be a QBF-formula of form $\exists \vec{X_S} \phi_S(\vec{X_S})$
					and let $U$ be a QBF-formula of form $\exists \vec{X_U} \phi_U(\vec{X_U})$.
					We assume again w.l.o.g.~that the sets of variables occurring in $S$ and $U$ are disjoint, i.e., they are standardized apart.
					For this type of formula, deciding if $S$ is satisfiable and $U$ is unsatisfiable are well-known \np- and \conp-complete problems, respectively.
					Thus, deciding both is complete for $D^P_1$.

					The problem is a special case of case (i) with $\vec{Y_S} = \vec{Y_U} = \emptyset$.
					In this case,
					the sets of rules~(\ref{def:irComplexityMIRChecking:5})--(\ref{def:irComplexityMIRChecking:6}) and~(\ref{def:irComplexityMIRChecking:10})--(\ref{def:irComplexityMIRChecking:11})
					in the above encoding are empty, which makes the program head-cycle free.
					Then, the guesses by the disjunctive rules~(\ref{def:irComplexityMIRChecking:1})--(\ref{def:irComplexityMIRChecking:2}) can be rewritten to unstratified negation
					and the program becomes normal.
					Hence, the original problem of deciding whether $S$ is satisfiable and $U$ is unsatisfiable can be reduced to checking a candidate IR for being a subset-minimal IR of an ordinary normal program,
					which proves hardness of the latter problem for class $D^P_1$.

				\noindent\textbf{Membership:}
					By Proposition~\ref{prop:irComplexityIRChecking},
					checking if a candidate IR $R = (R^{+}, R^{-})$ is an IR is feasible in $\Pi^P_2$ for (i) and in \conp for (ii).
					The problem is thus reducible to an UNSAT-instance of a QBF of form $\exists \vec{X} \forall \vec{Y} \phi(\vec{X}, \vec{Y})$ for (i) and $\exists \vec{X} \phi(\vec{X})$ for (ii), respectively.

					In order to check if is also subset-minimal,
					it suffices to check if no $R' = (R'^{+}, R'^{-})$ with $R'^{+} \subseteq R^{+}$ and $R'^{-} \subseteq R^{-}$ and $|(R^{+} \cup R^{-}) \setminus (R'^{+} \cup R'^{-})| = 1$
					is an IR; each such check is feasible in $\Sigma^P_2$ for (i) and \np for (ii)
					and thus reducible to a SAT-instance of a QBF of form $\exists \vec{X} \forall \vec{Y} \phi(\vec{X}, \vec{Y})$ for (i) and $\exists \vec{X} \phi(\vec{X})$ for (ii), respectively.
					Since only linearly many such checks are necessary, they can be combined into a single SAT-instance which is only polynomially larger than the original instance.

					Therefore, the problem of checking a candidate IR for being a subset-minimal IR can be reduced to a pair of independent SAT- and an UNSAT-instances
					over QBFs of form $\exists \vec{X} \forall \vec{Y} \phi(\vec{X}, \vec{Y})$ for (i) and $\exists \vec{X} \phi(\vec{X})$ for (ii), respectively,
					which proves membership results as stated.
			}

			Finally, we further consider the check for the existence of a subset-minimal IR.
			However, these results are immediate consequences of those by Proposition~\ref{prop:irComplexityIRExistence}
			as there is an IR iff there is a subset-minimal IR:

			\addProposition{prop:irComplexityIRExistenceMinimal}{
				Given a \hex-program $P$ and a domain $D$.
				Deciding if there is a subset-minimal IR of $P$ wrt.~$D$ is
				\begin{enumerate}[(i)]
					\item $\Sigma^P_3$-complete for general \hex-programs and for ordinary disjunctive ASP-programs, and
					\item $\Sigma^P_2$-complete if $P$ is an ordinary disjunction-free program.
				\end{enumerate}
			}

			\addProof{prop:irComplexityIRExistenceMinimal}{
				Each subset-minimal IR is in particular an IR, and if there is an IR $R$ then either $R$ or a smaller one is a subset-minimal IR.
				Thus, there is an IR iff there is a subset-minimal one and the results by Proposition~\ref{prop:irComplexityIRExistence} carry over.
			}

			We summarize the complexity results in Table~\ref{tab:complexity}.
			
			\begin{table}
				\centering
				\setlength{\extrarowheight}{0.3ex}
				\begin{tabular}{|l||c|c|c|c|}
					\hline
					Reasoning problem	& \multicolumn{3}{c|}{Program class} & Prop. \\
								& Normal ASP & Disjunctive ASP & General \hex & \\
					\hline
					\hline
					Checking an IR candidate & $\mathit{coNP}$-c & $\Pi^P_2$-c  & $\Pi^P_2$-c & \ref{prop:irComplexityIRChecking} \\
					Checking existence of an IR & $\Pi^P_2$-c & $\Pi^P_3$-c & $\Pi^P_3$-c & \ref{prop:irComplexityIRExistence} \\
					Checking a subset-minimal IR candidate & $D^P_1$-c & $D^P_2$-c & $D^P_2$-c & \ref{prop:irComplexityMIRChecking} \\
					Checking existence of a minimal IR & $\Pi^P_2$-c & $\Pi^P_3$-c & $\Pi^P_3$-c & \ref{prop:irComplexityIRExistenceMinimal} \\
					\hline
				\end{tabular}
				\caption{Summary of Complexity Results}
				\label{tab:complexity}
			\end{table}

	\section{Techniques for Computing Inconsistency Reasons}
	\label{sec:computing}

		In this section we discuss various methods for computing IRs.
		For ordinary normal programs we present a \emph{meta-programming} approach, extending the one from Section~\ref{sec:embedding}, which encodes the computation of IRs as a disjunctive program.
		For general \hex-programs this is not possible due to complexity reasons (except with an exponential blowup) and we present a procedural algorithm instead.

		\subsection{Inconsistency Reasons for Normal Ground ASP-Programs}

			If the program $P$ at hand is normal and does not contain external atoms, then one can construct a positive disjunctive meta-program $M^P_{\mathit{gr}}$ which
			checks consistency of $P$, as shown in Section~\ref{sec:embedding}.
			We recapitulate the properties of $M \cup M^P_{\mathit{gr}}$ as by Proposition~\ref{prop:genericRewriting} as follows:
			\begin{itemize}
				\item $M \cup M^P_{\mathit{gr}}$ is always consistent;
				\item if $P$ is inconsistent, then $M \cup M^P_{\mathit{gr}}$ has a single answer set $I_{\mathit{sat}} = A(M \cup M^P_{\mathit{gr}})$ containing all atoms in $M \cup M^P_{\mathit{gr}}$
					including a dedicated atom $\mathit{noAS}$ which does not appear in $P$; and
				\item if $P$ is consistent, then the answer sets of $M \cup M^P_{\mathit{gr}}$ correspond one-to-one to those of $P$ and none of them contains $\mathit{noAS}$. 
			\end{itemize}

			Then, the atom $\mathit{noAS}$ in the answer set(s) of $M \cup M^P_{\mathit{gr}}$ represents inconsistency of the original program $P$.
			One can then extend $M \cup M^P_{\mathit{gr}}$ in order to compute the inconsistency reasons of $P$ as follows.
			We construct
			\begin{align}
				\tau(D, P) = & \ M \cup M^P_{\mathit{gr}} \\
								\cup & \ \{ a^{+} \vee a^{-} \vee a^{x} \mid a \in D \} \ \cup \label{def:incReasonEncodingN:1} \\
								\cup & \ \{ a \leftarrow a^{+}; \ \leftarrow a, a^{-}; \ a \vee \bar{a} \leftarrow a^{x} \mid a \in D \} \label{def:incReasonEncodingN:2} \\
								\cup & \ \{ \leftarrow \naf \mathit{noAS} \}\text{,} \label{def:incReasonEncodingN:3}
			\end{align}
			where $a^{+}$, $a^{-}$, $a^{x}$ and $\bar{a}$ are new atoms for all atoms $a \in D$.

			Informally, the idea is that
			rules~(\ref{def:incReasonEncodingN:1}) guess all possible candidate IRs $R = (R^{+}, R^{-})$, where $a^{+}$ represents $a \in R^{+}$, $a^{-}$ represents $a \in R^{-}$
			and $a^{x}$ represents $a \not\in R^{+} \cup R^{-}$.
			Rules~(\ref{def:incReasonEncodingN:2}) guess all possible sets of input facts wrt.~the currently guessed IR,
			where $\bar{a}$ represents that $a$ is not a fact.
			We know from Proposition~\ref{prop:genericRewriting} that $M \cup M^P_{\mathit{gr}}$ derives $\mathit{noAS}$ iff $P$ together with the facts from rules~(\ref{def:incReasonEncodingN:2}) is inconsistent.
			This allows the constraint~(\ref{def:incReasonEncodingN:3}) to eliminate all candidate IRs,
			for which not all possible sets of input facts lead to inconsistency.
			
			One can show that the encoding is sound and complete wrt.~the computation of IRs:

			\addProposition{prop:incReasonComputationGroundN}{
				Let $P$ be an ordinary normal program and $D$ be a domain.
				Then $(R^{+}, R^{-})$ is an IR of $P$ wrt.~$D$
				iff $\tau(D, P)$ has an answer set $I \supseteq \{ a^{\sigma} \mid \sigma \in \{+, -\}, a \in R^{\sigma} \}$.
			}

			\addProof{prop:incReasonComputationGroundN}{
				The rules~(\ref{def:incReasonEncodingN:1}) guess all possible explanations $(R^{+}, R^{-})$,
				where $a^{+}$ represents $a \in R^{+}$, $a^{-}$ represents $a \in R^{-}$ and $a^{x}$ represents $a \not\in R^{+} \cup R^{-}$.
				The rules~(\ref{def:incReasonEncodingN:2}) then guess all possible sets of input facts
				$F \subseteq D$ with $R^{+} \subseteq F$ and $R^{-} \cap F = \emptyset$ wrt.~the previous guess for $(R^{+}, R^{-})$:
				if $a \in R^{+}$ then $a$ must be true, if $a \in R^{-}$ then $a$ cannot be true, and if $a \not\in R^{+} \cup R^{-}$ then $a$ can either be true or not.

				Now by the properties of $M \cup M^P_{\mathit{gr}}$ we know that $A(M \cup M^P_{\mathit{gr}})$ is an answer set of $M \cup M^P_{\mathit{gr}}$ iff $P$ is inconsistent wrt.~the current facts
				computed by rules~(\ref{def:incReasonEncodingN:2}).
				By minimality of answer sets, $A(\tau(D, P))$ is an answer set of $\tau(D, P)$.

				Rules~(\ref{def:incReasonEncodingN:3}) ensure that also the atoms $a^{+}$, $a^{-}$ and $a^{x}$ are true for all $a \in D$.
				Since $M \cup M^P_{\mathit{gr}}$ is positive, this does not harm the property of being an answer set wrt.~$M \cup M^P_{\mathit{gr}}$.
			}

			We provide a tool which allows for computing inconsistency reasons of programs, which is available from~\url{https://github.com/hexhex/inconsistencyanalysis}.
			The tool expects as command-line parameters a normal ASP-program $P$ (given as filename) and a comma-separated list of atoms to specify a domain $D$.
			Its output is another ASP-program $P'$ whose answer sets correspond to the IRs of $P$ wrt.~$D$; more precisely, each $I \in \mathcal{AS}(P)$
			contains atoms of kind $\mathit{rp}(\cdot)$ and $\mathit{rm}(\cdot)$ to specify the sets $R^{+}$ and $R^{-}$ of an IR $(R^{+}, R^{-})$, respectively.
			
		\subsection{Inconsistency Reasons for General Ground \hex-Programs}
		\label{sec:computing:ground}

			Using meta-programming approaches for computing IRs has its limitations depending on the class of the program at hand.
			Suppose the IRs of a program $P$ wrt.~a domain $D$ can be computed by a meta-program;
			then this meta-program is consistent iff an IR of $P$ wrt.~$D$ exists. Therefore, consistency checking over the meta-program must necessarily have a complexity not lower than
			the one of deciding existence of an IR of $P$.
			However, we have shown in Proposition~\ref{prop:irComplexityIRExistence} that deciding if a general \hex-program
			has an IR is $\Sigma^P_3$-complete, while consistency checking over a general \hex-program is only $\Sigma^P_2$-complete.
			But then, unless $\Sigma^P_3 = \Sigma^P_2$, computing the IRs of a general \hex-program cannot be polynomially reduced to a meta-\hex-program
			(using a non-polynomial reduction is possible but uncommon, which is why we do not follow this possibility).
			We present two possible remedies.

			\paragraph{An incomplete meta-programming approach}

				For \hex-programs $P$ without disjunctions we can specify an encoding for computing its IRs, which is sound but not complete.
				One possibility is to make use of a rewriting of $P$ to an ordinary ASP-program $\hat{P}$, which was previously used for evaluating \hex-programs.
				In a nutshell, each external atom $\ext{g}{\vec{p}}{\vec{c}}$ in $P$ is replaced by an ordinary \emph{replacement atom} $e_{\amp{g}[\vec{p}]}(\vec{c})$
				and rules $e_{\amp{g}[\vec{p}]}(\vec{c}) \leftarrow \naf \mathit{ne}_{\amp{g}[\vec{p}]}(\vec{c})$ and $\mathit{ne}_{\amp{g}[\vec{p}]}(\vec{c}) \leftarrow \naf e_{\amp{g}[\vec{p}]}(\vec{c})$
				are added to guess the truth value of the former external atom.
				However, the answer sets of the resulting \emph{guessing program} $\hat{P}$
				do not necessarily correspond to answer sets of the original program $P$.
				Instead, for each answer set it must be checked if the guesses are correct.
				An answer set $I$ of the guessing program $\hat{P}$
				is called a \emph{compatible set} of $P$, if $\extsem{g}{\hat{I}}{\vec{p}}{\vec{c}} = \T$ iff
				$e_{\amp{g}[\vec{p}]}(\vec{c}) \in \hat{I}$ for
				all external atoms $\amp{g}[\vec{p}](\vec{c})$ in $P$.
				Each answer set of $P$ is the projection of some compatible set of $P$.

				One can exploit the rewriting $\hat{P}$ to compute (some) IRs of $P$.
				To this end, one constructs $\tau(D, \hat{P})$ and computes its answer sets.
				This yields explanations for the inconsistency of the guessing program $\hat{P}$ rather than the actual \hex-program $P$.
				The \hex-program $P$ is inconsistent whenever the guessing program $\hat{P}$ is, and every inconsistency reason for $\hat{P}$ is also one for $P$.
				Hence, the approach is sound, but
				since $\hat{P}$ might be consistent even if $P$ is inconsistent, it follows that the approach is not complete:
			
				\addProposition{prop:soundnessNoCompleteness}{
					Let $P$ be a \hex-program and $D$ be a domain.
					Then each IR of $\hat{P}$ wrt.~$D$ is also an IR of $P$ wrt.~$D$, i.e., the use of $\hat{P}$ is sound wrt.~the computation of IRs.
				}
			
				\addProof{prop:soundnessNoCompleteness}{
					As each answer set of $P$ is the projection of a compatible set of $\hat{P}$ to the atoms in $P$,
					inconsistency of $\hat{P} \cup \toFacts{F}$ implies inconsistency of $\hat{P} \cup \toFacts{F}$ for all $F \subseteq D$.
				}

				The following example shows that using $\hat{P}$ for computing IRs is not complete.
				
				\begin{example}
					Consider the \hex-program $P = \{ p \leftarrow q, \ext{\mathit{neg}}{p}{} \}$ and domain $D = \{ q \}$.
					An IR is $(\{ q \}, \emptyset)$ because $P \cup \toFacts{F}$ for $F = \{ q \}$ is inconsistent.
					However, the guessing program $\hat{P}$ extended with $F$
					\begin{align*}
						\hat{P} \cup \{ q \leftarrow \} =& \{ e_{\amp{\mathit{neg}}[p]} \leftarrow \naf \mathit{ne}_{\amp{\mathit{neg}}[p]} \\
														& \phantom{\{} \mathit{ne}_{\amp{\mathit{neg}}[p]} \leftarrow \naf e_{\amp{\mathit{neg}}[p]} \\
														& \phantom{\{} p \leftarrow q, e_{\amp{\mathit{neg}}[p]} \\
														& \phantom{\{} q \leftarrow \}
					\end{align*}
					is consistent and has the answer set $\hat{I} = \{ e_{\amp{\mathit{neg}}[p]}, p, q \}$; therefore $(\{ q \}, \emptyset)$ is not found as an IR when using $\hat{P}$ (actually there is no IR of $\hat{P}$ wrt.~$D$).
				\end{example}
				
				In the previous example, the reason why no inconsistency reason is found is that $\hat{I}$ is an answer set of $\hat{P}$ but the value of
				$e_{\amp{\mathit{neg}}[p]}$ guessed for the external replacement atom
				differs from the actual value of $\ext{\mathit{neg}}{p}{}$ under $\hat{I}$, i.e., $\hat{I}$ is not compatible.

			\paragraph{An incomplete procedural algorithm}

				Beginning from this section, we need to distinguish unassigned and false atoms
				and denote assignments as sets of signed literals, as discussed in Section~\ref{sec:prelim}.
				
				Our algorithm is an extension of the
				state-of-the-art evaluation algorithm for ground \hex-programs.
				The evaluation algorithm is in turn based on conflict-driven nogood learning (CDNL),
				which has its origins in SAT solving (cf.~e.g.~\citeN{MSLM09HBSAT}).
				Here, a \emph{nogood} is a set $\{ L_1, \dotsc, L_n \}$ of ground literals
				$L_i, 1 \le i \le n$.
				An assignment $\Assignment$ is a \emph{solution} to a nogood $\delta$
				if $\delta \not\subseteq \Assignment$,
				and to a set of nogoods $\Delta$ if $\delta \not\subseteq \Assignment$ for all $\delta \in \Delta$.
				Note that according to this definition, partial assignments (i.e., assignments such that for some $a \in A$ we have $\T a \not\in A$ and $\F a \not\in A$)
				might be solutions to nogoods, even if supersets thereof are not; this definition is by intend and does not harm in the following.
				The basic idea is to represent the program at hand as a set of nogoods and try to construct an assignment, which satisfies all nogoods,
				where both deterministic propagation and guessing is used. The distinguishing feature of CDNL is that the nogood set is \emph{dynamically extended}
				with new nogoods, which are learned from conflict situations, such that the same conflict is not constructed again.

				The CDNL-based approach for \hex-programs is shown in Algorithm~\ref{alg:hexcdnl}.
				The input is a program $\Program$, a set of input facts $F$,
				and an \emph{inconsistency handler function} $h$; the latter is a preparation for our extensions below and returns always $\bot$ for answer set computation.
				The output is an answer set of $\Program \cup \toFacts{F}$ if there is one, and $\bot$ otherwise.

				\begin{algorithm}[t]
					\caption{\CDNLHEX}
					\label{alg:hexcdnl}
					\DontPrintSemicolon

					\KwIn{ground program $\Program$, input atoms $F$, an inconsistency handler function $h$ (with a nogood set and an assignment as parameters)}
					\KwOut{an answer set of $\Program \cup \toFacts{F}$ if existing, and $h(\Delta, \AssignmentP)$ otherwise (where $\Delta$ resp.~$\AssignmentP$ are the nogood set resp.~assignment at the time of inconsistency discovery)}
					\smallskip

					\tcp{Initialization}
					\nlset{(a)}{%
					\label{alg:hexcdnl:a}%
					$\Program \leftarrow \Program \cup \toFacts{F}$; $\Delta \leftarrow \Delta_{\ProgramP}; \AssignmentP \leftarrow \emptyset; \mathit{current\_dl} \leftarrow 0$\;
					\For{all facts $a \leftarrow \in \Program$}{
						$\AssignmentP \leftarrow \AssignmentP \circ (\T a)$\;
						$\decisionlevel[a] \leftarrow \mathit{current\_dl}$\;
					}
					\For{$a \not\in H(r) \text{ for all } r \in \Program$}{
						$\AssignmentP \leftarrow \AssignmentP \circ (\F a)$\;
						$\decisionlevel[a] \leftarrow \mathit{current\_dl}$\;
					}
					}%

					\medskip
					\tcp{Main loop}
					\While{true}{
						\nlset{(b)}{%
						\label{alg:hexcdnl:b}%
						$(\AssignmentP, \Delta) \leftarrow \text{\Propagation}(\ProgramP, \Delta, \AssignmentP)$\;
						}
						\nlset{(c)}{%
						\label{alg:hexcdnl:c}%
						\uIf{for some nogood $\delta \in \Delta$ we have by $\delta \subseteq \AssignmentP$}{
							\lIf{$\mathit{current\_dl} = 0$}{
								\Return{$h(\Delta, \AssignmentP)$}
							}
							Analyze conflict, add learned nogood to $\Delta$, backjump and update $\mathit{current\_dl}$\;%
						}
						}
						\nlset{(d)}{%
						\label{alg:hexcdnl:d}%
						\uElseIf{$\AssignmentP$ is complete}{
							\uIf{guesses are not correct or not minimal wrt.~the reduct}{
								$\Delta \leftarrow \Delta \cup \{ \AssignmentP \}$\;
								Analyze conflict, add learned nogood to $\Delta$, backjump and update $\mathit{current\_dl}$\;%
							}\Else{
								\Return{$\AssignmentP|_{A(\Program)}$}\;
							}
						}
						}
						\nlset{(e)}{%
						\label{alg:hexcdnl:e}%
						\uElseIf{\text{Heuristics evaluates } $\amp{g}[\vec{y}]$ and $\Lambda(\amp{g}[\vec{y}], \AssignmentP) \not\subseteq \Delta$}{
							$\Delta \leftarrow \Delta \cup \Lambda(\amp{g}[\vec{y}], \AssignmentP)$\;
						}
						}
						\nlset{(f)}{%
						\label{alg:hexcdnl:f}%
						\Else{ Let $\sigma \in \{ \T, \F \}$ and $a \in A(\ProgramP)$ with $\{ \T a, \F a \} \cap \AssignmentP = \emptyset$\;
							$\mathit{current\_dl} \leftarrow \mathit{current\_dl} + 1$\;
							$\decisionlevel[a] \leftarrow \mathit{current\_dl}$\;
							$\AssignmentP \leftarrow \AssignmentP \circ (\sigma a)$\;
						}
						}
					}
				\end{algorithm}
				
				The initialization is done at \ref{alg:hexcdnl:a}.
				The given \hex-program $\Program$ is extended with facts $\toFacts{F}$
				and transformed to an ordinary ASP-program $\ProgramP$ by replacing each external atom
				$\ext{g}{\vec{y}}{\vec{x}}$ in~$\Program$ by an ordinary \emph{replacement atom} $e_{\amp{g}[\vec{y}]}(\vec{x})$
				and adding a rule~$e_{\amp{g}[\vec{y}]}(\vec{x}) \vee \mathit{ne}_{\amp{g}[\vec{y}]}(\vec{x}) \leftarrow$ to guess its value.
				Program $\ProgramP$ is then represented as set of nogoods $\Delta$, which
				consists of nogoods $\Delta_{\ProgramP}$
				stemming from Clark's completion~\cite{c1977} and singleton loop nogoods~\cite{gks2012-aij};
				the former basically encode the rules as implications, while the latter control support of atoms.
				The nogood set will be expanded as the search space is traversed.

				As further initializations, we set the computed answer set $\AssignmentP$ of the guessing program $\ProgramP$ initially to the empty list;
				for the sake of the algorithm, assignments are seen as lists such that the order of assignments is retrievable later.
				The \emph{current decision level (dl)} is initially $0$ and incremented for every guess.
				The initialization further assigns facts immediately to true and atoms which do not appear in any heads to false (both at dl $0$),
				where the decision levels of atoms are stored in array $\mathit{dl}$ such that $\decisionlevel[a] \in \mathbb{N}_0$ for all atoms $a$.

				After the initialization, the algorithm performs deterministic propagation such as unit propagation at \ref{alg:hexcdnl:b}.
				Further propagation techniques such as unfounded set propagation can be added.
				Each implied literal is assigned at the maximum current decision of the literals which implied it.
				
				The algorithm detects conflicts, learns additional nogoods and backtracks at \ref{alg:hexcdnl:c}.
				If the conflict is on decision level $0$,
				the instance is inconsistent and the callback function $h$ is notified (with nogoods $\Delta$ and assignment $\AssignmentP$ over $\Program$ as input)
				to determine the return value in case of inconsistency; as said above, this callback serves as a preparation for our extension of the algorithm below
				and is instantiated with just $h_{\bot}(\Delta, \AssignmentP) = \bot$ (independent of $\Delta$ and $\AssignmentP$) for answer set computation.
				
				If the assignment is complete at \ref{alg:hexcdnl:d}, the algorithm must check if the guesses of replacement atoms coincide with the
				real truth values of external atoms,
				and if it represents a model which is also subset-minimal wrt.~the reduct.
				If this is not the case, the assignment is added as nogood in order to discard it; actual implementations are more involved to keep memory consumption small~\cite{Drescher08conflict-drivendisjunctive}.
				
				Next, the algorithm may perform theory propagation at \ref{alg:hexcdnl:e} driven by a heuristics.
				That is, an external source $\amp{g}$ with input $\vec{y}$ may be evaluated under assignment $\AssignmentP$
				to learn parts of its behavior in form of a set of nogoods. We abstractly use a so-called \emph{learning function} $\Lambda$
				to refer to the set of nogoods $\Lambda(\amp{g}[\vec{y}], \AssignmentP)$ learned from such an evaluation step.

				Finally, if none of the other cases applies, we have to guess a truth value at \ref{alg:hexcdnl:f}.
				
				\smallskip
				\noindent We refer to \citeN{efkrs2014-jair} for a more extensive study of this evaluation approach, who also
				proved soundness and completeness of Algorithm~\ref{alg:hexcdnl}:
				
				\addProposition{prop:correctnessHexCDNL}{
					Let $\Program$ be a program and $F \subseteq D$ be input atoms from a domain $D$.
					If we have that $\text{\CDNLHEX}(\Program, F, h_{\bot})$ returns
					\begin{inparaenum}[(i)]
						\item\label{prop:correctnessHexCDNL:correctness} an assignment $\Assignment$, then $\Assignment$ is an answer set of $\Program \cup \toFacts{F}$;
						\item\label{prop:correctnessHexCDNL:completeness} $\bot$, then $\Program \cup \toFacts{F}$ is inconsistent.
					\end{inparaenum}
				}
				
				\addProof{prop:correctnessHexCDNL}{
					See~\citeN{efkrs2014-jair}.
				}

				Our approach for computing IRs for ground programs is based on \emph{implication graphs},
				which are used for conceptual representation of the current status and assignment history of the solver,
				cf.~e.g.~\citeN{HandbookOfSAT2009}.
				However, in Algorithm~\ref{alg:hexcdnl} the implication graph is only implicitly represented by the order of assignments.

				Intuitively, the nodes of an implication graph represent (already assigned) literals or conflicts, their decision levels,
				and the nogoods which implied them. Predecessor nodes represent implicants. Nodes without predecessors represent guesses.
				Formally:
				
				\begin{definition}
					An \emph{implication graph} is a directed graph $\langle V, E \rangle$,
					where $V$ is a set of triplets $\langle L, \mathit{dl}, \delta \rangle$, denoted $L@\mathit{dl}/\delta$, where
					$L$ is a signed literal or $\bot$, $\mathit{dl} \in \mathbb{N}_0$ is a decision level,
					$\delta \in \Delta \cup \{ \bot \}$ is a nogood, and $E$ is a set of unlabeled edges.
				\end{definition}
				
\begin{figure}[h]
					\centering
					\footnotesize
					\beginpgfgraphicnamed{implicationgraph}
					\begin{tikzpicture}[->,>=stealth',shorten >=1pt,auto,node distance=1cm and 1.2cm,inner sep=1pt,outer sep=0pt,semithick]
							\node         (a)                    {$\T a@1/\bot$};
							\node         (b) [below right=of a] {$\F b@1/\delta_1$};
							\node         (c) [above right=of b] {$\T c@1/\delta_2$};
							\node         (d) [below right=of c] {$\T d@2/\bot$};
							\node         (e) [above right=of d] {$\T e@2/\delta_3$};
							\node         (f) [right=of e]       {$\bot@2/\delta_4$};

							\path (a) edge [->] node {} (b)
								  (a) edge [->] node {} (c)
								  (b) edge [->] node {} (c)
								  (c) edge [->] node {} (e)
								  (d) edge [->] node {} (e)
								  (d) edge [->] node {} (f)
								  (e) edge [->] node {} (f);
					\end{tikzpicture}
					\endpgfgraphicnamed

					\caption{Implication graph of the program in Example~\ref{ex:implicationgraph}}
					\label{fig:implicationgraph}
				\end{figure}
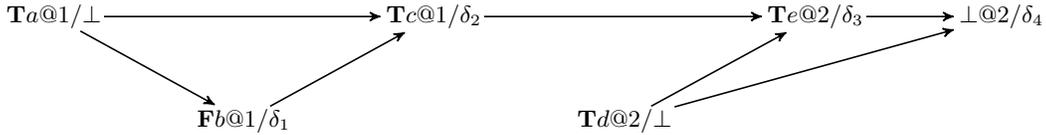

				\begin{example}
					\label{ex:implicationgraph}
					Let \\
					$\Delta = \big\{ \delta_1\colon \{ \T a, \T b \}, \delta_2\colon \{ \T a, \F b, \F c \}, \delta_3\colon \{ \T c, \T d, \F e \}, \delta_4\colon \{ \T d, \T e \} \big\}$.
					An implication graph is shown in Figure~\ref{fig:implicationgraph}.
					Here, $\T a@1/\bot$ is a guess at decision level $1$,
					$\F b@1/\delta_1$ is implied by $\T a$ using $\delta_1$,
					$\T c@1/\delta_2$ is implied by $\T a$ and $\F b$ using $\delta_2$,
					$\T d@2/\bot$ is a guess at decision level $2$,
					and $\T e@2/\delta_3$ is implied by $\T c$ and $\T d$ using $\delta_3$.
					Then $\delta_4$ is violated due to $\T d$ and $\T e$.
				\end{example}

				In order to compute an IR wrt.~a domain $D$ in case of inconsistency,
				we reuse Algorithm~\ref{alg:hexcdnl} but let it call Algorithm~\ref{alg:incReason} after an inconsistency is detected
				by using $\text{InconsistencyAnalysis}(D, \Delta, \AssignmentP)$ as inconsistency handler function instead of $h_{\bot}$.

				Observe that for answer set computation (i.e., using $h = h_{\bot}$),
				soundness of Algorithm~\ref{alg:hexcdnl} implies that $h(\Delta, \AssignmentP)$ is called at some point
				because it is the only way to return $\bot$.
				Note that this can only happen after a violated nogood has been identified at decision level $0$.
				But this implies that at the time the inconsistency of an instance is detected, all assignments are deterministic.
				
				\begin{algorithm}[t]
					\caption{InconsistencyAnalysis}
					\label{alg:incReason}
					\DontPrintSemicolon

					\KwIn{domain $D$, set of nogoods $\Delta$ representing program $\Program$, assignment $\AssignmentP$ conflicting with $\Delta$}
					\KwOut{inconsistency reason of $\Program$ wrt.~$D$}
					\smallskip

					\nlset{(a)}{%
					\label{alg:incReason:a}
					Let $\delta \in \Delta$ such that $\delta \subseteq \AssignmentP$\;
					}
					\nlset{(b)}{%
					\label{alg:incReason:b}
					\While{there is a $\sigma a \in \delta$ with $a \not\in D$}{
						Let $\epsilon \in \Delta$ s.t.~$\overline{\sigma a} \in \epsilon$ and $\epsilon \setminus \overline{\sigma a} \subseteq \AssignmentP'$ for some $\AssignmentP = \AssignmentP' \circ (\sigma a) \circ \AssignmentP$\; 
						\nlset{(c)}{%
						\label{alg:incReason:c}
						$\delta \leftarrow (\delta \setminus \{ \sigma a \}) \cup (\epsilon \cup \overline{\sigma a})$\;
						}
						$\AssignmentP \leftarrow \AssignmentP'$\;
					}
					}
					\nlset{(d)}{%
					\label{alg:incReason:d}
					\Return{$(\{ a \mid \T a \in \delta \}, \{ a \mid \F a \in \delta \})$}\;
					}
				\end{algorithm}

				Now the basic idea of Algorithm~\ref{alg:incReason} is as follows.
				We start at~\ref{alg:incReason:a} with a nogood $\delta \in \Delta$ which is currently violated; such a nogood always exists because otherwise Algorithm~\ref{alg:hexcdnl} would not have called $h(\Delta, \AssignmentP)$.
				Since each literal $\sigma a$ in this nogood was assigned at decision level $0$, each of them was assigned by unit propagation as a consequence of previously assigned literals.
				We then iteratively (cf.~loop at~\ref{alg:incReason:b}) resolve this nogood
				with the cause of one of its literals $\sigma a$ at~\ref{alg:incReason:c}, i.e., with the nogood $\epsilon$ which implied literal $\sigma a$.
				The loop at~\ref{alg:incReason:b} repeats this step until the nogood contains only literals over explanation atoms from $D$;
				since these literals imply the originally violated nogood and are thus responsible for inconsistency,
				$\delta$ represents an IR which we return at~\ref{alg:incReason:d}.
				The iterative resolution corresponds to the replacement of the literal by its predecessors in the implication graph
				(where the implication graph itself is implicitly represented by the implicants of literals).
				
				For a given set $D$, using $h_{\mathit{analyse}}^D(\Delta, \AssignmentP) = \text{InconsistencyAnalysis}(D, \Delta, \AssignmentP)$
				allows then for exploiting Algorithm~\ref{alg:hexcdnl} to compute IRs:

				\addProposition{prop:algGetIncReason}{
					Let $\Program$ be a program and $F \subseteq D$ be input atoms from a domain $D$.
					If we have that $\text{\CDNLHEX}(\Program, F, h_{\mathit{analyse}}^D)$ returns
					\begin{inparaenum}[(i)]
						\item\label{prop:algGetIncReason:correctness} an assignment $\Assignment$, then $\Assignment$ is an answer set of $\Program \cup \toFacts{F}$;
						\item\label{prop:algGetIncReason:completeness} a pair $R = (R^{+}, R^{-})$ of sets of atoms, then $\Program \cup \toFacts{F}$ is inconsistent and $R$ is an inconsistency reason of $\Program$ wrt.~$D$.
					\end{inparaenum}
				}
				
				\addProof{prop:algGetIncReason}{
					Case~(\ref{prop:algGetIncReason:correctness}) follows from case~(\ref{prop:correctnessHexCDNL:correctness}) of Proposition~\ref{prop:correctnessHexCDNL}.
					
					For case~(\ref{prop:algGetIncReason:completeness}), it suffices to focus on Algorithm~\ref{alg:incReason}
					and show that $\text{InconsistencyAnalysis}(D, \Delta, \AssignmentP)$ returns an IR of $\Program$ wrt.~domain $D$,
					where $\Delta$ is the set of nogoods from Algorithm~\ref{alg:hexcdnl} after ending up at
					with an assignment $\AssignmentP$ which violates $\Delta$ at decision level $0$.

					The main idea of the proof is then follows. We show that if we currently evaluate the program with input $F \subseteq D$ and Algorithm~\ref{alg:incReason} returns $R = (R^{+}, R^{-})$,
					then, whenever $\text{\CDNLHEX}(\Program, J, h_{\mathit{analyse}}^D)$ is called for some $J \subseteq D$ with $J \subseteq R^{+}$ and $J \cap R^{-} = \emptyset$,
					the nogood $\delta_0$ initially violated during inconsistency analysis (i.e., the value of $\delta$ before entering the loop in Algorithm~\ref{alg:incReason})
					can also be added during evaluation of $\Program \cup \toFacts{J}$.
					This means that the same situation which has led to the detection of inconsistency can be reconstructed,
					which implies that also $\Program \cup \toFacts{J}$ is inconsistent as required.

					To this end, first observe that the initial nogood set $\Delta_{\widehat{\Program \cup \toFacts{F}}}$ over the input $F \subseteq D$ used for inconsistency analysis,
					and the initial nogood set $\Delta_{\widehat{\Program \cup \toFacts{J}}}$ for a (possibly) different input $J \subseteq D$ can differ only in unary nogoods over atoms in $D$.
					This is because $\Delta_{\widehat{\Program \cup \toFacts{F}}} = \Delta_{\ProgramP} \cup \Delta_{\toFacts{F}}$ and $\Delta_{\widehat{\Program \cup J}} = \Delta_{\ProgramP} \cup \Delta_{\toFacts{J}}$
					due to the fact that Clark's completion is created rule-wise~\cite{c1977},
					and that the singleton loop nogoods for an atom $a$ depend only on rules $r$ with $a \in H(r)$~\cite{gks2012-aij}; since $D$ and heads of $\Program$ do not share any atoms by Definition~\ref{def:inconsistencyReason}, their singleton loop nogoods are independent.
					In particular, all nogoods in $\Delta_{\ProgramP}$ are independent of $F$ resp.~$J$,
					and $\Delta_{\toFacts{F}}$ resp.~$\Delta_{\toFacts{J}}$ contain a nogood $\{ \F a \}$ for each atom atom $a \in D$ which also occurs in $F$ resp.~$J$, and a nogood $\{ \T a \}$ for each atom $a \in D$ which does not occur in $F$ resp.~$J$.

					Next, note that all nogoods $N$ learned at a later point contain a literal $\T a$ resp.~$\F a$ for each $a \in D$ whose presence resp.~absence in $F$ was a prerequesite for the nogood to be added
					(in the sense that for $J \subseteq D$, if $a \in J$ for each $\T a \in N$ and $a \not\in J$ for each $\F a \in N$, then $N$ can also be learned during evaluation of $\Program \cup \toFacts{J}$ without eliminating answer sets).
					This is shown by induction on the number $n$ of resolution steps performed to derive $N$.
						For the base case $n = 0$, nogood $N$ must have been added either
							(1) due to a violation of the compatibility or minimality criterion at~\ref{alg:hexcdnl:d},
							or (2) by theory propagation at~\ref{alg:hexcdnl:e}.
							In case (1) it contains a literal over each $a \in D$ according to its value in $F$,
							in case (2) it is independent of $F$ and thus can also be learned when the program is evaluated under a (possibly) different input $J$.
						For $n \rightarrow n + 1$, nogood $N$ is the resolvent of two other nogoods $N_1$ and $N_2$; these \emph{cannot} be nogoods from $\Delta_{\toFacts{F}}$ because all in $\Delta_{\toFacts{F}}$
						are unary over $D$ but literals over $D$ are never resolved due to assignment at decision level $0$.
							But then the claim holds for $N_1$ and $N_2$ either because they come from $\Delta_{\ProgramP}$ (which is independent of $F$), or because they have been added at at later point, in which case the claim holds by induction hypothesis.
							Moreover, atoms $D$ are never resolved during conflict analysis because they are assigned at decision level $0$ at~\ref{alg:hexcdnl:a},
							thus all literals over $D$ contained in $N_1$ and $N_2$ are still contained in $N$. But then the claim holds also for nogood $N$ derived by $n + 1$ resolution steps.

					But then, the initially violated nogood $\delta_0$ selected by Algorithm~\ref{alg:incReason}
					also contains a literal $\T a$ resp.~$\F a$ for each $a \in D$ whose presence resp.~absence in $F$ was a prerequesite for this nogood to be added.
					Akin to resolution during conflict analysis, the loop in Algorithm~\ref{alg:incReason} does never resolve such literals, hence the final nogood $\delta_1$ (i.e., the value of $\delta$ after the loop)
					still contains all literals over atoms $D$
					whose presence resp.~absence in $F$ was a prerequisite for the conflict to be derived.
					Using the sets $R^{+} = \{ a \mid \T a \in \delta_1 \}$ resp.~$R^{-} = \{ a \mid \F a \in \delta_1 \}$
					for the IR exactly guarantees that atoms, whose presence resp.~absence in $F$ was a prerequisite to derive the conflict nogood and all its transitive parent nogoods in the resolution,
					is still given if the program is evaluated under a (possibly) different input $J \subseteq D$, provided that $J \subseteq R^{+}$ and $J \cap R^{-} = \emptyset$.
					But then, for such a $J$, the same conflict can be reproduced during evaluation of $\Program \cup \toFacts{J}$, hence $\Program \cup \toFacts{J}$ is inconsistent.
					This is exactly what $R = (R^{+}, R^{-})$ being an IR means.
				}

		\subsection{Computing Inconsistency Reasons for General Non-Ground \hex-Programs}
		\label{sec:computing:nonground:incompleteprocedural}

			Next, we extend the above approach to non-ground programs.
			Grounders fo ASP and \hex{}-programs instantiate non-ground programs in such a way
			that the resulting ground program is still equivalent to the original one wrt.~answer set computation.
			However, in general the equivalence is not given any more wrt.~computation of inconsistency reasons.
		
			Grounding algorithms typically do not
			use the naive grounding $\mathit{grnd}_{\mathcal{C}}(\Program)$ which
			substitutes variables $\mathcal{V}$ in $\Program$ by constants from $\mathcal{C}$ in all possible ways (and is even infinite in general).
			Instead, optimizations are performed to keep the grounding small.
			However, the exact algorithms for performing such optimizations, and therefore the output, depend on the grounder in use.
			In fact, there is large room for correct grounding procedures as they can output
			any \emph{\underline{o}ptimized \underline{g}round program} $\mathit{og}_{\mathcal{C}}(\Program)$
			as long as $\mathcal{AS}(\mathit{og}_{\mathcal{C}}(\Program)) = \mathcal{AS}(\Program)$ holds;
			it does not even be a subset of $\mathit{grnd}_{\mathcal{C}}(\Program)$.
			This allows grounders also to optimize \emph{within} rules (and not just drop irrelevant rules as a whole) and introduce auxiliary rules,
			which is also done in practice.
			Due to these grounder-specific optimizations we intentionally do not introduce a fixed definition of an optimized grounding here.
			However, we assume in the following
			that an arbitrary but fixed optimized grounding of a program $\Program$ wrt.~a set of constants $\mathcal{C}$
			accessible via $\mathit{og}_{\mathcal{C}}(\Program)$.

			These optimizations prohibit the direct reduction of the computation of an IR for a (possibly) non-ground program $\Program$ wrt.~a domain $D$
			to the ground case because the grounding step may have optimized program parts away, which can be relevant when the input facts change.
			To demonstrate this,
			suppose we compute the optimized grounding $\Program_g = \mathit{og}_{\mathcal{C}}(\Program \cup \toFacts{F})$
			of a program $\Program$ with input facts $\toFacts{F}$ for some $F \subseteq D$.
			Then we may try to reuse Algorithm~\ref{alg:hexcdnl} for IR computation by passing $\Program_g \setminus \toFacts{F}$ as program and $F$ as input,
			i.e., we call $\text{\CDNLHEX}(\Program_g \setminus \toFacts{F}, F, h_{\mathit{analyse}}^D)$.
			If this call returns a pair $R = (R^{+}, R^{-})$,
			we have then by Proposition~\ref{prop:algGetIncReason} that
			$R$ is an IR for $\Program_g \setminus \toFacts{F}$, i.e., $(\Program_g \setminus \toFacts{F}) \cup \toFacts{J}$
			is inconsistent for all $J \subseteq D$ with $R^{+} \subseteq J$ and $R^{-} \cap J = \emptyset$.
			However, $R$ is not necessarily an IR for $\Program$ wrt.~$D$ because for a different set of facts $F' \subseteq D$
			we may have $\mathit{og}_{\mathcal{C}}(\Program \cup \toFacts{F'}) \setminus \toFacts{F'} \not= \mathit{og}_{\mathcal{C}}(\Program \cup \toFacts{F}) \setminus \toFacts{F}$.

			\begin{example}
				\label{ex:ogrounding}
				Consider $\Program = \{ q(X) \leftarrow p(X); \ \leftarrow \naf q(1); \ \leftarrow a \}$ and $D = \{ a, p(1) \}$.
				For input $F = \emptyset$,
				we have $\Program_g = \mathit{og}_{\mathcal{C}}(\Program \cup \toFacts{F}) \setminus \toFacts{F} = \{ \leftarrow \naf q(1); \ \leftarrow a \}$.
				Inconsistency analysis of $\Program_g$ wrt.~$F$ may yield $R = (\emptyset, \emptyset)$, which is an IR of $\Program_g$ wrt.~$D$,
				but not of $\Program$ wrt~$D$ as $\Program \cup \toFacts{\{ p(1) \}}$ is consistent.
			\end{example}

			\appendToProofs{
				\paragraph{Preparation for Proposition~\ref{prop:irNonground}}
				As a preparation for the upcoming proofs,
				we observe that one possible solution for computing IRs of a non-ground program is to use a sufficiently large but finite grounding,
				which is suitable for answer set computation wrt.~any set of input atoms from $D$.
				More precisely, for a program $\Program$ and a domain $D$,
				let $\mathit{eg}_{\mathcal{C}, D}(\Program)$ be
				an \emph{\underline{e}xhaustive \underline{g}rounding} $\Program$
				with the properties that
				(i) it is finite,
				(ii) $\mathit{eg}_{\mathcal{C}, D}(\Program) \subseteq \mathit{grnd}_{\mathcal{C}}(\Program)$,
				and
				(iii) $\mathcal{AS}(\mathit{eg}_{\mathcal{C}, D}(\Program) \cup \toFacts{F}) = \mathcal{AS}(\Program \cup \toFacts{F})$ for all $F \subseteq D$;
				safety conditions guarantee that such a grounding exists.
				We assume w.l.o.g.~that $\mathit{eg}_{\mathcal{C}, D}(\Program)$ is minimal, i.e., no proper subset has these properties.

				\begin{example}[cont'd]
					\label{ex:pogrounding}
					For the program $\Program = \{ q(X) \leftarrow p(X); \ \leftarrow \naf q(1); \ \leftarrow a \}$ and domain $D = \{ a, p(1) \}$ from Example~\ref{ex:ogrounding},
					consider the ground program $\mathit{eg}_{\mathcal{C}, D}(\Program) = \{ q(1) \leftarrow p(1); \ \leftarrow \naf q(1); \ \leftarrow a \}$;
					it is correct for answer set computations over $\mathit{eg}_{\mathcal{C}, D}(\Program) \cup \toFacts{F}$ for all $F \subseteq D$.
				\end{example}

				One can then show:
			}
			
			\addLemmaOutsourcedOnly{lem:exhaustiveGrounding}{
				For a program $\Program$ and a domain $D$,
				an IR for $\mathit{eg}_{\mathcal{C}, D}(\Program)$ wrt.~$D$ is also an IR of $\Program$ wrt.~$D$.
			}
			
			\addProof{lem:exhaustiveGrounding}{
				If $R = (R^{+}, R^{-})$ is an IR of $\mathit{eg}_{\mathcal{C}, D}(\Program)$,
				then $\mathit{eg}_{\mathcal{C}, D}(\Program) \cup \toFacts{F}$ is inconsistent for all $F \subseteq D$ with $R^{+} \subseteq F$ and $R^{-} \cap F = \emptyset$.
				But $\mathcal{AS}(\mathit{eg}_{\mathcal{C}, D}(\Program) \cup \toFacts{F}) = \mathcal{AS}(\Program \cup \toFacts{F})$ for all $F \subseteq D$,
				hence also $\Program \cup \toFacts{F}$ is inconsistent for all $F \subseteq D$.
			}

			\appendToProofs{
				However, since this grounding has to be correct for all possible inputs $F \subseteq D$, can be large
				and -- even worse -- grounding external atoms for a yet unknown input is expensive because an exponential number of
				evaluations is necessary to ensure that all relevant constants are respected in the grounding~\cite{efkr2016-aij}.
			}

			We thus use a \emph{\underline{p}artially \underline{o}ptimized \underline{g}rounding}
			$\mathit{pog}_{\mathcal{C}, F}(\Program)$ for a specific input $F \subseteq D$
			with the properties that
			(i) $\mathit{pog}_{\mathcal{C}, F}(\Program) \subseteq \mathit{grnd}_{\mathcal{C}}(\Program)$
			and (ii) $\mathcal{AS}(\mathit{pog}_{\mathcal{C}, F}(\Program)) = \mathcal{AS}(\Program \cup \toFacts{F})$.
			That is, 
			optimization is restricted to the elimination of rules, while changes within or additions of rules are prohibited. 
			A grounding procedure for \hex-programs with these properties was presented by~\citeN{efkr2016-aij}.


			\appendToProofs{
				Our computation of IRs of a (possibly non-ground) program $\Program$ wrt.~a domain $D$ and a given input $F \subseteq D$
				is now based on the grounding $\mathit{pog}_{\mathcal{C}, F}(\Program)$.
				This has the advantage that the grounding must be computed only for the specific input $F$,
				while the property that $\mathit{pog}_{\mathcal{C}, F}(\Program) \subseteq \mathit{grnd}_{\mathcal{C}}(\Program)$
				can be exploited to ensure that the IR, which is actually computed for the ground program, carries over to the non-ground program.
				
				We first show that it is possible to replace an atom $a$ in the heads of \emph{some} rules of a ground program $\Program$ by a new atom $a'$, if a rule $a \leftarrow a'$ is added.
			}
			
			\addLemmaOutsourcedOnly{lem:atomRepalcement}{
				Let $\Program$ be a ground \hex-program, $\Program_s \subseteq \Program$ be a subprogram, and $a$ be an atom.
				We define:
				$$\Program' = \Program_s \cup \{ a \leftarrow a' \} \cup \{ H(r)|_{a \rightarrow a'} \leftarrow B(r) \mid r \in \Program \setminus \Program_s\}\text{,}$$
				where $a'$ is a new atom
				and $H(r)|_{a \rightarrow a'}$ is $H(r)$ after replacing $a$ by $a'$.
				Then $\mathcal{AS}(\Program')$ and $\mathcal{AS}(\Program)$ coincide modulo $a'$.
			}
			
			\addProof{lem:atomRepalcement}{
				We first consider the program
				$\Program'' = \Program_s \cup \{ a \leftarrow a'; \ a' \leftarrow a \} \cup \{ H(r)|_{a \rightarrow a'} \leftarrow B(r) \mid r \in \Program \setminus \Program_s \}$ instead of $\Program'$,
				i.e., we also add the reverse rule $a' \leftarrow a$.
				Then $a$ and $a'$ are equivalent and each answer set $\Assignment$ of $\Program$ corresponds one-by-one to an answer set $\Assignment'' = \Assignment \cup \{ \sigma a' \mid \sigma a \in \Assignment \}$
				(the equivalence is also given in the search for smaller models in the reduct because $a \leftarrow a'$ is in the reduct wrt.~an assignment iff $a' \leftarrow a$ is).

				Now obseve that the only difference between $\Program''$ and $\Program'$ is that $a' \leftarrow a$ is missing in $\Program'$.
				We show now that
				(i) each answer set of $\Program''$ is equivalent to an answer set of $\Program'$
				and
				(ii) each answer set of $\Program'$ is equivalent to an answer set of $\Program'$, both modulo $a'$.

				(i)
				We show that for an answer set $\Assignment''$ of $\Program''$, either $\Assignment''$ or $(\Assignment'' \setminus\{ \T a' \}) \cup \{ \F a' \}$ is an answer set of $\Program'$.
				Obviously, $\Assignment''$ is a model of $\Program''$ and thus also of $\Program' (\subseteq \Program'')$.

				\begin{itemize}
					\item If $\Assignment''$ is a least model of $f \Program'^{\Assignment''}$, it is an answer set of $\Program'$.
				
					\item Otherwise, there is a smaller model $\Assignment''_{<}$ of $f \Program'^{\Assignment''}$.
					Then, since $\Assignment''$ is a least model of $f \Program''^{\Assignment''}$, $f \Program'^{\Assignment''}$ must differ from $f \Program''^{\Assignment''}$ (only) in the absence of $a' \leftarrow a$.
					But then $(\Assignment'' \setminus\{ \T a' \}) \cup \{ \F a' \}$ must be a least model of $f \Program'^{\Assignment''}$.
					This is because since removal of $a' \leftarrow a$ caused $\Assignment''$ to be not a minimal model anymore, then obviously $a'$ lost support.
					Moreover, none of the remaining atoms can be switched to false in addition:
					since $a'$ does not occur in any rule body (except for the rule $a \leftarrow a'$; but $a$ must be true because it is true in $\Assignment''$ which is a least model $f \Program''^{\Assignment''}$)
					switching it to false cannot lead to further atoms losing their support.
					In this case, $(\Assignment'' \setminus\{ \T a' \}) \cup \{ \F a' \}$ is also a model of $\Program'$ and $f \Program'^{\Assignment''} = f \Program'^{\Assignment'}$, hence it is an answer set of $\Program'$.
				\end{itemize}
				
				(ii)
				We show that for an answer set $\Assignment'$ of $\Program'$, either $\Assignment'$ or $(\Assignment' \setminus\{ \F a' \}) \cup \{ \T a' \}$ is an answer set of $\Program''$.
				
				\begin{itemize}
					\item If $\Assignment'$ is a model of $\Program''$,
					since $\Assignment'$ is a subset-minimal model of $f \Program'^{\Assignment'}$
					and $f \Program''^{\Assignment'} \supseteq f \Program'^{\Assignment'}$, it is also a subset-minimal model of $f \Program''^{\Assignment'}$
					and thus an answer set of $\Program''$.
					
					\item If $\Assignment'$ is not a model of $\Program''$,
					then, since it is a model of $\Program'$, rule $a' \leftarrow a \in \Program''$ must be violated under $\Assignment'$; this implies $\T a \in \Assignment'$.
					But then, $(\Assignment' \setminus\{ \F a' \}) \cup \{ \T a' \}$ satisfies this rule and is a model of $\Program''$,
					because no other rule can be violated by switching $a'$ to true (the only rule which contains $a'$ in its body is $a \leftarrow a'$, but $a$ is also true in $\Assignment'$).
					We have that $\Assignment'$ is a subset-minimal model of $f \Program'^{\Assignment'}$.
					We further have that $f \Program''^{(\Assignment' \setminus\{ \F a' \}) \cup \{ \T a' \}} = f \Program'^{\Assignment'} \cup \{ a' \leftarrow a \}$.
					Then $(\Assignment' \setminus\{ \F a' \}) \cup \{ \T a' \}$ must be a subset-minimal model of $f \Program''^{(\Assignment' \setminus\{ \F a' \}) \cup \{ \T a' \}}$:
					$a'$ cannot be switched to false due to rule $a' \leftarrow a$
					and no other atom can be switched to false because $\Assignment'$ (and thus $(\Assignment' \setminus\{ \F a' \}) \cup \{ \T a' \}$, which coincides with $\Assignment'$ on such atoms)
					is subset-minimal wrt.~$f \Program'^{\Assignment'}$ (to this end, observe again that $a'$ does not appear in any rule bodies in $\Program'$ other than in rule $a \leftarrow a'$,
					hence switching it to true cannot allow any other atom to become false).
					But then $(\Assignment' \setminus\{ \F a' \}) \cup \{ \T a' \}$ is an answer set of $\Program''$.
				\end{itemize}
			}

			The main idea is then as follows.
			We add for all atoms $a$ a rule $a \leftarrow a'$ to $\mathit{pog}_{\mathcal{C}, F}(\Program)$
			in order to encode the potential support by unknown rules, where the new atom $a'$ stands for the situation that such a rule fires.
			Then an IR $R = (R^{+}, R^{-})$ of the extended program, which does not contain any such atom,
			does not depend on the absence of $a'$. Hence, adding such a rule cannot invalidate the IR and thus the IR is also valid for~$\mathit{eg}_{\mathcal{C}, F}(\Program)$ (and thus for $\Program$).
			Formally we can show:

			\appendToProofs{
				We can now show the main proposition:
			}

			\addProposition{prop:irNonground}{
				Let $\Program$ be a program and $F \subseteq D$ be input atoms from a domain $D$.
				Then an IR $R = (R^{+}, R^{-})$ of $\mathit{pog}_{\mathcal{C}, F}(\Program) \cup \{ a \leftarrow a' \mid a \in A(\mathit{pog}_{\mathcal{C}, F}(\Program)) \}$ wrt.~$D \cup \{ a' \mid a \in A(\mathit{pog}_{\mathcal{C}, F}(\Program)) \}$
				s.t.~$(R^{+} \cup R^{-}) \cap \{ a' \mid a \in A(\mathit{pog}_{\mathcal{C}, F}(\Program)) \} = \emptyset$ is an IR of $\Program$ wrt.~$D$.
			}

			\addProof{prop:irNonground}{
				In the following, let $Y = \{ a_1, \ldots, a_n \}$
				be the set
				$Y = A(\mathit{pog}_{\mathcal{C}, F}(\Program))$ of atoms in $\mathit{pog}_{\mathcal{C}, F}(\Program)$
				and $\Program_g = \mathit{pog}_{\mathcal{C}, F}(\Program)$.
				Let $\Program_{\mathit{eg}} = \mathit{eg}_{\mathcal{C}, D}(\Program)$ be the exhaustive grounding of $\Program$.
				Then by Lemma~\ref{lem:atomRepalcement} we have that
				$\Program_{\mathit{eg}} = \Program_g \cup \{ r \mid r \in \Program_{\mathit{eg}} \setminus \Program_g \}$
				is equivalent to
				$\Program_{\mathit{eg}'} = \Program_g \cup \{ a \leftarrow a' \mid a \in Y \} \cup \{ H(r)|_{Y \rightarrow Y'} \leftarrow B(r) \mid r \in \Program_{\mathit{eg}} \setminus \Program_g \}$.\footnote{
					Here, $H(r)|_{Y \rightarrow Y'}$ abbreviates $H(r)|_{a_1 \rightarrow a_1'}|_{\cdots}|_{a_n \rightarrow a_n'}$.
				}
				Then for all sets of input atoms $J \subseteq D$ we have that $\Program_{\mathit{eg}} \cup \toFacts{J}$ and $\Program_{\mathit{eg}'} \cup \toFacts{J}$ have the same answer sets, modulo atoms $a'$ for $a \in Y$.

				The idea of the proof is to show for all nogoods,
				which are added during inconsistency analysis over $\Program_g \cup \{ a \leftarrow a' \mid a \in Y \} \cup \toFacts{F}$ for some $F \subseteq D$,
				that they
				can either also be added during evaluation of $\Program_{\mathit{eg}'} \cup \toFacts{J}$ for any $J \subseteq D$ with $R^{+} \subseteq J$ and $R^{-} \cap J = \emptyset$ for an IR $R = (R^{+}, R^{-})$ of $\Program_g$,
				or contain an atom $a'$ or $\beta_{\{ a' \}}$ for $a \in Y$.
				That is, the inconsistency can either be reconstructed during evaluation of $\Program_{\mathit{eg}'} \cup \toFacts{J}$ and thus $R$ is also an IR of $\Program$ wrt.~$D$,
				or $R$ contains an atom which violates the precondition of the proposition, i.e., $R$ is found not to carry over to $\Program$.
				
				To this end,
				we first show that all initial nogoods in $\Delta_{\Program_g \cup \{ a \leftarrow a' \mid a \in Y \}}$ are either
				(i) also in $\Delta_{\Program_{\mathit{eg}'}}$,
				or (ii) contain $a'$ or an $\beta_{\{a'\}}$ (the auxiliary variable representing the body of $a \leftarrow a'$~\cite{c1977}) for some $a \in Y$.
				Consider a nogood in $\Delta_{\Program_g \cup \{ a \leftarrow a' \mid a \in Y \}}$.
				If the nogood comes from Clark's completion of $\Program_g$, then it is also in $\Delta_{\Program_{\mathit{eg}'}}$ because $\Program_{\mathit{eg}'} \supseteq \Program_g$ and Clark's completion is created rule-wise, cf.~\citeN{c1977}.
				If the nogood comes from Clark's completion of $\{ a \leftarrow a' \mid a \in Y \}$, then it contains $a'$ or $\beta_{\{ a' \}}$ for some $a \in Y$.
				If the nogood comes from the singleton loop nogood for an atom other than those in $Y$, it is also in $\Delta_{\Program_{\mathit{eg}'}}$ since the nogood depends only on rules which contain such an atom in their head~\cite{Drescher08conflict-drivendisjunctive};
					however, $\Program_{\mathit{eg}} \setminus \Program_g$ does not define such an atom because $Y$ includes all ud-atoms (these are all atoms appearing in a head of $\Program_{\mathit{eg}} \setminus \Program_g$) and each $a \in Y$ in a head of $\Program_{\mathit{eg}} \setminus \Program_g$ is replaced by $a'$.
				If the nogood comes from the singleton loop nogood for an atom $a \in Y$, then it contains $a'$ or $\beta_{\{ a' \}}$ due to the rule $a \leftarrow a'$.
				Therefore, all nogoods in $\Delta_{\Program_g \cup \{ a \leftarrow a' \mid a \in Y \}}$ are either (i) also in $\Delta_{\Program_{\mathit{eg}'}}$ or (ii) contain $a'$ or an $\beta_{\{a'\}}$ for some $a \in Y$.

				Moreover, all nogoods $N$ learned at a later point either contain a literal $\T a'$ resp.~$\F a'$ for some $a \in Y$, or can also be learned when performing inconsistency analysis over $\Program_{\mathit{eg}'} \cup \toFacts{J}$ for some $J \subseteq D$.
				This is shown by induction on the number $n$ of resolution steps performed to derive $N$.
					For the base case $n = 0$, nogood $N$ must have been added either
						(i) due to a violation of the compatibility or minimality criterion at~\ref{alg:hexcdnl:d},
						or (ii) by theory propagation at~\ref{alg:hexcdnl:e}.
						In case (i) it contains a literal $\F a'$ for all $a \in Y$,
						in case (ii) the nogood is independent of rules in $\Program_{\mathit{eg}} \setminus \Program_g$.
					For $n \rightarrow n + 1$, nogood $N$ is the resolvent of two other nogoods $N_1$ and $N_2$.
						But then the claim holds for $N_1$ and $N_2$ either because they come from $\Delta_{\Program_g \cup \{ a \leftarrow a' \mid a \in Y \}}$
						(in which case they also also in $\Delta_{\Program_{\mathit{eg}'}}$ as shown above),
						or because they have been added at at later point, in which case the claim holds by induction hypothesis.
						Moreover, atoms $a'$ are never resolved during conflict analysis because they are assigned at decision level $0$ at~\ref{alg:hexcdnl:a},
						thus all literals $a'$ for $a \in Y$ contained in $N_1$ and $N_2$ are still contained in $N$. But then the claim holds also for nogood $N$ derived by $n + 1$ resolution steps.

				But then each IR $R = (R^{+}, R^{-})$ of $\Program_g \cup \{ a \leftarrow a' \mid a \in Y \}$
				with $(R^{+} \cup R^{-}) \cap \{ a', \beta_{\{a'\}} \mid a \in Y \} = \emptyset$ is also an IR of $\Program_{\mathit{eg}'}$
				because the conflict can also be derived during conflict analysis over $\Program_{\mathit{eg}'} \cup \toFacts{J}$ for some $J \subseteq D$.
				Moreover, since $\Program_{\mathit{eg}}$ is equivalent to $\Program_{\mathit{eg}'}$ modulo atoms $a'$ for $a \in Y$,
				each IR of $\Program_{\mathit{eg}'}$ is an IR (wrt.~$D$) of $\Program_{\mathit{eg}}$ and thus, by definition of the semantics of non-ground programs, also of $\Program$.
				Hence, an IR $R = (R^{+}, R^{-})$ of $\Program_g \cup \{ a \leftarrow a' \mid a \in Y \}$ with $(R^{+} \cup R^{-}) \cap \{ a', \beta_{\{a'\}} \mid a \in Y \} = \emptyset$ is also an IR of $\Program$.
				Finally, since during conflict analysis all atoms are assigned at decision level $0$ and $\beta_{\{ a' \}}$ occurs only in binary nogoods for all $a \in Y$,
				it will always have an implicant and will be resolved away, hence, it suffices to check the candidate IR $R = (R^{+}, R^{-})$ for $(R^{+} \cup R^{-}) \cap \{ a' \mid a \in Y \} = \emptyset$.
			}

			\begin{example}[cont'd]
				\label{ex:pogrounding2}
				For the program $\Program = \{ q(X) \leftarrow p(X); \ \leftarrow \naf q(1); \ \leftarrow a \}$ and domain $D = \{ a, p(1) \}$ from Example~\ref{ex:ogrounding},
				and $F = \emptyset$ we have $\mathit{pog}_{\mathcal{C}, F}(\Program) = \{ \leftarrow \naf q(1); \ \leftarrow a \}$.
				However, for computing inconsistency reasons we
				extend the program by the rule $q(1) \leftarrow q(1)'$
				to $\{ q(1) \leftarrow q(1)' \} \cup \mathit{pog}_{\mathcal{C}, F}(\Program)$
				to express that $q(1)$ might be supported by unknown rules, which have been optimized away when grounding wrt.~specific facts $F = \emptyset$.
				Then $\langle \emptyset, \{ q(1)' \} \rangle$ is an IR of $\{ q(1) \leftarrow q(1)' \} \cup \mathit{pog}_{\mathcal{C}, F}(\Program)$ wrt.~$D \cup \{ q(1)' \}$,
				but since it contains $q(1)'$, it is excluded is an IR of $\Program$ wrt.~$D$.
				In contrast, $\langle \emptyset, \{ a \} \rangle$ as an IR of $\{ q(1) \leftarrow q(1)' \} \cup \mathit{pog}_{\mathcal{C}, F}(\Program)$ wrt.~$D \cup \{ q(1)' \}$,
				and since it does not contain $q(1)'$, it is also an IR of $\Program$ wrt.~$D$.
			\end{example}

\nop{ 
			\addProposition{prop:irNongroundUFSProp}{
				Proposition~\ref{prop:irNonground} still holds if $\text{\Propagation}$ in Algorithm~\ref{alg:hexcdnl} also performs unfounded set propagation as by~\citeN{Drescher08conflict-drivendisjunctive}.
			}

			\addProof{prop:irNongroundUFSProp}{
				We have to extend the proof of Proposition~\ref{prop:irNonground}
				by showing that it holds also for all nogoods $N$ added by unfounded set propagation,
				that $N$ either contains a literal $\T a'$ resp.~$\F a'$ for some $a \in Y$, or could also be learned when performing inconsistency analysis over $\Program_{\mathit{eg}'} \cup \toFacts{J}$ for some $J \subseteq D$.
				Without loss of generality, we assume for this proof that nogood set $\Delta$ is subject to a subsumption check,
				i.e., nogoods which are subsumed by another nogood wrt.~set inclusion are automatically eliminated.

				Akin to Proposition~\ref{prop:algGetIncReasonUFSProp}, consider an atom $a \in Y$ and an unfounded set $U$.
				If $a' \in U$, then either
					(i) $N$ is independent of $a$ (i.e., it can be learned also during inconsistency analysis over $\Program_{\mathit{eg}'} \cup \toFacts{J}$ for a (possibly) different input $J$),
					or (ii) contains $\T a'$ (in this case nogood $\{ \T a' \}$ is already in $\Delta$ and subsumes $N \supseteq \{ \T a' \}$, which is therefore not added due to the subsumption check).
				If $a \not\in U$, then $N$ contains a literal $\F a'$ if absence of supporting rules for $a'$ was necessary to justify satisfaction of a supporting rule independently of $U$.
			}

			Note that in Proposition~\ref{prop:irNonground}, set $Y$ must include all ud-atoms but can include additional atoms
			but can contain additional atoms.
			Thus, in the simplest case, one could just use $Y = A(\mathit{pog}_{\mathcal{C}, F}(\Program))$,
			but in order to ensure that many IRs of $\mathit{pog}_{\mathcal{C}, F}(\Program)$ can be identified as IRs of $\Program$,
			it is advantageous to keep the set small.
}

\nop{ 
			\medskip
			\noindent\emph{Using a procedural algorithm.}
			A sound and complete alternative is using a procedural algorithm which implements Proposition~\ref{prop:irCharacterization}; a naive one is shown in Algorithm~\ref{alg:irComputation}.

			\begin{algorithm}[t]
				\caption{IRComputation}
				\label{alg:irComputation}
				\DontPrintSemicolon

				\KwIn{general \hex-program $P$ and a domain $D$}
				\KwOut{all IRs of $P$ wrt.~$D$}

				\smallskip

				\For{all classical models $M$ of $P$}{
					\For{all $R^{+} \subseteq D$ such that $R^{+} \not\subseteq M$}{
						\For{all $R^{-} \subseteq D$ such that $R^{+} \cap R^{-} \not= \emptyset$}{
							Output $(R^{+}, R^{-})$\;
						}				
					}
					\For{all $R^{-} \subseteq D$ such that there is a UFS $U$ of $P$ wrt.~$M$ with $U \cap M \not= \emptyset$ and $U \cap (D \setminus R^{-}) = \emptyset$}{
						\For{all $R^{+} \subseteq D$ such that $R^{+} \cap R^{-} \not= \emptyset$}{
							Output $(R^{+}, R^{-})$\;
						}				
					}
				}
			\end{algorithm}
			
			\addProposition{prop:algIrComputation}{
				For a general \hex-program $P$ and a domain $D$,
				$\text{IRComputation}(P, D)$ outputs all IRs of $P$ wrt.~$D$.
			}
			
			\addProof{prop:algIrComputation}{
				The algorithm enumerates all pairs of sets $(R^{+}, R^{-})$ as by Proposition~\ref{prop:irCharacterization}.
			}
		
			However, in general efficient computation of IRs is a challenging problem by itself, which calls for algorithmic optimizations.
			We will present one such algorithm in the next section in context of our envisaged application for efficient \hex-program solving.
}

\nop{ 

		\subsection{Optimizations}
		\label{sec:computing:nonground:incompleteproceduraloptimizations}
		
			In the previous subsection we have extended the grounding $\mathit{pog}_{\mathcal{C}, F}(\Program)$
			by rules $a \leftarrow a'$ for all atoms appearing in the grounding in order to express that they might
			be defined by rules which have been optimized away wrt.~the current set $F$ of input facts.
			This is overly cautious as, in general, not all of them might be actually defined by rules which have been dropped form the grounding.

			To this end, we introduce the concept of \emph{(potentially) underdefined atoms}.
			Intuitively, it captures atoms in $\mathit{pog}_{\mathcal{C}, F}(\Program)$ of a program $\Program$ with a specific input $F \subseteq D$ from a domain $D$,
			whose defining rules may be incomplete,
			i.e., which occur in rule heads of $\mathit{eg}_{\mathcal{C}, D}(\Program) \setminus \mathit{pog}_{\mathcal{C}, F}(\Program)$.
			That is, these atoms might have support wrt.~another selection of input atoms, which cannot be seen in the current grounding.
			
			\begin{definition}[Underdefined Atom (ud-atom)]
				\label{def:udatom}
				Let $\Program$ be a program and $F \subseteq D$ be input atoms from a domain $D$.
				An atom $a$ is \emph{underdefined wrt.~$\Program$ and $F$},
				if there is a rule $r \in \mathit{eg}_{\mathcal{C}, D}(\Program) \setminus \mathit{pog}_{\mathcal{C}, F}(\Program)$ such that $a \in H(r)$.
			\end{definition}

			\begin{example}[cont'd]
				Consider the program $\Program$, domain $D$ and input $F = \emptyset$ from Example~\ref{ex:pogrounding}
				and suppose $\mathit{pog}_{\mathcal{C}, F}(\Program) = \{ \leftarrow \naf q(1); \ \leftarrow a \}$.
				The atom $q(1)$ is underdefined wrt.~$\Program$ and $F$
				because $q(1) \leftarrow p(1) \in \mathit{eg}_{\mathcal{C}, D}(\Program) \setminus \mathit{pog}_{\mathcal{C}, F}(\Program)$.
			\end{example}

			However, deciding definitely if an atom is underdefined without
			explicitly constructing $\mathit{eg}_{\mathcal{C}, D}(\Program)$ (which we want to avoid as discussed)
			is challenging and appears to be impossible in general.
			We thus go for a conservative approach and use $\mathit{pog}_{\mathcal{C}, F}(\Program)$ to identify atoms which are \emph{definitely not underdefined};
			for all others we assume that they might be.
			We first compute an \emph{envelope} $\mathit{env}(\Program)$ of $\Program$, i.e.,
			a superset of all atoms which are relevant when evaluating $\Program$ under some $F \subseteq D$:

			\begin{definition}
				For a program $\Program$ we let
				\begin{align*}
					\mathit{env}(\Program) = \{ & a \mid \T a \in \Assignment, \Assignment \in \mathcal{AS}(\Program') \}, \text{where} \\
					\Program' = \{ & \sigma(a) \leftarrow B^{+}_o(r) \mid r \in \Program, a \in H(r) \cup B^{-}_o(r), \\
									& \sigma = \{ X \rightarrow \upsilon \mid X \in \mathit{vars}(a) \setminus \mathit{vars}(B^{+}_o(r)) \} \}\text{.}
				\end{align*}
			\end{definition}
			
			Intuitively, we construct a positive program to compute in its unique answer set the maximum set of relevant atoms,
			where an atom is considered relevant if it occurs in the head of some rule in the optimized grounding, and thus might be contained in some answer set.
			Atoms which are not relevant wrt.~this definition cannot be contained in any answer set and do not need to be respected in the grounding.
			To avoid exhaustive grounding, we abstract from the values which may be introduced by external atoms and treat them just as \underline{u}nknown values represented by a dedicated constant $\upsilon$.
			To this end, we add for each rule $r \in \Program$ a separate normal rule
			\begin{itemize}
				\item whose head atom comes from $H(\sigma(r)) \cup B^{-}_o(\sigma(r))$,
				\item whose body is $B^{+}_o(r)$, and
				\item all variables which appear only in the head of the resulting rule are replaced by the new constant $\upsilon$.
			\end{itemize}
			
			One can formally show:
			
			\addLemma{lem:domainExplorationProgram}{
				For a program $\Program$, input $F \subseteq D$ and a minimal model $\Assignment$ of $\mathit{eg}_{\mathcal{C}, D}(\Program) \cup \toFacts{F}$,
				whenever $\T a \in \Assignment$ then $a$ coincides with some $a' \in \mathit{env}(\Program)$ modulo constants $\upsilon$.
			}
			
			\addProof{lem:domainExplorationProgram}{
				Splitting up rules into a normal rule for each head atom and default-negated atom
				leads to a program with a unique answer set, which contains all atoms that can occur in a subset-minimal model of the original program or a subset thereof.
				This is because no other atom is supported by any rule, or occurs in a default-negated part which could force the atom to be true in order to satisfy the rule, and thus cannot occur in any minimal model
				of any program $\Program' \subseteq \mathit{eg}_{\mathcal{C}, D}(\Program)$ due to the minimality criterion.
				Moreover, removal of external atoms and substitution of all variables, which do not occur elsewhere in the rule body,
				by $\upsilon$ abstracts from values introduced by external atoms. That is, $\upsilon$ represents that (possibly multiple) unknown values
				may occur at this position. Hence, $\mathit{env}(\Program)$ contains all atoms which can occur in a minimal model of $\Program' \subseteq \mathit{eg}_{\mathcal{C}, D}(\Program)$,
				where atoms containing $\upsilon$ represent sets of unknown atoms.
			}

			One can further show that rules, whose bodies are never satisfied under minimal models, are superfluous:

			\addLemma{lem:domainExploration}{
				For a program $\Program$ and $r \in \Program$, if $\Assignment \not\models B^{+}_o(r)$ for all minimal models
				$\Assignment$ of $\Program \setminus \{ r \}$, then $\mathcal{AS}(\Program) = \mathcal{AS}(\Program \setminus \{ r \})$.\footnote{
					Here, minimal models are meant in the classical sense. This includes all answer sets but possibly even more assignments.
				}
			}
			
			\addProof{lem:domainExploration}{
				We have to show that $\Assignment \in \mathcal{AS}(\Program)$ iff $\Assignment \in \mathcal{AS}(\Program \setminus \{ r \})$.
			
				$(\Rightarrow)$
				Let $\Assignment \in \mathcal{AS}(\Program)$. We make a case distinction.
				\begin{itemize}
					\item If we have $f (\Program \setminus \{ r \})^{\Assignment} = f \Program^{\Assignment}$, then $\Assignment$ is an answer set of $\Program \setminus \{ r \}$.
						This is because it is subset-minimal model of $f \Program^{\Assignment}$ and therefore, due to the equivalence of the reducts, also of $f (\Program \setminus \{ r \})^{\Assignment}$.
					\item If we have $f (\Program \setminus \{ r \})^{\Assignment} \subsetneq f \Program^{\Assignment}$, then $\Assignment \models B(r)$ and thus in particular $\Assignment \models B^{+}_o(r)$.
						Then $\Assignment$ cannot be a subset-minimal model of $\Program \setminus \{ r \}$ because this would imply, by precondition, that $\Assignment \not\models B^{+}_o(r)$.
						However, it is still a (non-minimal) model of $\Program \setminus \{ r \}$ because it is a model of $\Program$.
						But if $\Assignment$ is a model but not a subset-minimal model of $\Program \setminus \{ r \}$, then there is a smaller subset-minimal model $\Assignment'$.				
						But then it holds by precondition that $\Assignment' \not\models B^{+}_o(r)$ and thus $\Assignment' \models r$, hence $\Assignment'$ is also a model of $f \Program^{\Assignment}$, which is smaller than $\Assignment$.
						This contradicts the assumption that $\Assignment$ is an answer set of $\Program$.
				\end{itemize}
				
				$(\Leftarrow)$
				Let $\Assignment \in \mathcal{AS}(\Program \setminus \{ r \})$.
				Then it is also a minimal model of $\Program \setminus \{ r \}$ and, by precondition, we have that $\Assignment \not\models B^{+}_o(r)$.
				But then $f (\Program \setminus \{ r \})^{\Assignment} = f \Program^{\Assignment}$
				and, since $\Assignment$ is a subset-minimal model of $f (\Program \setminus \{ r \})^{\Assignment}$, it is also one of $f \Program^{\Assignment}$.
				Thus it is an answer set of $\Program$.
			}

			We now check for an atom if it \emph{might} be derived by a rule in $\mathit{eg}_{\mathcal{C}, D}(\Program)$,
			that is not already included in $\mathit{pog}_{\mathcal{C}, F}(\Program)$.
			That is, we identify atoms which are certainly not underdefined.

			\addProposition{prop:domainExploration}{
				Let $\Program$ be a program and $F \subseteq D$ be input atoms from a domain $D$.
				If an atom $a$ unifies with some $h \in H(r)$ for some $r \in \Program$ using $\rho\colon \mathcal{V} \rightarrow \mathcal{C}$,
				and for all $\rho'\colon \mathcal{V} \rightarrow \mathcal{C}$ with $\rho'(\rho(B^{+}_o(r))) \subseteq \mathit{env}(\Program)$
				we have that $\rho'(\rho(r)) \in \mathit{pog}_{\mathcal{C}, F}(\Program)$,
				then $a$ is \emph{not} underdefined wrt.~$\Program$ and $F$.
			}
			
			\addProof{prop:domainExploration}{
				If an atom $a$
				unifies with some $h \in H(r)$ for some $r \in \Program$ using $\rho\colon \mathcal{V} \rightarrow \mathcal{C}$ as unifier,
				then $a$ might be supported by a rule in $\mathit{eg}_{\mathcal{C}, D}(\Program)$.
				More precisely, it might be supported by all instances $\rho'(\rho(r))$ of $r$ with $\rho'\colon \mathcal{V} \rightarrow \mathcal{C}$.
				However, for instances $\rho'(\rho(r))$ with a $\rho'\colon \mathcal{V} \rightarrow \mathcal{C}$ such that
				$\rho'(\rho(B^{+}_o(r))) \not\subseteq \mathit{env}(\Program)$ we have by Lemma~\ref{lem:domainExplorationProgram}
				that $\rho'(\rho(B^{+}_o(r)))$ cannot be satisfied under any minimal model of any
				$\Program' \subseteq \mathit{eg}_{\mathcal{C}, D}(\Program) \cup \toFacts{J}$ for any $J \subseteq D$,
				in particular not under those of $(\mathit{eg}_{\mathcal{C}, D}(\Program) \cup \toFacts{J}) \setminus \rho'(\rho(r))$.
				Then we have by Lemma~\ref{lem:domainExploration}
				that $\mathcal{AS}(\mathit{eg}_{\mathcal{C}, D}(\Program \cup \toFacts{J})) = \mathcal{AS}((\mathit{eg}_{\mathcal{C}, D}(\Program) \cup \toFacts{J}) \setminus \rho'(\rho(r)))$ for all $J \subseteq D$.
				But then, due to assumed minimality of $\mathit{eg}_{\mathcal{C}, D}(\Program \cup \toFacts{J})$, it does not contain $\rho'(\rho(r))$.

				But then only the rules $\rho'(\rho(r))$ with $\rho'\colon \mathcal{V} \rightarrow \mathcal{C}$ such that $\rho'(\rho(B^{+}_o(r))) \subseteq \mathit{env}(\Program)$
				might support $a$. If all of them are already in $\mathit{pog}_{\mathcal{C}, F}(\Program)$, then $a$ cannot be underdefined.
			}
		
			The idea of the proof is as follows. Set $\mathit{env}(\Program)$ contains all atoms which might be relevant when evaluating $\Program \cup F$ for any $F \subseteq D$,
			i.e., atoms not in $\mathit{env}(\Program)$ can be disregarded.
			To decide if an atom $a$ is definitely not underdefined,
			we check for all rules $r \in \Program$ whose head unifies with $a$, and thus defines it, if all relevant ground instances
			of $r$ wrt.~$\mathit{env}(\Program)$ are already included in $\mathit{pog}_{\mathcal{C}, F}(\Program)$.
			
			As a generalization of Proposition~\ref{prop:irNonground} we can then show:
			
			\addProposition{prop:irNongroundGen}{
				Let $\Program$ be a program, $F \subseteq D$ be input atoms from a domain $D$,
				and $Y$ be a set of atoms including all ud-atoms wrt.~$\Program$ and $F$.
				Then an IR $R = (R^{+}, R^{-})$ of $\mathit{pog}_{\mathcal{C}, F}(\Program) \cup \{ a \leftarrow a' \mid a \in Y \}$ wrt.~$D \cup \{ a' \mid a \in Y \}$
				with $(R^{+} \cup R^{-}) \cap \{ a' \mid a \in Y \} = \emptyset$ is an IR of $\Program$ wrt.~$D$.
			}

			\addProof{prop:irNongroundGen}{
				The proof for Proposition~\ref{prop:irNonground} still goes through as is,
				even if $Y$ (see first sentence of the proof)
				is set to an arbitrary set of atoms which includes all ud-atoms wrt.~$\Program$ and $F$.
			}
}

	\section{Application: Exploiting Inconsistency Reasons for \hex-Program Evaluation}
	\label{sec:hexprogramevaluation}
	
		We now exploit inconsistency reasons for improving the evaluation algorithm for \hex-programs.
		In this section we focus on a class of programs which cannot be efficiently tackled by existing approaches.
		Current evaluation techniques inefficient for programs with guesses that are separated from
		constraints by external atoms which (i)~are \emph{nonmonotonic}, and (ii)~\emph{introduce new values} to the program (called \emph{value invention}).
		Here, an external atom $\ext{g}{\vec{p}}{\cdot}$ is called monotonic if
		the output never shrinks when more input atoms become true;
		formally: if for all output vectors $\vec{c}$
		and assignments $\Assignment$, $\Assignment'$ with $\{ \T a \in \Assignment' \} \supseteq \{ \T \in \Assignment \}$
		it is guaranteed that $\Assignment \models \ext{g}{\vec{p}}{\vec{c}}$ implies $\Assignment' \models \ext{g}{\vec{p}}{\vec{c}}$.
		Otherwise it is called nonmonotonic.
		
		We start with a recapitulation of current evaluation techniques and point out their bottlenecks.

		\subsection{Existing Evaluation Techniques for \hex-Programs}
		\label{sec:hexprogramevaluationn:existing}

			Currently, there is a \emph{monolithic} evaluation approach which evaluates the program as a whole,
			and another one which \emph{splits} it into multiple program components.
		
			\leanparagraph{Monolithic Approach}
			Safety criteria for \hex-programs guarantee the existence of a finite grounding,
			which allows for evaluating a program by separate grounding an solving phases akin to ordinary ASP.
			A grounding algorithm was presented by~\citeN{efkr2016-aij},
			the solving algorithm was recapitulated in Algorithm~\ref{alg:hexcdnl} in Section~\ref{sec:computing}.
			
			Since external sources may introduce new constants, determining the set of relevant constants
			requires the external sources to be evaluated already during grounding.
			While this set is finite, it is in general expensive to compute
			because an up to exponential number of evaluation calls is necessary to construct a single ground instance of a rule.
			Intuitively, this is the case because nonmonotonic external atoms which provide value invention
			must be evaluated under all possible inputs to observe all possible output values.
			We demonstrate this with an example.

			\begin{example}
				\label{ex:committee}
				Suppose we want to form a committee of employees.
				Some pairs of persons should not be together in the committee due to conflicts of interests
				(cf.~independent sets of a graph).
				The competences of the committee depend on the involved persons.
				For instance, it can decide in technical questions only if a certain number of members has expert knowledge in the field.
				The competences can depend nonmonotonically on the members.
				For instance, while overrepresentation of a department might not be forbidden altogether,
				it can make it lose authorities such as assigning more resources to this department.
				Constraints define the competences the committee should have.
				This is encoded by the program
				\begin{align*}
					P = \{ r_1\colon& \mathit{in}(X) \vee \mathit{out}(X) \leftarrow \mathit{person}(X). \\
							r_2\colon&  \leftarrow\mathit{in}(X), \mathit{in}(Y), \mathit{conflict}(X,Y). \\
							r_3\colon& \mathit{comp}(X) \leftarrow \ext{\mathit{competences}}{\mathit{in}}{X}. \\
							r_4\colon& \leftarrow \naf \mathit{comp}(\mathit{technical}), \naf \mathit{comp}(\mathit{financial}). \} \cup F\text{,}
				\end{align*}
				where $F$ is supposed to be a set of facts over predicate $\mathit{person}$, which specify the available employee.
				
				Rule $r_1$ guesses candidate commitees, rule $r_2$ excludes conflicts of interest
				(defines as facts over predicate $\mathit{conflict}$), $r_3$ determines competences of the candidate,
				and $r_4$ defines that we do not want committees which can neither decide in technical nor financial affairs.
			\end{example}

			Suppose program $P$ from Example~\ref{ex:committee} is grounded as a whole before solving starts.
			Then, without further information, the grounder must evaluate $\ext{\mathit{competences}}{\mathit{in}}{X}$
			under all possible candidate committees, i.e.,
			under all consistent assignments $\Assignment \subseteq \{ \T \mathit{in}(a), \F \mathit{in}(a) \mid \mathit{person}(a) \leftarrow \in F \}$
			to make sure that all possible competences $X$ are observed.
			In other words, the grounder must determine the set
			$$\{ \vec{c} \mid \Assignment \models \ext{\mathit{competences}}{\mathit{in}}{\vec{c}} \text{ for some consistent } \Assignment \subseteq \{ \T \mathit{in}(a), \F \mathit{in}(a) \mid \mathit{person}(a) \leftarrow \in F \} \}\text{.}$$
			While many candidate committees (and their competences) turn out to be irrelevant when solving the ground program
			because they are already eliminated by $r_2$, this is not discovered in the grounding phase.

			Note that this bottleneck occurs only if the external atom is (i) nonmonotonic, and (ii) can introduce new values to the program.
			If condition (i) would not be given, i.e., if the external atom would be known to the grounder to be monotonic
			(that is, the set of competences can only increase if more people join the committee),
			then it would suffice to evaluate it under the maximal assignment $\Assignment = \{ \T \mathit{in}(a) \mid \mathit{person}(a) \leftarrow \in F \}$
			to observe all relevant constants.
			If condition (ii) would not be given, i.e., it is known to the grounder that the external atom does not introduce constants which are not in the program,
			then the grounder would not need to evaluate it at all during grounding since all relevant constants are already in the program.

			Hence, for the considered class of programs, the approach usually suffers a bottleneck due to a large number of external calls during grounding.
			
			\leanparagraph{Splitting Approach}
			An alternative evaluation approach is implemented by a \emph{model-building framework} based on program splitting~\cite{efikrs2016-tplp}.
			Based on dependencies between rules, the program is split it into components, called \emph{(evaluation) units},
			which are arranged in an \emph{acyclic evaluation graph};
			acyclicity means informally that a later unit cannot derive an atom (i.e., have it in some rule head)
			that already occurred in predecessor units. The approach exploits a theorem similar to the splitting theorem by~\citeN{lifs-turn-94}.
			
			In a nutshell, the evaluation works unit-wise: each unit is separately grounded and solved, beginning from the units without predecessors.
			The framework first computes the answer sets of units without predecessors.
			For each answer set, the successor units are extended with the true atoms from the answer set as facts,
			and it itself grounded and solved. This procedure is repeated recursively,
			where the final answer sets are extracted from the leaf units.

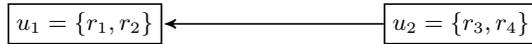
\begin{figure}[h]
				\centering
				\footnotesize
				\beginpgfgraphicnamed{committee}
				\begin{tikzpicture}[->,>=stealth',shorten >=1pt,auto,node distance=5cm,semithick]
					\node         (u1) [draw]                   {$u_1=\{r_1, r_2\}$};
					\node         (u2) [draw,right of=u1]       {$u_2=\{r_3, r_4\}$};

					\path (u2) edge [->] node {} (u1);
				\end{tikzpicture}
				\endpgfgraphicnamed

				\caption{Evaluation graph of the program from Example~\ref{ex:committee}}
				\label{fig:committee}
			\end{figure}

			\begin{example}[cont'd]
				\label{ex:committee2}
				Reconsider program $P$ from Example~\ref{ex:committee}.
				It can be partitioned into units $u_1 = \{ r_1, r_2 \}$ and $u_2 = \{ r_3, r_4 \}$,
				where $u_2$ depends on $u_1$ because it uses atoms from $u_1$, but does not redefine them, cf.~Figure~\ref{fig:committee}.
				For the evaluation, we first compute the answer sets $\mathcal{AS}(u_1)$ of program component $u_1$,
				which represent all committee candidates, excluding those with conflicts of interest (due to $r_2 \in u_1$).
				For each answer set $\Assignment \in \mathcal{AS}(u_1)$,
				we ground and solve $u_2$ extended with the true atoms from $\Assignment$ as facts,
				i.e., we compute $\mathcal{AS}(u_2 \cup \{ a \leftarrow \mid \T a \in \Assignment \})$.
				The answer sets of the overall program correspond one-by-one to
				$\bigcup_{\Assignment \in \mathcal{AS}(u_1)} \mathcal{AS}(u_2 \cup \{ a \leftarrow \mid \T a \in \Assignment \})$.
			\end{example}
			
			Observe that when grounding $u_2 \cup \{ a \leftarrow \mid \T a \in \Assignment \}$
			for a certain answer set $\Assignment$ of $u_1$ in Example~\ref{ex:committee2},
			the input to the external atom $\ext{\mathit{competences}}{\mathit{in}}{X}$
			consists of facts only, i.e., it is already fixed.
			This is exploited to determine all relevant constants with a single external call.
			
			However, the drawback of this approach is that it splits the guessing part from the checks:
			rule $r_4$ is separated from the guess in rule $r_1$
			and a conflict with $r_4$ during the evaluation of $u_2 \cup \{ a \leftarrow \mid \T a \in \Assignment \}$
			for some $\Assignment \in \mathcal{AS}(u_1)$
			cannot be propagated to unit $u_1$ to exclude further guesses, which may lead to a related conflict in $u_2$.

			Hence, for the considered class of programs, this technique also suffers a bottleneck,
			but one of a different nature than for the monolithic approach.
			While grounding a single program unit for a fixed input from its predecessor unit is now efficient,
			the problem is now that later units must be instantiated for many different inputs from predecessor units.
			In the example, unit $u_2$ needs to be grounded for all answer sets of $u_1$, although many of them
			may make $u_2$ inconsistent because of the same reason.

			Although the framework supports arbitrary acyclic evaluation graphs as stated above,
			we only consider the special case of a list of units in this paper (as in Example~\ref{ex:committee2}).
			This is because it is sufficient for explaining the main idea of our new evaluation technique
			and saves us from the need to introduce the complete framework formally, which is cumbersome
			and would distract from the core idea. However, in practice the idea can be easily applied
			also to general acyclic evaluation graphs.

\nop{
		\subsection{The Model-Building Framework for \hex-Program Evaluation}
		\label{sec:hexprogramevaluationn:framework}

			\newcommand{\ipar}{\textsc{i}}
			\newcommand{\opar}{\textsc{o}}
			\newcommand{\modtypei}{\textsc{i}}
			\newcommand{\modtypeo}{\textsc{o}}
		
			The basic idea is based on a notion of rule dependencies.
			Moreover, the new framework is also more flexible because it abstractly
			uses \emph{evaluation graphs} which define the ordering of program evaluation,
			while the former approach hard-coded this in the algorithms.
			\begin{definition}[Rule Dependencies]
				\label{def:ruledep}
				Let $\Program$ be a program with rules $r, s \in \Program$.
				We denote by $r \rightarrow_m s$ (resp.~$r \rightarrow_n s$)
				that $r$ \emph{depends monotonically} (resp.~\emph{depends nonmonotonically}) on $s$.
				We define:
				\begin{enumerate}[(i)]
					\item\label{def:ruledep:i} If $a \in B^{+}(r), b \in H(s)$ and $a \sim b$, then $r \rightarrow_m s$.
					\item\label{def:ruledep:ii} If $a \in B^{-}(r), b \in H(s)$ and $a \sim b$, then $r \rightarrow_n s$.
					\item\label{def:ruledep:iii} If $a \in H(r), b \in H(s)$ and $a \sim b$, then both $r \rightarrow_m s$ and $s \rightarrow_m r$.
					\item\label{def:ruledep:iv} If $a \in B(r)$ is an external atom of form $\ext{g}{\vec{Y}}{\vec{X}}$ where $\vec{Y} = Y_1, \ldots, Y_n$,
								the input $Y_i = p$ for $1 \le i \le \mathit{ar}_{\ipar}(\amp{g})$ has $\mathit{type}(\amp{g}, i) = \mathbf{pred}$, and $b \in H(s)$ is an atom of form $p(\vec{Z})$, then
								\begin{itemize}
									\item $r \rightarrow_m s$ if $\amp{g}$ is monotonic (in all predicate parameters) and $a \in B^{+}(r)$; and
									\item $r \rightarrow_n s$ otherwise.
								\end{itemize}
				\end{enumerate}
			\end{definition}
			
			Note that the dependency in Condition~(\ref{def:ruledep:iv}) is considered monotonic
			only if the external atom is monotonic in \emph{all} parameters.
			This is because the former formalization (and implementation) of \hex{} did not
			distinguish between different parameters, i.e., there was only a global monotonicity attribute
			of each external predicate. In this thesis we have a more fine-grained approach and
			use a separate monotonicity attribute for each predicate parameter, which allows
			for using a more liberal definition of rule dependencies (monotonic dependencies are usually
			advantageous over nonmonotonic ones). However, as this thesis does not focus
			on the evaluation framework, we stick with the former definition at this point
			and leave the formalization of the relaxed notion for future work.
			
			\begin{example}[ctd.]
				\label{ex:frameworkctd}
				For the program from Example~\ref{ex:framework} we have the following dependencies,
				which are visualized in Figure~\ref{fig:exruledep}:
				\begin{itemize}
					\item $r_1 \rightarrow_m r_2$ and $r_2 \rightarrow_m r_1$ by (\ref{def:ruledep:iii})
					\item $r_3 \rightarrow_n r_1$ and $r_3 \rightarrow_n r_2$ by (\ref{def:ruledep:iv})
					\item $r_4 \rightarrow_n r_1$ and $r_4 \rightarrow_n r_2$ by (\ref{def:ruledep:iv})
					\item $r_5 \rightarrow_m r_1$ and $r_5 \rightarrow_m r_2$ by (\ref{def:ruledep:i})
					\item $r_6 \rightarrow_m r_3$ by (\ref{def:ruledep:i})
					\item $r_7 \rightarrow_m r_5$ by (\ref{def:ruledep:i})
				\end{itemize}
				
				\begin{figure}
					\centering
					\beginpgfgraphicnamed{ruledependencies}
					\begin{tikzpicture}[->,>=stealth',shorten >=1pt,auto,node distance=2.8cm,
										semithick]
						\node         (r1)        {$r_1$};
						\node         (r2) [right of=r1]       {$r_2$};
						\node         (r3) [below of=r1]       {$r_3$};
						\node         (r4) [below of=r2]       {$r_4$};
						\node         (r5) [above of=r1]       {$r_5$};
						\node         (r6) [left of=r3]       {$r_6$};
						\node         (r7) [left of=r5]       {$r_7$};

						\path	(r1) edge [<->] node {$\rightarrow_m$}  (r2)
								(r3) edge [->] node {$\rightarrow_n$}  (r1)
								(r3) edge [->,bend right] node {$\rightarrow_n$}  (r2)
								(r4) edge [->,bend left] node {$\rightarrow_n$}  (r1)
								(r2) edge [<-] node {$\rightarrow_n$}  (r4)
								(r5) edge [->] node {$\rightarrow_m$}  (r1)
								(r5) edge [->,bend left] node {$\rightarrow_m$}  (r2)
								(r6) edge [->] node {$\rightarrow_m$}  (r3)
								(r7) edge [->] node {$\rightarrow_m$}  (r5)
						;
					\end{tikzpicture}
					\endpgfgraphicnamed
					
					\caption{Rule Dependencies of the Program from Examples~\ref{ex:framework} and~\ref{ex:frameworkctd}}
					\label{fig:exruledep}
				\end{figure}
			\end{example}
			
			The dependency graph is used to construct the \emph{evaluation graph} which controls
			the overall evaluation of the program.
			The evaluation graph is composed of extended pre-groundable \hex{}-programs as nodes,
			which are called \emph{evaluation units} in this context. The edges of the evaluation graph
			connect the evaluation units acyclically and are derived from the dependency relation between rules.
			More formally, we introduce the following concepts.
			
			\begin{definition}[(Evaluation) Unit]
				\label{def:evalunit}
				An \emph{(evaluation) unit} is an extended pre-groundable \hex{}-program.
			\end{definition}
			\begin{definition}[Evaluation Graph]
				\label{def:evalgraph}
				An \emph{evaluation graph} $\mathcal{E} = \langle V, E \rangle$ of a program $\Program$
				is a directed acyclic graph; vertices $V$ are evaluation units and $\mathcal{E}$ has the following properties:
				\begin{enumerate}[(a)]
					\item $\bigcup_{u \in V} u = \Program$, i.e., every rule $r \in \Program$ is contained in at least one unit;
					\item for every non-constraint $r \in \Program$, it holds that $\big|\{ u \in V \mid r \in u \}\big| = 1$, i.e.,
						$r$ is contained in exactly one unit;
					\item for each nonmonotonic dependency $r \rightarrow_n s$ between rules $r, s \in \Program$
						and for all $u \in V$ with $r \in u$ and $v \in V$ with $s \in v$ st.~$u \not=v$,
						there exists an edge $(u, v) \in E$, i.e., nonmonotonic dependencies between rules have
						corresponding edges everywhere in $\mathcal{E}$; and
					\item for each monotonic dependency $r \rightarrow_m s$ between rules $r, s \in \Program$,
						there exists one $u \in V$ with $r \in u$ such that $E$ contains all edges $(u, v)$ with $v \in V$, $s \in v$ and $v \not= u$,
						i.e., there is (at least) one unit in $\mathcal{E}$ where all monotonic dependencies from $r$ to other rules have
						corresponding outgoing edges in $\mathcal{E}$.
				\end{enumerate}
			\end{definition}
			
			We denote by $\mathit{preds}_{\mathcal{E}}(u)$ the predecessors of unit $u$ in $\mathcal{E} = \langle V, E \rangle$,
			i.e., $\mathit{preds}_{\mathcal{E}}(u) = \big\{ v \in V \mid (u, v) \in E \big\}$.
			For units $u$, $w$ we write $u < w$ if there exists a path from $u$ to $w$ in $\mathcal{E}$
			and $u \le w$ if $u < w$ or $u = w$.
			Moreover, for a unit $u \in V$ let $u^{<} = \bigcup_{w \in U, u < w} w$ and $u^{\le} = u^{<} \cup \{ u \}$.

			Informally, the edges of $\mathcal{E}$
			cover the rule dependencies in the sense that if $r \in v$ depends on $s \in w$ with $w \not= v \in V$,
			then there must be an edge from $v$ to $w$ in $E$.
			For the sake of simplicity of the formal results,
			it is advantageous to introduce an empty \emph{final evaluation unit} $u_{\mathit{final}}$ which depends
			on all other units, as shown in the following example.
			
			\begin{example}[ctd.]
				\label{ex:frameworkctdevalgraph}
				A valid evaluation graph for the program in Example~\ref{ex:framework} is
				$\mathcal{E} = \langle V, E \rangle$ with
				\begin{align*}
					V = \big\{& u_1 = \{ r_1, r_2 \}, u_2 = \{ r_3 \}, u_3 = \{ r_4 \}, u_4 = \{ r_5 \}, u_5 = \{ r_6, r_7 \}, u_{\mathit{final}} = \emptyset \big\} \\
					E = \big\{& (u_2, u_1), (u_3, u_1), (u_4, u_1), (u_5, u_2), (u_5, u_4), \\
						& (u_{\mathit{final}}, u_1), (u_{\mathit{final}}, u_2), (u_{\mathit{final}}, u_3), (u_{\mathit{final}}, u_4), (u_{\mathit{final}}, u_5) \big\}
				\end{align*}
				as visualized in Figure~\ref{fig:ex:frameworkctd:evalgraph} (where $u_{\mathit{final}}$ is omitted).
				\begin{figure}[h]
					\centering
					\beginpgfgraphicnamed{evaluationgraph}
					\begin{tikzpicture}[%
						start chain,
						node distance=1cm,
						every on chain/.style={join=by ->},
						every join/.style={line width=1.25pt}]
						\matrix (m) [matrix of nodes, 
						column sep=0.5mm,
						row sep=8mm,
						nodes={draw, 
						  line width=0.7pt,
						  text width=4.5cm
						},
						box/.style={
						  line width=1.0pt,
						  minimum width=3.1cm,
						  minimum height=9mm
						}
						]
						{
						  & |[box]| $\mathit{team}1(a) \vee \mathit{team}1(b) \leftarrow$ $\mathit{team}1(b) \vee \mathit{team}1(c) \leftarrow$  &  \\
						  |[box]| $\mathit{team}2(X) \leftarrow\amp{\mathit{diff}}$ \newline \hspace*{2mm} $[\mathit{employee}, \mathit{team}1](X)$  &
						  |[box]| $\mathit{team}{1a}(X) \leftarrow \amp{\mathit{diff}}$ \newline \hspace*{2mm} $[\mathit{team}1, \mathit{qualification}](X)$ &
						  |[box,text width=4.3cm]| $\mathit{team}{1b}(X) \leftarrow \mathit{team}1(X),$ \newline \hspace*{2mm} $\mathit{qualification}(X)$ \\
						  & |[box]| $\mathit{bonus}(X) \leftarrow \mathit{team}2(X)$ $\mathit{bonus}(X) \leftarrow \mathit{team}{1b}(X)$  & \\
						};
						\draw (m-1-2) node[below left=9mm] {$u_1$};
						\draw (m-2-1) node[below=6mm] {$u_2$};
						\draw (m-2-2) node[below=6mm] {$u_3$};
						\draw (m-2-3) node[below=6mm] {$u_4$};
						\draw (m-3-2) node[below=6mm] {$u_5$};
						%
						{ [start chain,every on chain/.style={join}, every join/.style={line width=1.25pt}]
						  \path[line width=1pt,->] (m-2-1) edge node [right] {} (m-1-2);
						  \path[line width=1pt,->] (m-2-2) edge node [right] {} (m-1-2);
						  \path[line width=1pt,->] (m-2-3) edge node [right] {} (m-1-2);
						  \path[line width=1pt,->] (m-3-2) edge node [right] {} (m-2-1);
						  \path[line width=1pt,->] (m-3-2) edge node [right] {} (m-2-3);
						};
					\end{tikzpicture}
					\endpgfgraphicnamed
					
					\caption{Evaluation Graph of Example~\ref{ex:frameworkctdevalgraph} without $u_{\mathit{final}}$}
					\label{fig:ex:frameworkctd:evalgraph}
				\end{figure}
				\begin{figure}[h]
					\renewcommand\arraystretch{2.4}
					\centering
					\setlength{\tabcolsep}{.3pt}
					\beginpgfgraphicnamed{answersetgraph}
					\begin{tikzpicture}[%
						remember picture,
						text centered,
						start chain,
						node distance=0.35cm,
						every on chain/.style={join=by ->},
						every join/.style={line width=1.25pt}]
						\tikzstyle{unit} = [line width=1.0pt, auto]
						\tikzstyle{model} = [line width=1.0pt, auto, text width=2.1cm]
						\tikzstyle{line} = [draw, line width=1.25pt, join=by ->]
						\matrix (m) [matrix of nodes, 
						ampersand replacement=\&,
						column sep=1mm,
						row sep=8mm,
						inner sep=1pt,
						nodes={draw, 
						  line width=0.7pt,
						  anchor=center, 
						  text centered
						}
						]
						{
							\&
							|[unit]|{
								\begin{tabular}{c c}
									\multicolumn{2}{c}{
										\tikz{\node[model,label=above:$m_1^{\modtypei}$](m1){$\emptyset$};}
									} \\
									\tikz{\node[model,label=above:$m_2^{\modtypeo}$](m2){$\{\mathit{team}1(a),$ $\mathit{team}1(c)\}$};} & \tikz{\node[model,label=above:$\ m_3^{\modtypeo}$](m3){$\{\mathit{team}1(b)\}$};}
								\end{tabular}
								}
							\& \\
							|[unit]|{
								\begin{tabular}{c c}
									\tikz{\node[model,label=above:$m_4^{\modtypei}$](m4){$\mathit{int}(m^{\modtypeo}_2)$};} & \tikz{\node[model,label=above:$m_5^{\modtypei}$](m5){$\mathit{int}(m^{\modtypeo}_3)$};} \\
									\tikz{\node[model,label=below left:$m_6^{\modtypeo}$](m6){$\{\mathit{team}2(b)\}$};} & \tikz{\node[model,label=below:$m_7^{\modtypeo}$](m7){$\{ \mathit{team}2(a),$ $\mathit{team}2(c) \}$};}
								\end{tabular}
								}
							\&
							|[unit]|{
								\begin{tabular}{c c}
									\tikz{\node[model,label=above left:$m_8^{\modtypei}$](m8){$\mathit{int}(m^{\modtypeo}_2)$};} & \tikz{\node[model,label=above left:$m_9^{\modtypei}$](m9){$\mathit{int}(m^{\modtypeo}_3)$};} \\
									\tikz{\node[model,label=below:$m_{10}^{\modtypeo}$](m10){$\{\mathit{team}{1a}(a)\}$};} & \tikz{\node[model,label=below:$m_{11}^{\modtypeo}$](m11){$\{ \mathit{team}{1a}(b) \}$};}
								\end{tabular}
								}
							\&
							|[unit]|{
								\begin{tabular}{c c}
									\tikz{\node[model,label=above:$m_{12}^{\modtypei}$](m12){$\mathit{int}(m^{\modtypeo}_2)$};} & \tikz{\node[model,label=above:$m_{13}^{\modtypei}$](m13){$\mathit{int}(m^{\modtypeo}_3)$};} \\
									\tikz{\node[model,label=below:$m_{14}^{\modtypeo}$](m14){$\{\mathit{team}{1b}(c)\}$};} & \tikz{\node[model,label=below:$m_{15}^{\modtypeo}$](m15){$\emptyset$};}
								\end{tabular}
								}
							\\
							\&
							|[unit]|{
								\begin{tabular}{c c}
									\tikz{\node[model,label=above:$m_{16}^{\modtypei}$](m16){$\mathit{team}{2}(b),$ $\mathit{team}{1b}(c)$};} & \tikz{\node[model,label=above:$m_{17}^{\modtypei}$](m17){$\mathit{team}{2}(a),$ $\mathit{team}{2}(c)$};} \\
									\tikz{\node[model,xshift=-2cm,label=below:$m_{18}^{\modtypeo}$](m18){$\{\mathit{bonus}(b),$ $\mathit{bonus}(c)\}$};} & \tikz{\node[model,label=below:$m_{19}^{\modtypeo}$](m19){$\{ \mathit{bonus}(a),$ $\mathit{bonus}(c) \}$};}
								\end{tabular}
								}
							\&
							\\
						};
						\draw (m-1-2) node[below=16mm] {$u_1$};
						\draw (m-2-1) node[below=16mm] {$u_2$};
						\draw (m-2-2) node[below=15mm] {$u_3$};
						\draw (m-2-3) node[below=15mm] {$u_4$};
						\draw (m-3-2) node[below=19mm] {$u_5$};
						%
						{ 
						};
						\path[draw,line width=1pt,->] (m2) -- (m1);
						\path[draw,line width=1pt,->] (m3) -- (m1);
						\path[draw,line width=1pt,->] (m4) -- (m2);
						\path[draw,line width=1pt,->] (m5) -- (m3);
						\path[draw,line width=1pt,->] (m6) -- (m4);
						\path[draw,line width=1pt,->] (m7) -- (m5);
						\path[draw,line width=1pt,->] (m8) -- (m2);
						\path[draw,line width=1pt,->] (m9) -- (m3);
						\path[draw,line width=1pt,->] (m10) -- (m8);
						\path[draw,line width=1pt,->] (m11) -- (m9);
						\path[draw,line width=1pt,->] (m12) -- (m2);
						\path[draw,line width=1pt,->] (m13) -- (m3);
						\path[draw,line width=1pt,->] (m14) -- (m12);
						\path[draw,line width=1pt,->] (m15) -- (m13);
						\path[draw,line width=1pt,->] (m16) -- (m6);
						\path[draw,line width=1pt,->] (m16) -- (m14);
						\path[draw,line width=1pt,->] (m17) -- (m7);
						\path[draw,line width=1pt,->] (m17) -- (m15);
						\path[draw,line width=1pt,->] (m18) -- (m16);
						\path[draw,line width=1pt,->] (m19) -- (m17);
					\end{tikzpicture}
					\endpgfgraphicnamed
					
					\caption{Answer Set Graph of Example~\ref{ex:frameworkctd2}}
					\label{fig:ex:frameworkctd:modelgraph}
				\end{figure}
			\end{example}
			
			Note that there exist in general multiple valid evaluation graphs for a given program.
			However, it was shown that every strongly domain-expansion safe \hex-program
			has at least one evaluation graph, which is crucial for the applicability of the framework in the general setting.
			The construction of a concrete evaluation graph is the job of so-called \emph{evaluation heuristics}
			which can be plugged into the framework. The heuristics may have significant influence on efficiency
			and we will develop a new heuristics in Section~\ref{sec:domainexpansion:greedyheuristics}.

			An important concept for the model building process is that of \emph{first ancestor intersection units}.
			This will allow us in the following to decide whether the output models of multiple predecessor units
			origin from the same ancestor.
			
			\begin{definition}[First Ancestor Intersection Unit (FAI)]
				\label{def:fai}
				For an evaluation graph $\mathcal{E} = \langle V, E \rangle$ and distinct units $v, w \in V$,
				we say that $w$ \emph{is a first ancestor intersection unit (FAI) of} $v$ if there exist paths $p_1 \not= p_2$
				from $v$ to $w$ in $E$ such that $p_1$ and $p_2$ share no nodes apart from $v$ and $w$. We denote by $\mathit{fai}(v)$
				the set of all FAIs of a unit $v$.
			\end{definition}
			
			\begin{example}[ctd.]
				\label{ex:fai}
				In the evaluation graph from Example~\ref{ex:frameworkctd}, $u_1$ is the only FAI of $u_5$ because there
				exist two paths $u_5, u_2, u_1$ and $u_5, u_4, u_1$ from $u_5$ to $u_1$, but no other.
			\end{example}
			
			Evaluation is then based on an \emph{answer set graph}, which interrelates models at evaluation units in the
			evaluation graph.
			Each unit $u$ has assigned a set of \emph{input models} $\mathit{i\mbox{-}ints}(u)$ and
			a set of \emph{output models} $\mathit{o\mbox{-}ints}(u)$.

			\begin{definition}[Interpretation Structure]
				\label{def:interpretationstructure}
				An \emph{interpretation structure}
				for an evaluation graph $\mathcal{E} = \langle V, E \rangle$
				is a labeled directed acyclic graph
				$\mathcal{I} = \langle M, F, \mathit{unit}, \mathit{type}, \mathit{int} \rangle$
				where each node $M \subseteq \mathcal{I}_{\mathit{id}}$ is
				from a countable set $\mathcal{I}_{\mathit{id}}$ of identifiers, eg.~from $\mathbb{N}$,
				and $\mathit{unit}\colon\ M \rightarrow V$, $\mathit{type}\colon\ M \rightarrow \{ \modtypei, \modtypeo \}$
				and $\mathit{int}\colon\ M \rightarrow 2^{\mathit{HB}_{\Program}}$ are total node labeling functions.
			\end{definition}
			
			Given a unit $u \in V$ of an evaluation graph $\mathcal{E} = \langle V, E \rangle$,
			we denote for an interpretation structure $\mathcal{I}$ by
			$\mathit{i\mbox{-}ints}_{\mathcal{I}}(u) = \big\{ m \in M \mid \mathit{unit}(m) = u$ \text{ and } $\mathit{type}(m)=\modtypei \big\}$
			the \emph{input interpretations},
			and by 
			$\mathit{o\mbox{-}ints}_{\mathcal{I}}(u) = \big\{ m \in M \mid \mathit{unit}(m) = u$ \text{ and } $\mathit{type}(m)=\modtypeo \big\}$
			the \emph{output interpretations} at unit $u$, respectively.
			Given vertex $m \in M$, we further denote by
			$$\mathit{int}^{+}(m) = \mathit{int}(m) \cup \bigcup \big\{ \mathit{int}(m') \mid m' \in M \text{ and } m' \text{ is reachable from } m \text{ in } \mathcal{I} \big\}$$
			the \emph{expanded interpretation} of $m$.
			
			An interpretation structure is called \emph{interpretation graph}, if the edge relation satisfies
			some further properties.
			
			\begin{definition}[Interpretation Graph]
				\label{def:interpretationgraph}
				An \emph{interpretation graph} $\mathcal{I} = \{ M, F, \mathit{unit}, \mathit{type}, \mathit{int} \}$
				for an evaluation graph $\mathcal{E} = \langle V, E \rangle$
				is an interpretation structure which fulfills for every $u \in V$ the following properties:
				\begin{enumerate}
					\item[(IG-I)] \emph{I-connectedness:} for every $m \in \mathit{o\mbox{-}ints}_{\mathcal{I}}(u)$ the structure contains exactly one outgoing edge $(m, m') \in F$
									and $m' \in \mathit{i\mbox{-}ints}_{\mathcal{I}}(u)$ is an i-interpretation at unit $u$;
					\item[(IG-O)] \emph{O-connectedness:} for every $m \in \mathit{i\mbox{-}ints}_{\mathcal{I}}(u)$
									and for every predecessor unit $u_i \in \mathit{preds}_{\mathcal{E}}(u)$ of $u$,
									there is exactly one outgoing edge $(m, m_i) \in F$ and $m_i \in \mathit{o\mbox{-}ints}_{\mathcal{I}}(u_i)$
									(every $m_i$ is an o-interpretation at the respective unit $u_i$);
					\item[(IG-F)] \emph{FAI intersection}%
									\footnote{This property is called \emph{FAI intersection} because it implies that
									for any units $u$ and $v \in \mathit{fai}(u)$, all
									paths in the interpretation graph from some $m \in \mathit{i\mbox{-}ints}_{\mathcal{I}}(u)$
									to an output model $m' \in \mathit{o\mbox{-}ints}_{\mathcal{I}}(v)$
									share the same $m'$, i.e., the paths `intersect' at FAI units.
									In fact, in order to check the property for a unit $u$, it suffices to consider all $v \in \mathit{fai}(u)$
									because other ancestor units between $u$ and $v$ are not reachable via multiple paths,
									thus the property cannot be violated.}%
									\emph{:}
									for every $m \in \mathit{i\mbox{-}ints}_{\mathcal{I}}(u)$, let $\mathcal{I}'$ be the subgraph of $\mathcal{I}$ reachable
									from $m$, and let $\mathcal{E}'$ be the subgraph of $\mathcal{E}$ reachable from $u$. Then $\mathcal{I}'$ contains exactly one o-interpretation
									at each evaluation unit of $\mathcal{E}'$; and
					\item[(IG-U)] \emph{Uniqueness:} for each pair of distinct vertices $m_1, m_2 \in M, m_1 \not= m_2$ with $\mathit{unit}(m_1) = \mathit{unit}(m_2) = u$
								the expanded interpretation of $m_1$ and $m_2$ differs, formally $\mathit{int}^{+}(m_1) \not= \mathit{int}^{+}(m_2)$.
				\end{enumerate}
			\end{definition}

			\begin{example}[ctd.]
				\label{ex:frameworkctdmodelgraph}
				Figure~\ref{fig:ex:frameworkctd:modelgraph} shows an interpretation graph for the evaluation graph of Example~\ref{ex:frameworkctd}.
				Actually it shows an answer set graph, which is a special interpretation graph and is introduced next.
			\end{example}

			It is intended that an output model $m^{\modtypeo}$ of a unit $u$ results from the corresponding input model $m^{\modtypei}$,
			if Algorithm~\ref{alg:EvaluateExtendedPreGroundable} is called for the program $u$
			augmented by $m^{\modtypei}$ interpreted as facts, and corresponds to an answer set of $u^{\le}$.
			However, the definition of an interpretation graph does not refer to the \hex{}-programs
			in the evaluation units. Thus, it is not yet guaranteed that the output interpretations
			of the final evaluation unit are really the intended answer sets of the program.
			This requires the further notion of an \emph{answer set graph}.
			
			\begin{definition}[Answer Set Graph]
				\label{def:answersetgraph}
				Given an evaluation graph $\mathcal{E} = \langle V, E \rangle$,
				an \emph{answer set graph} is an interpretation graph $\mathcal{I}$ for $\mathcal{E}$ such
				that for each unit $u \in V$ it holds that
				\begin{enumerate}[(i)]
					\item every expanded input interpretation in $\mathit{i\mbox{-}ints}_{\mathcal{I}}(u)$
						is an answer set of $u^{<}$, i.e., $\mathit{int}^{+}(m) \in \mathcal{AS}(u^{<})$ for all $m \in \mathit{i\mbox{-}ints}_{\mathcal{I}}(u)$;
					\item every expanded output interpretation in $\mathit{i\mbox{-}ints}_{\mathcal{I}}(u)$
						is an answer set of $u^{\le}$, i.e., $\mathit{int}^{+}(m) \in \mathcal{AS}(u^{\le})$ for all $m \in \mathit{o\mbox{-}ints}_{\mathcal{I}}(u)$; and
					\item every input interpretation at $u$ is the union of the output interpretations it depends on, i.e.,
						$\mathit{int}(m) = \bigcup_{(m,m_i) \in F} \mathit{int}(m_i)$.
				\end{enumerate}
			\end{definition}
		
		\subsection{Using the Framework for Model Building}
		\label{sec:domainexpansion:framework:answersetgraph}
	
			For the description of the evaluation algorithm we need to introduce the concept of \emph{joins}.
			This will allow us to decide which combinations of output models of predecessor units serve
			as an input model to a successor unit.

			\begin{definition}[Join]
				\label{def:join}
				Let $\mathcal{I} = \langle M, F, \mathit{unit}, \mathit{type}, \mathit{int} \rangle$
				be an interpretation graph
				for an evaluation graph $\mathcal{E} = \langle V, E \rangle$.
				Let $u \in V$ be an evaluation unit and $u_1, \ldots, u_k$ be all units on which $u$
				depends.
				Let $m_i \in \mathit{o\mbox{-}ints}(u_i)$ for $1 \le i \le k$ be output models of predecessor
				units of $u$.
				
				Then the \emph{join} $m_1 \Join \cdots \Join m_k = \bigcup_{1 \le i \le k} m_i$
				is defined if for each $u' \in \mathit{fai}(u)$ there exists exactly
				one model $m' \in \mathit{o\mbox{-}ints}(u')$ reachable from some model $m_i$, $1 \le i \le k$,
				and undefined otherwise.
			\end{definition}

			Intuitively, the concept ensures that only those combinations of output models
			form an input model to an evaluation unit, which result from one common ancestor model
			in the model graph.

			During program evaluation,
			Algorithm~\ref{alg:BuildAnswerSets} starts from an empty answer set graph and expands it to the final answer set graph
			as follows.
			The algorithm iteratively selects an evaluation unit $u$ such that all direct predecessors $u_1, \ldots, u_k$
			have already been processed.
			Then the algorithm computes in the first step all input interpretations
			in Parts~\ref{alg:BuildAnswerSets:a} and~\ref{alg:BuildAnswerSets:b} (which is the empty interpretation for units without predecessors)
			and in a second step all
			output interpretations of $u$
			in Part~\ref{alg:BuildAnswerSets:d}. Both steps can be described in terms of updates to the answer set graph.
			The input interpretations of $u_{\mathit{final}}$ correspond then to the answer sets of $\Program$ (cf.~Part~\ref{alg:BuildAnswerSets:c}),
			which is formalized by the following theorem.
			
			\begin{theorem}
				\label{thm:BuildAnswerSets}
				Given an evaluation graph $\mathcal{E} = (V, E)$ of a program $\Program$,
				BuildAnswerSets returns $\mathcal{AS}(\Program)$.
			\end{theorem}
			
			This is demonstrated with a final example.
			
			\begin{algorithm}
				\SetAlgoRefName{BuildAnswerSets}
				\caption{}
				\label{alg:BuildAnswerSets}
				\label{errata:24}
				\DontPrintSemicolon

				\KwIn{evaluation graph $\mathcal{E} = (V, E)$ for a \hex{}-program $\Program$ with a unit $u_{\mathit{final}}$ that depends on all other units in $V$}
				\KwOut{all answer sets of $\Program$}
				\smallskip
				$M \leftarrow \emptyset, F \leftarrow \emptyset, \mathit{unit} \leftarrow \emptyset, \mathit{type} \leftarrow \emptyset, \mathit{int} \leftarrow \emptyset, U \leftarrow V$\;
				\While{$U \not= \emptyset$}{
					Choose $u \in U$ st.~$\mathit{preds}_{\mathcal{E}}(u) \cap U = \emptyset$\;
					Let $\{u_1, \ldots, u_k\} = \mathit{preds}_{\mathcal{E}}(u)$\;
					\nlset{(a)}{%
					\label{alg:BuildAnswerSets:a}%
					\uIf{$k = 0$}{
						$m \leftarrow \mathit{max}(M) + 1$\;
						$M \leftarrow M \cup \{m\}$\;
						$\mathit{unit}(m) \leftarrow u, \mathit{type}(m) \leftarrow \modtypei, \mathit{int}(m) \leftarrow \emptyset$\;
					}}
					\nlset{(b)}{%
					\label{alg:BuildAnswerSets:b}%
					\Else{
						\For{$m_1 \in \mathit{o\mbox{-}ints}(u_1), \ldots, m_k \in \mathit{o\mbox{-}ints}(u_k)$}{
							\If {$J = m_1 \Join \cdots \Join m_k$ is defined}{
								$m \leftarrow \mathit{max}(M) + 1$\;
								$M \leftarrow M \cup \{m\}$\;
								$F \leftarrow F \cup \big\{(m, m_i) \mid 1 \le i \le k\big\}$\;
								$\mathit{unit}(m) \leftarrow u, \mathit{type}(m) \leftarrow \modtypei, \mathit{int}(m) \leftarrow J$\;
							}
						}
					}
					}
					\nlset{(c)}{%
					\label{alg:BuildAnswerSets:c}%
					\If{$u = u_{\mathit{final}}$}{
						\Return $\mathit{i\mbox{-}ints}(u_{\mathit{final}})$\;
					}
					}
					\nlset{(d)}{%
					\label{alg:BuildAnswerSets:d}%
					\For{$m' \in \mathit{i\mbox{-}ints}(u)$}{
						$O \leftarrow \text{EvaluateExtendedPreGroundable}\big(u, \mathit{int}(m')\big)$\;
						\For{$o \in O$}{
							$m \leftarrow \mathit{max}(M) + 1$\;
							$M \leftarrow M \cup \{m\}$\;
							$F \leftarrow F \cup \big\{(m, m') \mid 1 \le i \le k\big\}$\;
							$\mathit{unit}(m) \leftarrow u, \mathit{type}(m) \leftarrow \modtypeo, \mathit{int}(m) \leftarrow o$\;
						}
					}
					}
					$U \leftarrow U \setminus \{u\}$\;
				}
			\end{algorithm}

			\begin{example}[ctd.]
				\label{ex:frameworkctd2}
				We now describe the construction of the answer set graph as depicted
				in Figure~\ref{fig:ex:frameworkctd:modelgraph} (without final unit $u_{\mathit{final}}$).
				The evaluation graph in Example~\ref{ex:frameworkctd} has a single unit $u_1$
				without predecessors. Its only input model is the empty one,
				i.e., $\mathit{i\mbox{-}ints(u_1)} = \{ m_1^{\modtypei} = \emptyset \}$.
				
				The algorithm chooses $u_1$ and computes the set of output models for input model $\emptyset$,
				which is $\mathit{o\mbox{-}ints(u_1)} = \big\{ m_2^{\modtypeo} = \{\mathit{team}1(a), \mathit{team}1(c)\}, m_3^{\modtypeo} = \{\mathit{team}1(b)\} \big\}$.
				
				In the next step, one of the components $u_2$, $u_3$ or $u_4$ can be chosen for evaluation
				because for each of them the single predecessor unit $u_1$ has already been processed.
				For $u_2$ and input model $m_4^{\modtypei} = m_2^{\modtypeo} = \{\mathit{team}1(a), \mathit{team}1(c)\}$,
				the unique output model is $m_6^{\modtypeo} = \{ \mathit{team}2(b) \}$,
				and for input model $m_5^{\modtypei} = m_3^{\modtypeo} = \{\mathit{team}1(b)\}$,
				the unique output model is $m_7^{\modtypeo} = \{ \mathit{team}2(a), \mathit{team}2(c) \}$.
				For $u_3$ and input model $m_8^{\modtypei} = m_2^{\modtypeo} = \{\mathit{team}1(a), \mathit{team}1(c)\}$,
				the unique output model is $m_{10}^{\modtypeo} = \{ \mathit{team}{1a}(a) \}$,
				and for input model $m_{9}^{\modtypei} = m_3^{\modtypeo} = \{\mathit{team}1(b)\}$,
				the unique output model is $m_{11}^{\modtypeo} = \{ \mathit{team}{1a}(b) \}$.
				For $u_4$ and input model $m_{12}^{\modtypei} = m_2^{\modtypeo} = \{\mathit{team}1(a), \mathit{team}1(c)\}$,
				the unique output model is $m_{14}^{\modtypeo} = \{ \mathit{team}{1b}(c) \}$,
				and for input model $m_{13}^{\modtypei} = m_3^{\modtypeo} = \{\mathit{team}1(b)\}$,
				the unique output model is $m_{15}^{\modtypeo} = \emptyset$.
				
				Then the algorithm chooses $u_5$ for evaluation.
				The first step is the computation of the input models of $u_5$.
				Because $u_5$ has two predecessor units $u_2$ and $u_4$ and each of them has
				two output models $m_6^{\modtypeo}$, $m_7^{\modtypeo}$ resp.~$m_{14}^{\modtypeo}$, $m_{15}^{\modtypeo}$,
				there are four possible combinations.
				However, only the joins $m_{16}^{\modtypei} = m_6^{\modtypeo} \Join m_{14}^{\modtypeo} = \{ \mathit{team}2(b), \mathit{team}{1b}(c) \}$
				and $m_{17}^{\modtypei} = m_7^{\modtypeo} \Join m_{15}^{\modtypeo} = \{ \mathit{team}2(a), \mathit{team}2(c) \}$
				are defined, because for the common ancestor unit $u_1$ of $u_5$,
				there is exactly one output model $m_2^{\modtypeo} \in \mathit{o\mbox{-}ints}(u_1)$ reachable from $m_6^{\modtypeo}, m_{14}^{\modtypeo}$
				resp.~$m_3^{\modtypeo} \in \mathit{o\mbox{-}ints}(u_1)$ from $m_7^{\modtypeo}, m_{15}^{\modtypeo}$.
				In contrast, from $m_6^{\modtypeo}, m_{15}^{\modtypeo}$ and $m_7^{\modtypeo}, m_{14}^{\modtypeo}$, both output models
				$m_2^{\modtypeo}, m_3^{\modtypeo}$ of $u_1$ are reachable.
				In the second step, the output models of $u_5$ are determined: for $m_{16}^{\modtypei}$
				the unique output model is $m_{18}^{\modtypeo} = \{ \mathit{bonus}(b), \mathit{bonus}(c) \}$
				and for $m_{17}^{\modtypei}$ the unique output model is
				$m_{19}^{\modtypeo} = \{ \mathit{bonus}(a), \mathit{bonus}(c) \}$.
				
				Finally, unit $u_{\mathit{final}}$ is chosen for evaluation. Actually, as this is the final unit and contains no rules,
				only the input models need to be determined. We have $5$ units with $2$ output models each, thus we have $2^5$ possible combinations.
				However,
				only $m_{20}^{\modtypei} = m_{2}^{\modtypeo} \Join m_{6}^{\modtypeo} \Join m_{10}^{\modtypeo} \Join m_{14}^{\modtypeo} \Join m_{18}^{\modtypeo} =
				\{ \mathit{team}_{1}(a), \mathit{team}_{1}(c), \mathit{team}_{1a}(a), \mathit{team}_{1b}(c), \mathit{team}_2(b), \mathit{bonus}(b), \mathit{bonus}(c) \}$
				(with the single reachable model $m_2^{\modtypeo}$ at the common ancestor unit $u_1$)
				and $m_{21}^{\modtypei} = m_{3}^{\modtypeo} \Join m_{7}^{\modtypeo} \Join m_{11}^{\modtypeo} \Join m_{15}^{\modtypeo} \Join m_{19}^{\modtypeo} =
				\{ \mathit{team}_{1}(b), \mathit{team}_{1a}(b), \mathit{team}_2(a), \mathit{team}_2(c), \mathit{bonus}(a), \mathit{bonus}(c) \}$
				(with the single reachable model $m_3^{\modtypeo}$ at the common ancestor unit $u_1$)
				are defined.
				These models are the answer sets of the program.
			\end{example}

		\begin{figure}[h]
			\centering
			\beginpgfgraphicnamed{committee}
			\begin{tikzpicture}[->,>=stealth',shorten >=1pt,auto,node distance=5cm,semithick]
				\node         (u1) [draw]                   {$u_1=\{r_1, r_2\}$};
				\node         (u2) [draw,right of=u1]       {$u_2=\{r_3, r_4\}$};
				\path (u2) edge [->] node {} (u1);

				\path (u1) edge [->,dashed,bend left] node {add answer set as input atoms} (u2);
				\node         (inc) [below=1cm of u2]       {detect inconsistency};
				\path (u2) edge [->,dashed] node {} (inc);
				\node         (ir) [draw,dashed,left of=inc]       {inconsistency reason $R$};
				\path (inc) edge [->,dashed] node {compute} (ir);
				
				\path (ir) edge [->,dashed] node {add as constraint $c_R$} (u1);
			\end{tikzpicture}
			\endpgfgraphicnamed
			\caption{Visualization of trans-unit learning}
			\label{fig:tulearning}
		\end{figure}

		we will demonstrate the two techniques with the next example.
		While monotonic and antimonotonic external atoms can be efficiently grounded without splitting,
		grounding nonmonotonic external atoms, which do not depend only on facts, requires exponentially many calls to external sources.
		Splitting can solve this grounding bottleneck, but it introduces a solving bottleneck
		because subsequent constraints cannot be propagated back.
}

		\subsection{Trans-Unit Propagation}
		\label{sec:hexprogramevaluation:transunitpropagation}

			To overcome the bottleneck we envisage at keeping the program splits for the sake of efficient grounding,
			but still allow for propagating conflicts back from later units to predecessor units.
			This can be seen as conflict-driven learning over multiple evaluation units,
			which is why we call out technique \emph{trans-unit (tu-)propagation}.
			The main idea is to associate an
			IR $R = (R^{+}, R^{-})$ of a later unit $u$
			with a constraint $c_R = \leftarrow R^{+}, \{ \naf a \mid a \in R^{-} \}$
			which we propagate to predecessors $u'$ of $u$.
			This is in order to eliminate exactly those interpretations already earlier,
			which would lead to inconsistency of $u$ anyway.

			Recall that for a domain $D$ of explanation atoms and an inconsistency reason $R = (R^{+}, R^{-})$ wrt.~a program $\Program$,
			all programs $\Program \cup \toFacts{F}$ with $F \subseteq D$ such that $R^{+} \subseteq F$ and $R^{-} \cap F = \emptyset$ are inconsistent.
			One can show that adding $c_R$ to $\Program$ does not eliminate answer sets:

			\addProposition{prop:constraintLearning}{
				For all \hex-programs $\Program$ and IRs $R = (R^{+}, R^{-})$ of $\Program$ wrt.~a domain $D$,
				we have that $\mathcal{AS}(\Program \cup \toFacts{F}) = \mathcal{AS}(\Program \cup \{ c_R \} \cup \toFacts{F})$ for all $F \subseteq D$.
			}

			\addProof{prop:constraintLearning}{
				We have that $\mathcal{AS}(\Program \cup \toFacts{F}) \supseteq \mathcal{AS}(\Program \cup \{ c_R \} \cup \toFacts{F})$
				since the addition of constraints can never generate additional answer sets but only eliminate them.
				Thus, it suffices to show that all answer sets of $\mathcal{AS}(\Program \cup \toFacts{F})$
				are also answer sets of $\mathcal{AS}(\Program \cup \{ c_R \} \cup \toFacts{F})$.
			
				To this end, note that the constraint $c_R$ is violated by $F \subseteq D$ iff $R^{+} \subseteq F$ and $R^{-} \cap F = \emptyset$.
				However, since $R$ is an IR, by Definition~\ref{def:inconsistencyReason} we have that $\Program \cup \toFacts{F}$ is inconsistent anyway
				for such an $F$, hence the addition of $c_R$ cannot eliminate answer sets of $\mathcal{AS}(\Program \cup \toFacts{F})$.
				%
			}

			Moreover and more importantly, such a constraint can also be added to predecessor units:
			for an inconsistency reason $R$ of a unit $u$ wrt.~a domain $D$,
			one can add the constraint $c_R$ to all predecessor units $u'$ of $u$ such that all atoms in $D$ are defined in $u'$ or its own transitive predecessors.
			Intuitively, this is the case due to acyclicity of the evaluation graph:
			since an already defined atom cannot be redefined in a later component, a constraint can be checked in the hierarchical evaluation
			of the evaluation units as soon as all relevant atoms from $D$ (which are the only ones that can appear in $c_R$) have been defined.
			This potentially eliminates wrong guesses earlier.

			\addProposition{prop:nogoodPushing}{
				For an evaluation unit $u$ and IR $R = (R^{+}, R^{-})$ of $u$ wrt.~a domain $D$,
				constraint $c_R$ can be added to all predecessor units $u'$
				s.t.~all atoms in $D$ are defined in $u'$ or one of its own transitive predecessors.
			}

			\addProof{prop:nogoodPushing}{
				If $c_R$ is violated under the current input $F$ of unit $u$,
				$u \cup \toFacts{F}$ is inconsistent by Proposition~\ref{prop:constraintLearning}.
				Due to acyclicity of the dependency graph, atoms defined in $u'$ or one of its own predecessors
				cannot be redefined in successor units of $u'$. Hence, the truth values of such atoms
				in $F$ are definitely known once $u'$ was evaluated.
				But then adding $c_R$ to $u$ to $u'$
				does not eliminate answer sets since $u \cup \toFacts{F}$ is inconsistent anyway.
			}

			\begin{example}[cont'd]
				\label{ex:committee3}
				Assume that $\mathit{joe}$ and $\mathit{sue}$ are the only technicians
				and $\mathit{alyson}$ is the only economist. Then,
				$u_2$ is always inconsistent if none of these three persons is selected,
				independent of other choices.
				If the current input to unit $u_2$ is $F = \{ \mathit{in}(\mathit{jack}), \mathit{in}(\mathit{joseph}) \}$, then
				by exploiting nogoods learned from the external source $\amp{\mathit{competences}}$,
				the IR $R = (\emptyset, \{ \mathit{in}(\mathit{joe}), \mathit{in}(\mathit{sue}), \mathit{in}(\mathit{alyson}) \})$
				of $u_2$ wrt.~$D = \{ \mathit{in}(\mathit{joe}), \mathit{in}(\mathit{sue}),$ $\mathit{in}(\mathit{alyson}) \}$ can be determined,
				and the constraint $c_R = \leftarrow \naf \mathit{in}(\mathit{joe}), \naf \mathit{in}(\mathit{sue}),$ $\naf \mathit{in}(\mathit{alyson})$
				can be added to $u_1$.
			\end{example}

			While in general there can be multiple predecessor units $u'$ satisfying the precondition of Proposition~\ref{prop:nogoodPushing},
			in our implementation and experiments we add $c_R$ to the top-most one.
			For analytical reasons, this is always better than (and subsumes) adding them
			to later units since as it eliminates invalid candidates as early as possible.
			Moreover, the proposition holds for any set $D$ of domain atoms; in practice when computing an inconsistency explanation for a unit $u$,
			we use the set of all atoms which appear already in some predecessor unit of $u$.

		\subsection{Implementation and Experiments}
		\label{sec:hexprogramevaluation:implementation}
		
			\definecolor{Gray}{gray}{0.9}
			\definecolor{LightGray}{gray}{0.95}
			\newcolumntype{g}{>{\columncolor{LightGray}}r}


			We now evaluate our approach using a benchmark suite.
			For the experiments, we integrated our techniques into the reasoner \dlvhex{}\footnote{\url{www.kr.tuwien.ac.at/research/systems/dlvhex}} with
			\gringo{} and \clasp{} from the Potassco suite\footnote{\url{https://potassco.org}} as backends.
			All benchmarks were run
			on a Linux machine with two 12-core AMD
			Opteron 6176 SE CPUs and 128 GB RAM;
			timeout was 300 secs and memout 16 GB per instance.
			We used the \emph{HTCondor} load distribution system\footnote{\url{http://research.cs.wisc.edu/htcondor}}
			to ensure robust runtimes
			(i.e., deviations of runs on the same instance are negligible).

			\leanparagraph{Setting and hypotheses}
			Based on the program class we want to assess,
			we selected our benchmarks such that guessing and checking
			is separated by nonmonotonic external atoms with value invention.
			For each benchmark we compare three configurations:
			(i) evaluation as a \textbf{monolithic} program,
			(ii) evaluation using \textbf{splitting},
			and (iii) evaluation using splitting and \textbf{trans-unit (tu-)propagation}.
			Our hypothesis is that
			(i) suffers a bottleneck from grounding the program as a whole, (ii) suffer a bottleneck from repetitive instantiation of later program units
			(where the overall effort is distributed over grounding and solving),
			whereas (iii) outperforms the other two approaches as it can restrict the grounding and propagate throughout the whole program at the same time.
			In the tables we show the overall runtime, grounding time, solving time, and for tu-propagation also the time needed for computing inconsistency reasons
			as by Algorithms~\ref{alg:hexcdnl} and~\ref{alg:incReason}.
			All time specifications are averaged over all instances of the respective size.
			Importantly, other computations besides grounding, solving and computing inconsistency reasons
			(such as preprocessing, running the model-building framework, etc)
			can cause the overall runtime to be higher than the sum of the other time specifications.
			Numbers in parentheses show the number of timeout instances.
			The instances we used for our benchmarks are available from \url{http://www.kr.tuwien.ac.at/research/projects/inthex/inconsistency}.

			\leanparagraph{Configuration Problem}
			Consider the following configuration problem.
			We assemble a server cluster consisting of various components. For a given component selection the cluster has different properties
			such as its performance, power consumption, disk space, etc. Properties may depend not only on individual components but also on their interplay,
			and this dependency can be nonmonotonic.
			For instance, the selection of an additional component might make it lose the property of low energy consumption.
			We want the cluster to have certain properties.

			In order to capture also similar configuration problems (such as Example~\ref{ex:committee}),
			we use a more abstract formalization as a quadruple $(D, P, m, C)$,
			where $D$ is a \emph{domain}, $P$ is a set of \emph{properties},
			$m$ is a function which associates each selection $S \subseteq D$ of a domain elements with a set of properties $m(S) \subseteq P$,
			and $C$ is a set of constraints of form $C_i = (C_i^{+}, C_i^{-})$, where $C_i^{+} \subseteq 2^P$ and $C_i^{-} \subseteq 2^P$
			define sets properties which must resp.~must not be simultaneously given;
			a selection $S \subseteq D$ is a \emph{solution} if for all $(C_i^{+}, C_i^{-}) \in C$
			we have $C_i^{+} \not\subseteq m(S)$ or $m(S) \cap C_i^{-} \not= \emptyset$.
	%
	%
			%

			For the experiment we consider $10$ randomly generated instances for each size $n$
			with $n$ domain elements, $\lfloor \frac{n}{5} + 1 \rfloor$ properties, a random function $m$, and a random number of constraints,
			such that their count has an expected value of $n$.
			The results are shown in Table~\ref{tab:abstractConf}.
			One can observe that for all shown instance sizes, the \textbf{splitting} approach is the slowest, \textbf{monolithic} evaluation is the second-slowest,
			and \textbf{tu-propagation} is the fastest configuration.
			The grounding and solving times show that for the monolithic approach, the main source of computation costs is grounding,
			which comes from exponentially many external calls.
			In contrast, for the splitting approach the runtime is more evenly distributed to the grounding and the solving phase,
			which comes from repetitive instantiation and evaluation of later program units.
			With \textbf{tu-propagation} both grounding and solving is faster.
			This is because learned nogoods effectively spare grounding and solving of later program units, which would be inconsistent anyway.
			Hence, the approach clearly outperforms both existing ones. The observations are in line with our hypotheses.

			\begin{table}[t]
				\small
				\centering
				\setlength\tabcolsep{1.5pt}
				\begin{tabular}[t]{|r|r|r|r|r|r|r|r|r|r|r|}
					\hline
					size & \multicolumn{3}{c|}{monolithic} & \multicolumn{3}{c|}{splitting} & \multicolumn{4}{c|}{tu-propagation} \\
					& total & ground & solve & total & ground & solve & total & ground & solve & analysis \\
					\hline
5      & 0.13 ~~(0) & 0.01     & 0.01     & 0.13 ~~(0) & 0.01     & 0.02     & 0.14 ~~(0) & 0.01     & 0.00     & 0.00     \\
7      & 0.18 ~~(0) & 0.03     & 0.04     & 0.27 ~~(0) & 0.07     & 0.08     & 0.19 ~~(0) & 0.03     & 0.01     & 0.01     \\
9      & 0.37 ~~(0) & 0.10     & 0.14     & 0.90 ~~(0) & 0.32     & 0.42     & 0.38 ~~(0) & 0.13     & 0.03     & 0.02     \\
11      & 0.72 ~~(0) & 0.45     & 0.14     & 4.51 ~~(0) & 1.68     & 2.42     & 0.39 ~~(0) & 0.14     & 0.07     & 0.06     \\
13      & 3.04 ~~(0) & 1.87     & 0.91     & 20.35 ~~(0) & 7.73     & 11.23     & 1.64 ~~(0) & 0.89     & 0.19     & 0.17     \\
15      & 12.64 ~~(0) & 8.23     & 3.74     & 102.82 ~~(0) & 42.16     & 55.32     & 5.53 ~~(0) & 3.64     & 0.51     & 0.46     \\
17      & 47.32 ~~(0) & 34.24     & 11.09     & 300.00 (10) & 122.58     & 164.20     & 15.66 ~~(0) & 10.64     & 1.07     & 0.90     \\
19      & 184.35 ~~(2) & 142.21     & 14.07     & 300.00 (10) & 122.37     & 163.66     & 61.99 ~~(1) & 34.13     & 4.83     & 4.34     \\
21      & 300.00 (10) & 300.00     & n/a     & 300.00 (10) & 127.01     & 160.96     & 148.85 ~~(3) & 91.11     & 12.45     & 11.72     \\
23      & 300.00 (10) & 300.00     & n/a     & 300.00 (10) & 128.13     & 161.24     & 300.00 (10) & 184.68     & 26.17     & 25.00     \\
					\hline
				\end{tabular}
				\caption{Configuration Problem}
				\label{tab:abstractConf}
			\end{table}

\nop{
			\begin{table}[t]
				\tiny
				\centering
				\setlength\tabcolsep{1.5pt}
				\begin{tabular}[t]{|r|r|r|r|r|r|r|r|r|r|r|r|r|r|r|}
					\hline
					size & \multicolumn{3}{c|}{monolithic} & \multicolumn{3}{c|}{splitting} & \multicolumn{4}{c|}{tu-propagation} & \multicolumn{4}{c|}{tu-propagation-pud} \\
					& total & ground & solve & total & ground & solve & total & ground & solve & analysis & total & ground & solve & analysis \\
					\hline
6 (10) & 0.14 ~~(0) & 0.01 (0) & 0.02 (0) & 0.17 ~~(0) & 0.03 (0) & 0.04 (0) & 0.15 ~~(0) & 0.01 (0) & 0.01 (0) & 0.00 (0) & 0.15 ~~(0) & 0.01 (0) & 0.01 (0) & 0.01 (0) \\
7 (10) & 0.18 ~~(0) & 0.02 (0) & 0.04 (0) & 0.26 ~~(0) & 0.07 (0) & 0.08 (0) & 0.19 ~~(0) & 0.03 (0) & 0.01 (0) & 0.01 (0) & 0.20 ~~(0) & 0.03 (0) & 0.01 (0) & 0.01 (0) \\
8 (10) & 0.20 ~~(0) & 0.05 (0) & 0.04 (0) & 0.46 ~~(0) & 0.15 (0) & 0.18 (0) & 0.19 ~~(0) & 0.03 (0) & 0.02 (0) & 0.02 (0) & 0.19 ~~(0) & 0.03 (0) & 0.02 (0) & 0.02 (0) \\
9 (10) & 0.37 ~~(0) & 0.10 (0) & 0.14 (0) & 0.89 ~~(0) & 0.31 (0) & 0.42 (0) & 0.38 ~~(0) & 0.12 (0) & 0.04 (0) & 0.02 (0) & 0.38 ~~(0) & 0.12 (0) & 0.04 (0) & 0.03 (0) \\
10 (10) & 0.45 ~~(0) & 0.22 (0) & 0.10 (0) & 2.24 ~~(0) & 0.83 (0) & 1.16 (0) & 0.35 ~~(0) & 0.11 (0) & 0.06 (0) & 0.05 (0) & 0.36 ~~(0) & 0.11 (0) & 0.06 (0) & 0.06 (0) \\
11 (10) & 0.73 ~~(0) & 0.46 (0) & 0.14 (0) & 4.50 ~~(0) & 1.66 (0) & 2.43 (0) & 0.40 ~~(0) & 0.14 (0) & 0.07 (0) & 0.07 (0) & 0.41 ~~(0) & 0.14 (0) & 0.08 (0) & 0.07 (0) \\
12 (10) & 1.56 ~~(0) & 0.93 (0) & 0.45 (0) & 9.77 ~~(0) & 3.64 (0) & 5.37 (0) & 0.91 ~~(0) & 0.45 (0) & 0.10 (0) & 0.09 (0) & 0.92 ~~(0) & 0.46 (0) & 0.12 (0) & 0.10 (0) \\
13 (10) & 3.05 ~~(0) & 1.88 (0) & 0.91 (0) & 20.59 ~~(0) & 7.69 (0) & 11.48 (0) & 1.65 ~~(0) & 0.90 (0) & 0.19 (0) & 0.17 (0) & 1.68 ~~(0) & 0.91 (0) & 0.22 (0) & 0.19 (0) \\
14 (10) & 4.95 ~~(0) & 3.78 (0) & 0.84 (0) & 44.92 ~~(0) & 16.75 (0) & 25.19 (0) & 1.52 ~~(0) & 0.83 (0) & 0.20 (0) & 0.18 (0) & 1.54 ~~(0) & 0.83 (0) & 0.22 (0) & 0.20 (0) \\
15 (10) & 12.79 ~~(0) & 8.42 (0) & 3.73 (0) & 103.02 ~~(0) & 41.15 (0) & 56.47 (0) & 5.47 ~~(0) & 3.61 (0) & 0.51 (0) & 0.45 (0) & 5.57 ~~(0) & 3.65 (0) & 0.57 (0) & 0.52 (0) \\
16 (10) & 19.93 ~~(0) & 16.89 (0) & 2.13 (0) & 219.54 ~~(0) & 86.38 (0) & 121.83 (0) & 3.86 ~~(0) & 2.40 (0) & 0.57 (0) & 0.55 (0) & 3.92 ~~(0) & 2.40 (0) & 0.65 (0) & 0.62 (0) \\
17 (10) & 66.91 ~~(1) & 34.21 (0) & 10.32 (0) & 300.00 (10) & 119.96 (0) & 166.69 (0) & 15.49 ~~(0) & 10.55 (0) & 1.07 (0) & 0.91 (0) & 15.61 ~~(0) & 10.62 (0) & 1.18 (0) & 1.01 (0) \\
18 (10) & 144.80 ~~(3) & 71.26 (0) & 11.24 (0) & 300.00 (10) & 122.60 (0) & 165.66 (0) & 45.06 ~~(1) & 18.62 (0) & 6.32 (0) & 6.13 (0) & 45.06 ~~(1) & 18.55 (0) & 6.46 (0) & 6.26 (0) \\
19 (10) & 300.00 (10) & 0.00 (0) & 0.00 (0) & 300.00 (10) & 120.18 (0) & 166.98 (0) & 69.14 ~~(2) & 22.07 (0) & 3.68 (0) & 3.39 (0) & 69.30 ~~(2) & 22.29 (0) & 3.92 (0) & 3.64 (0) \\
20 (10) & 300.00 (10) & 0.00 (0) & 0.00 (0) & 300.00 (10) & 127.47 (0) & 162.34 (0) & 38.00 ~~(0) & 27.05 (0) & 3.43 (0) & 3.23 (0) & 38.61 ~~(0) & 27.38 (0) & 3.87 (0) & 3.67 (0) \\
21 (10) & 300.00 (10) & 0.00 (0) & 0.00 (0) & 300.00 (10) & 123.77 (0) & 164.35 (0) & 151.23 ~~(4) & 58.83 (0) & 10.14 (0) & 9.71 (0) & 151.34 ~~(4) & 59.36 (0) & 10.90 (0) & 10.48 (0) \\
22 (10) & 300.00 (10) & 0.00 (0) & 0.00 (0) & 300.00 (10) & 122.37 (0) & 166.04 (0) & 225.30 ~~(6) & 87.59 (0) & 11.68 (0) & 11.07 (0) & 224.62 ~~(6) & 86.98 (0) & 12.37 (0) & 11.77 (0) \\
23 (10) & 300.00 (10) & 0.00 (0) & 0.00 (0) & 300.00 (10) & 124.05 (0) & 165.25 (0) & 300.00 (10) & 101.11 (0) & 15.51 (0) & 14.90 (0) & 300.00 (10) & 101.84 (0) & 16.81 (0) & 16.20 (0) \\
					\hline
				\end{tabular}
				\caption{Configuration Problem}
				\label{tab:abstractConf}
			\end{table}
}

			\leanparagraph{Diagnosis Problem}
			We now consider diagnosis problems formalized as follows.
			We have a quintuple $\langle \mathcal{O}_d, \mathcal{O}_p, \mathcal{H}, \mathcal{C}, P \rangle$,
			where
			sets $\mathcal{O}_d$ and $\mathcal{O}_p$ are definite resp.~potential observations,
			$\mathcal{H}$ is a set of hypotheses, $\mathcal{C}$ is a set of constraints over the hypotheses,
			and $P$ is a logic program which defines the observations which follow from given hypotheses.
			Each constraint $C \in \mathcal{C}$ forbids certain combinations of hypotheses.
			Let $\bar{h}$ be a new atom for each $h \in \mathcal{H}$ which does not appear in $P$.
			Then a solution consists of a set $S_{\mathcal{H}} \subseteq \mathcal{H}$
			of hypotheses and a set of potential observations $S_{\mathcal{O}_p} \subseteq {\mathcal{O}}_p$
			such that (i)
			all answer sets of $P \cup \{ h \vee \bar{h} \leftarrow \mid h \in \mathcal{H} \}$, which contain all of $\mathcal{O}_d \cup S_{\mathcal{O}_p}$, contain also $S_{\mathcal{H}}$,
			and (ii)
			$C \not\subseteq S_{\mathcal{H}}$ for all $C \in \mathcal{C}$.
			Informally, $S_{\mathcal{H}}$ are necessary hypotheses to explain the observations.

			As a concrete medical example, definite observations are known symptoms and test results,
			potential observations are possible outcomes of yet unfinished tests and hypotheses are possible causes (e.g.~diseases, nutrition behavior, etc).
			Constraints exclude certain (combinations of) hypotheses
			because it is known from anamnesis and the patient's declaration
			that they do not apply.
			A solution of the diagnosis problem corresponds to a set of possible observations which, if confirmed by tests,
			imply certain hypotheses (i.e., medical diagnosis), which can be exploited to perform the remaining tests goal-oriented.

			We use $10$ random instances for each instance size, where the size is given by the number of observations;
			observations are definite with a probability of $20\%$ and potential otherwise.
			The results are shown in Table~\ref{tab:diagnosis}.
			Unlike for the previous benchmark, \textbf{monolithic} is now slower than \textbf{splitting}.
			This is because the evaluation of the external source (corresponding to evaluating $P$) is much more expensive
			now.
			Since with \textbf{monolithic}, grounding always requires exponentially many evaluations, while during solving this is not necessarily the case,
			the evaluation costs have a larger impact to \textbf{monolithic} than to \textbf{splitting}.
			However, as before, \textbf{tu-propagation} is clearly the fastest configuration.
			Due to randomization of the instances, larger instances can in some cases be solved faster than smaller instances.
			Again, the observations are in line with our hypotheses.

			\begin{table}[t]
				\small
				\centering
				\setlength\tabcolsep{1.5pt}
				\begin{tabular}[t]{|r|r|r|r|r|r|r|r|r|r|r|}
					\hline
					size & \multicolumn{3}{c|}{monolithic} & \multicolumn{3}{c|}{splitting} & \multicolumn{4}{c|}{tu-propagation} \\
					& total & ground & solve & total & ground & solve & total & ground & solve & analysis \\
					\hline
5      & 1.31 ~~(0) & 0.38     & 0.81     & 0.25 ~~(0) & 0.12     & 0.02     & 0.94 ~~(0) & 0.77     & 0.01     & 0.01     \\
7      & 3.62 ~~(0) & 1.69     & 1.81     & 0.80 ~~(0) & 0.60     & 0.08     & 2.02 ~~(0) & 1.79     & 0.04     & 0.04     \\
9      & 12.02 ~~(0) & 7.07     & 4.76     & 2.35 ~~(0) & 1.94     & 0.27     & 4.23 ~~(0) & 3.94     & 0.06     & 0.05     \\
11      & 75.92 ~~(0) & 42.04     & 32.62     & 13.28 ~~(0) & 11.43     & 1.40     & 49.61 ~~(0) & 48.12     & 0.28     & 0.26     \\
13      & 300.00 (10) & 300.00     & n/a    & 51.29 ~~(1) & 46.51     & 4.26     & 23.19 ~~(0) & 22.54     & 0.12     & 0.11     \\
15      & 300.00 (10) & 300.00     & n/a     & 244.22 ~~(7) & 222.73     & 18.28     & 153.54 ~~(5) & 150.25     & 0.44     & 0.41     \\
17      & 300.00 (10) & 300.00     & n/a     & 260.10 ~~(7) & 237.44     & 19.88     & 158.91 ~~(5) & 156.30     & 0.77     & 0.75     \\
19      & 300.00 (10) & 300.00     & n/a     & 300.00 (10) & 274.57     & 22.56     & 179.53 ~~(5) & 175.96     & 0.56     & 0.54     \\
21      & 300.00 (10) & 300.00     & n/a     & 300.00 (10) & 272.97     & 24.05     & 86.86 ~~(2) & 84.76     & 0.32     & 0.30     \\
23      & 300.00 (10) & 300.00     & n/a    & 300.00 (10) & 271.61     & 25.52     & 164.58 ~~(4) & 159.98     & 0.39     & 0.37     \\
25      & 300.00 (10) & 300.00     & n/a     & 300.00 (10) & 270.88     & 26.31     & 241.03 ~~(7) & 235.46     & 0.57     & 0.55     \\
27      & 300.00 (10) & 300.00     & n/a     & 300.00 (10) & 243.35     & 26.96     & 275.57 ~~(8) & 270.67     & 0.70     & 0.68     \\
29      & 300.00 (10) & 300.00     & n/a     & 300.00 (10) & 269.84     & 27.43     & 203.08 ~~(5) & 200.02     & 0.53     & 0.51     \\
31      & 300.00 (10) & 300.00     & n/a     & 300.00 (10) & 269.83     & 27.59     & 268.79 ~~(8) & 264.27     & 0.62     & 0.59     \\
33      & 300.00 (10) & 300.00     & n/a     & 300.00 (10) & 269.94     & 27.61     & 300.00 (10) & 293.03     & 0.70     & 0.68     \\
					\hline
				\end{tabular}
				\caption{Diagnosis Problem}
				\label{tab:diagnosis}
			\end{table}

\nop{
			\begin{table}[t]
				\tiny
				\centering
				\setlength\tabcolsep{1.5pt}
				\begin{tabular}[t]{|r|r|r|r|r|r|r|r|r|r|r|r|r|r|r|}
					\hline
					size & \multicolumn{3}{c|}{monolithic} & \multicolumn{3}{c|}{splitting} & \multicolumn{4}{c|}{tu-propagation} & \multicolumn{4}{c|}{tu-propagation-pud} \\
					& total & ground & solve & total & ground & solve & total & ground & solve & analysis & total & ground & solve & analysis \\
					\hline
5 (10) & 0.45 ~~(0) & 0.15 (0) & 0.25 (0) & 0.15 ~~(0) & 0.05 (0) & 0.02 (0) & 0.36 ~~(0) & 0.23 (0) & 0.04 (0) & 0.00 (0) \\
7 (10) & 1.09 ~~(0) & 0.58 (0) & 0.57 (0) & 0.38 ~~(0) & 0.24 (0) & 0.07 (0) & 0.71 ~~(0) & 0.56 (0) & 0.10 (0) & 0.02 (0) \\
9 (10) & 3.37 ~~(0) & 2.29 (0) & 1.43 (0) & 1.01 ~~(0) & 0.76 (0) & 0.24 (0) & 1.36 ~~(0) & 1.16 (0) & 0.19 (0) & 0.03 (0) \\
11 (10) & 17.90 ~~(0) & 12.18 (0) & 7.17 (0) & 4.96 ~~(0) & 4.01 (0) & 1.14 (0) & 10.85 ~~(0) & 9.76 (0) & 1.40 (0) & 0.21 (0) \\
13 (10) & 52.94 ~~(0) & 47.49 (0) & 10.66 (0) & 15.06 ~~(0) & 12.27 (0) & 3.52 (0) & 6.12 ~~(0) & 5.63 (0) & 0.78 (0) & 0.09 (0) \\
15 (10) & 276.45 ~~(5) & 259.09 (0) & 39.49 (0) & 120.62 ~~(1) & 96.47 (0) & 22.82 (0) & 105.57 ~~(3) & 88.65 (0) & 13.17 (0) & 3.83 (0) \\
17 (10) & 300.00 (10) & 0.00 (0) & 37.06 (0) & 186.29 ~~(4) & 156.42 (0) & 35.09 (0) & 152.81 ~~(4) & 132.69 (0) & 20.16 (0) & 6.48 (0) \\
19 (10) & 300.00 (10) & 0.00 (0) & 36.74 (0) & 259.67 ~~(6) & 226.65 (0) & 48.83 (0) & 141.97 ~~(3) & 126.88 (0) & 15.67 (0) & 2.84 (0) \\
21 (10) & 300.00 (10) & 0.00 (0) & 36.18 (0) & 279.25 ~~(8) & 246.54 (0) & 49.78 (0) & 70.87 ~~(2) & 65.78 (0) & 7.53 (0) & 0.92 (0) \\
23 (10) & 300.00 (10) & 0.00 (0) & 36.20 (0) & 300.00 (10) & 266.40 (0) & 50.60 (0) & 136.88 ~~(4) & 130.75 (0) & 13.75 (0) & 1.23 (0) \\
25 (10) & 300.00 (10) & 0.00 (0) & 35.25 (0) & 300.00 (10) & 270.79 (0) & 45.49 (0) & 187.14 ~~(5) & 181.55 (0) & 17.64 (0) & 1.36 (0) \\
27 (10) & 300.00 (10) & 0.00 (0) & 34.97 (0) & 300.00 (10) & 271.66 (0) & 44.59 (0) & 207.00 ~~(5) & 202.36 (0) & 19.16 (0) & 1.36 (0) \\
29 (10) & 300.00 (10) & 0.00 (0) & 34.79 (0) & 300.00 (10) & 272.64 (0) & 43.45 (0) & 166.81 ~~(3) & 163.30 (0) & 14.62 (0) & 0.99 (0) \\
31 (10) & 300.00 (10) & 0.00 (0) & 33.77 (0) & 300.00 (10) & 274.05 (0) & 41.09 (0) & 242.31 ~~(7) & 238.47 (0) & 20.72 (0) & 1.26 (0) \\
33 (10) & 300.00 (10) & 0.00 (0) & 34.23 (0) & 300.00 (10) & 274.69 (0) & 40.25 (0) & 267.79 ~~(7) & 263.70 (0) & 21.84 (0) & 1.34 (0) \\
35 (10) & 300.00 (10) & 0.00 (0) & 32.94 (0) & 300.00 (10) & 277.35 (0) & 36.09 (0) & 299.52 ~~(9) & 294.39 (0) & 23.23 (0) & 1.29 (0) \\
37 (10) & 300.00 (10) & 0.00 (0) & 34.28 (0) & 300.00 (10) & 277.64 (0) & 36.08 (0) & 300.00 (10) & 295.22 (0) & 23.09 (0) & 1.22 (0) \\
39 (10) & 300.00 (10) & 0.00 (0) & 33.76 (0) & 300.00 (10) & 278.51 (0) & 34.61 (0) & 300.00 (10) & 294.88 (0) & 22.38 (0) & 1.08 (0) \\
					\hline
				\end{tabular}
				\caption{Diagnosis Problem}
				\label{tab:diagnosis}
			\end{table}
}

			\leanparagraph{Analysis of Best-Case Potential}
			Finally, in order to analyze the potential of our approach in the best case, we use a synthetic program.
			Our program of size $n$ is as follows:
			\begin{align*}
				P = \{	& \mathit{dom}(1..n). \ \mathit{in}(X) \vee \mathit{out}(X) \leftarrow \mathit{dom}(X). \ \mathit{someIn} \leftarrow \mathit{in}(X). \\
						& \mathit{r}(X) \leftarrow \ext{\mathit{diff}}{\mathit{dom}, \mathit{out}}{X}. \ \leftarrow \mathit{r}(X), \mathit{someIn} \}
			\end{align*}
			It uses a domain of size $n$ and guesses a subset thereof. It then uses an external atom to compute the complement set.
			The final constraint encodes that the guessed set and the complement must not be nonempty at the same time, i.e.,
			there are only two valid guesses: either all elements are in or all are out.
			
			The results are shown in Table~\ref{tab:synthetic}.
			While \textbf{splitting} separates the rules in the second line from the others and must handle each guess independently,
			the \textbf{monolithic} approach must evaluate the external atom under all possible extensions of $\mathit{out}$.
			Both approaches are exponential.
			In contrast, \textbf{tu-propagation} learns for each non-empty guess a constraint which excludes all guesses that set $\mathit{someIn}$ to true;
			after learning a linear number of such constraints, only the two valid guesses remain. 

			\begin{table}[t]
				\small
				\centering
				\setlength\tabcolsep{1.5pt}
				\begin{tabular}[t]{|r|r|r|r|r|r|r|r|r|r|r|}
					\hline
					size & \multicolumn{3}{c|}{monolithic} & \multicolumn{3}{c|}{splitting} & \multicolumn{4}{c|}{tu-propagation} \\
					& total & ground & solve & total & ground & solve & total & ground & solve & analysis \\
					\hline
5     & 3.06 (0) & 2.96     & $<0.005$     & 0.18 (0) & 0.04     & 0.04     & 0.14 (0) & 0.00     & 0.01     & 0.01     \\
6     & 14.06 (0) & 13.96     & $<0.005$     & 0.28 (0) & 0.09     & 0.08     & 0.15 (0) & 0.00     & 0.02     & 0.02     \\
7     & 65.45 (0) & 65.35     & $<0.005$     & 0.51 (0) & 0.20     & 0.19     & 0.15 (0) & 0.00     & 0.02     & 0.02     \\
8     & 289.53 (0) & 289.43     & $<0.005$     & 1.03 (0) & 0.44     & 0.44     & 0.18 (0) & 0.01     & 0.03     & 0.03     \\
9     & 300.00 (1) & 300.00     & n/a     & 2.13 (0) & 0.95     & 0.99     & 0.19 (0) & 0.01     & 0.04     & 0.03     \\
10     & 300.00 (1) & 300.00     & n/a     & 4.62 (0) & 2.16     & 2.10     & 0.20 (0) & 0.01     & 0.04     & 0.04     \\
11     & 300.00 (1) & 300.00     & n/a     & 9.87 (0) & 4.67     & 4.63     & 0.23 (0) & 0.01     & 0.05     & 0.05     \\
12     & 300.00 (1) & 300.00     & n/a     & 19.62 (0) & 9.19     & 9.54     & 0.24 (0) & 0.02     & 0.06     & 0.06     \\
13     & 300.00 (1) & 300.00     & n/a     & 41.36 (0) & 19.21     & 20.51     & 0.27 (0) & 0.02     & 0.07     & 0.07     \\
14     & 300.00 (1) & 300.00     & n/a     & 85.66 (0) & 40.88     & 41.67     & 0.29 (0) & 0.02     & 0.08     & 0.08     \\
15     & 300.00 (1) & 300.00     & n/a     & 178.97 (0) & 85.65     & 86.74     & 0.32 (0) & 0.03     & 0.10     & 0.09     \\
					\hline
				\end{tabular}
				\caption{Synthetic Set Guessing}
				\label{tab:synthetic}
			\end{table}

\nop{
			\begin{table}[t]
				\tiny
				\centering
				\setlength\tabcolsep{1.5pt}
				\begin{tabular}[t]{|r|r|r|r|r|r|r|r|r|r|r|r|r|r|r|}
					\hline
					size & \multicolumn{3}{c|}{monolithic} & \multicolumn{3}{c|}{splitting} & \multicolumn{4}{c|}{tu-propagation} & \multicolumn{4}{c|}{tu-propagation-pud} \\
					& total & ground & solve & total & ground & solve & total & ground & solve & analysis & total & ground & solve & analysis \\
					\hline
5 (1) & 3.06 (0) & 2.96 (0) & 0 (0) & 0.19 (0) & 0.04 (0) & 0.04 (0) & 0.13 (0) & 0.00 (0) & 0.01 (0) & 0.01 (0) & 0.14 (0) & 0.00 (0) & 0.02 (0) & 0.01 (0) \\
6 (1) & 13.88 (0) & 13.78 (0) & 0 (0) & 0.29 (0) & 0.09 (0) & 0.08 (0) & 0.15 (0) & 0.00 (0) & 0.02 (0) & 0.02 (0) & 0.15 (0) & 0.00 (0) & 0.02 (0) & 0.02 (0) \\
7 (1) & 64.15 (0) & 64.02 (0) & 0 (0) & 0.50 (0) & 0.19 (0) & 0.18 (0) & 0.16 (0) & 0.01 (0) & 0.02 (0) & 0.02 (0) & 0.16 (0) & 0.00 (0) & 0.03 (0) & 0.03 (0) \\
8 (1) & 286.68 (0) & 286.57 (0) & 0 (0) & 1.00 (0) & 0.43 (0) & 0.42 (0) & 0.18 (0) & 0.01 (0) & 0.03 (0) & 0.03 (0) & 0.19 (0) & 0.01 (0) & 0.04 (0) & 0.04 (0) \\
9 (1) & 300.00 (1) & 0.00 (0) & 0 (0) & 2.14 (0) & 0.95 (0) & 0.98 (0) & 0.19 (0) & 0.01 (0) & 0.04 (0) & 0.03 (0) & 0.20 (0) & 0.01 (0) & 0.05 (0) & 0.05 (0) \\
10 (1) & 300.00 (1) & 0.00 (0) & 0 (0) & 4.59 (0) & 2.08 (0) & 2.15 (0) & 0.20 (0) & 0.01 (0) & 0.04 (0) & 0.04 (0) & 0.22 (0) & 0.01 (0) & 0.06 (0) & 0.06 (0) \\
11 (1) & 300.00 (1) & 0.00 (0) & 0 (0) & 9.58 (0) & 4.36 (0) & 4.64 (0) & 0.23 (0) & 0.01 (0) & 0.05 (0) & 0.05 (0) & 0.24 (0) & 0.01 (0) & 0.07 (0) & 0.07 (0) \\
12 (1) & 300.00 (1) & 0.00 (0) & 0 (0) & 19.72 (0) & 9.08 (0) & 9.76 (0) & 0.25 (0) & 0.02 (0) & 0.06 (0) & 0.06 (0) & 0.27 (0) & 0.02 (0) & 0.08 (0) & 0.08 (0) \\
13 (1) & 300.00 (1) & 0.00 (0) & 0 (0) & 40.68 (0) & 18.95 (0) & 20.11 (0) & 0.27 (0) & 0.02 (0) & 0.07 (0) & 0.07 (0) & 0.30 (0) & 0.02 (0) & 0.10 (0) & 0.10 (0) \\
14 (1) & 300.00 (1) & 0.00 (0) & 0 (0) & 84.20 (0) & 39.74 (0) & 41.40 (0) & 0.29 (0) & 0.02 (0) & 0.08 (0) & 0.08 (0) & 0.32 (0) & 0.02 (0) & 0.11 (0) & 0.11 (0) \\
15 (1) & 300.00 (1) & 0.00 (0) & 0 (0) & 177.13 (0) & 83.87 (0) & 86.79 (0) & 0.33 (0) & 0.03 (0) & 0.10 (0) & 0.09 (0) & 0.36 (0) & 0.03 (0) & 0.13 (0) & 0.13 (0) \\
					\hline
				\end{tabular}
				\caption{Synthetic Set Guessing}
				\label{tab:synthetic}
			\end{table}
}

		\nop{
		\begin{example}[Synthetic Example]
		\ 

		\begin{verbatim}
% monolithic: grounding problem
% because external atoms are
% evaluated under exponentially
% many interpretations

% splitting: solving problem
% because constraint at the
% bottom cannot eliminate guesses

% Unit 1:
% the following rules let the
% heuristics split before the main part
% (since monotonicity of &id is not known)
p(a). p(b). p(c). p(d).
q(a). q(c). q(e).
pi(X) :- p(X).
qi(X) :- q(X).
y(X) v n(X) :- #int(X).
someN :- n(X).

% ----- split -----

% Unit 2:
po(X) :- &id[pi](X).
qo(X) :- &id[qi](X).
allY :- not someN.

% ----- split -----

% Unit 3:
% main check
res(X) :- &setMinus[po,qo](X).
:- res(X), not allY.
		\end{verbatim}
		\end{example}
		}


	\section{Discussion and Related Work}
	\label{sec:related}

		\leanparagraph{Related work on the saturation technique}
		Our encoding for inconsistency checking is related to a technique towards automated integration of guess and check programs~\cite{DBLP:journals/tplp/EiterP06},
		but using a different encoding.
		While the main idea of simulating the computation of the least fixpoint of the (positive) reduct by guessing a derivation sequence
		is similar, our encoding appears to be conceptually simpler thanks to the use of conditional literals.
		Moreover, they focus on integrating programs,
		but do not discuss inconsistency checking or query answering over subprograms.
		We go a step further and introduce a language extension towards query answering over general subprograms,
		which is more convenient for average users.
		Also, their approach can only handle ground programs.


		\leanparagraph{Related work on modular programming}
		Related to our approach towards query answering over subprograms are \emph{nested \hex-programs}, which allow for accessing answer sets of subprograms using dedicated \emph{external atoms}~\cite{ekr2013-inap11}.
		However, \hex{} is beyond plain ASP and requires a more sophisticated solver.
		Similar extensions of ordinary ASP exist~\cite{DBLP:conf/asp/TariBA05}, but unlike our approach, they did not come with a compilation approach into a single program.
		Instead, \emph{manifold programs} compile both the meta and the called program into a single one, similarly to our approach~\cite{DBLP:conf/birthday/FaberW11}.
		But this work depends on weak constraints, which are not supported by all systems. Moreover, the encoding requires a separate copy of the subprogram for each atom,
		while ours requires only a copy for each query.

		The idea of representing a subprogram by atoms in the meta-program is related to approaches for ASP debugging (cf.~\citeN{TUW-167810,DBLP:journals/tplp/OetschPT10}).
		But the actual computation is different: while debugging approaches explain why a particular interpretation is not an answer set
		(and print the explanation to the user), we aim at detecting the inconsistency and continuing reasoning afterwards.
		which can help users to get a better understanding of the structure of the problem	
		Also the \emph{stable-unstable semantics} supports an explicit interface to (possibly even nested) oracles~\cite{DBLP:journals/tplp/BogaertsJT16}.
		However, there are no query atoms but the relation between the guessing and checking programs is realized via an extension of the semantics.

		Modular programming approaches such as by~\citeN{tba2005} or~\citeN{DBLP:journals/corr/JanhunenOTW14}
		might appear to be related to our evaluation approach at first glance. However, a major difference is that these approaches provide program modules
		as language features (i.e., offer them to the user) and thus also define a semantics based on modules. In contrast, we use them just within the solver
		as an evaluation technique, while the semantics of \hex-programs does not use modules.
		
		Finally, we remark that on an abstract level, the component-wise evaluation might be considered to
		be related to lazy grounding (see e.g.~\citeN{DBLP:journals/fuin/PaluDPR09} and \citeN{DBLP:journals/tplp/LefevreBSG17}),
		as pointed out by colleagues in informal discussions. However, different from traditional (more narrow) definitions of lazy grounding,
		our approach does not interleave the instantiation of single rules with search, but rather ground whole program components using pregrounding algorithms.
		

		\leanparagraph{Related work on evaluation algorithms}
		Our Algorithm~\ref{alg:incReason} is related to \emph{conflict analysis} in existing CDNL-based algorithms, cf.~e.g.~\citeN{gks2012-aij}.
		While both are based on iterative resolution of a conflicting nogood with the implicant of one of its literals, the stop criterion is different.
		Traditional conflict analysis does the resolution based on \emph{learning schemas} such as \emph{first UIP}, which resolves as long as there are multiple literals assigned at the conflicting decision level.
		In contrast, we have to take the atoms from the domain into account, from which the facts can come.
		
		Also related are previous evaluation algorithms for \hex-programs.
		While the previous state-of-the-art algorithm corresponds to Algorithm~\ref{alg:hexcdnl}
		using $h_{\bot}(\Delta, \AssignmentP) = \bot$ for parameter $h$,
		alternative algorithms have been developed as well.
		One of them was presented by~\citeN{eiterFR014}, who used \emph{support sets}~\cite{DBLP:journals/corr/abs-1106-1819}
		to represent sufficient conditions to make an external atom true. Such support sets are used to speed up the compatibility check at \ref{alg:hexcdnl:d}.
		However, as this technique shows its benefits only during the solving phase, it does not resolve the grounding issue addressed in this paper.
		Later, another evaluation approach based on support sets was developed; in contrast to the previous one it compiles external atoms away altogether~\cite{r2017a-aaai-inlining}.
		However, the size of the resulting rewritten program strongly depends on the type of external sources and is exponential in general. Thus, the technique is only practical
		for certain external sources which are known to have a small representation by support sets (which is not the case for the benchmarks discussed in this paper).
		
		\leanparagraph{Related work on debugging approaches}
		Debugging approaches explain why a particular interpretation is not an answer set of a certain program instance
		and print the explanation to the user to help him/her to find the reason why an expected solution is not an answer set.
		In contrast, our approach of inconsistency reasons aims at detecting classes of instances which make a program inconsistent.
		This is motivated by applications which evaluate programs with fixed proper rules
		under different sets of facts. Then, inconsistency reasons can be exploited to skip evaluations which are known to yield no answer sets,
		as we did for the sake of improving the evaluation algorithm for \hex-programs~\cite{efikrs2016-tplp}.
		Here, in order to handle programs with expanding domains (value invention),
		the overall program is partitioned into multiple \emph{program components} which are arranged in an acyclic \emph{evaluation graph} that is evaluated top-down.
		In contrast to that, previous work on inconsistency analysis was mainly in the context of
		debugging of answer set programs and faulty systems in general,
		based on symbolic diagnosis~\cite{reiter1987}.
		For instance, the approach by~\citeN{Syrjaenen06debugginginconsistent} computes inconsistency explanations
		in terms of either minimal sets of constraints which need to be removed in order to regain consistency,
		or of odd loops (in the latter case the program is called \emph{incoherent}).
		The realization is based on another (meta-)ASP-program.
		Also the more general approaches by~\citeN{lpnmr07a} and~\citeN{TUW-167810} rewrite the program to debug into a meta-program using dedicated control atoms.
		The goal is to support the human user to find reasons for undesired behavior of the program.
		Possible queries are, for instance, why a certain interpretation is not an answer set or why a certain atom is not true in an answer set.
		Explanations are then in terms of unsatisfied rules, only cyclically supported atoms, or unsupported atoms.
		
		\leanparagraph{Related work on inconsistency characterizations}
		%
		%
		%
		Inconsistency management has also been studied for more specialized formalisms
		such as for multi-context systems (MCSs)~\cite{DBLP:journals/ai/EiterFSW14} and DL-programs~\cite{Eiter2013}.
		
		MCSs are sets of knowledge-bases (called \emph{contexts}), which are implemented in possibly different formalisms that are abstractly identified by their belief sets
		and interconnected by so-called \emph{bridge rules}. Bridge rules look syntactically similar to ASP rules, but can access different contexts and derive atoms in a context based on information from other contexts.
		Such systems may become inconsistent even if the individual contexts are all consistent.
		Inconsistency diagnoses for MCSs use pairs of bridge rules which must be removed resp.~added unconditionally (i.e., their body is empty) to make the system consistent.

		Inconsistent management for DL-programs is about modifying the Abox of the ontology to restore consistency of the program.
		While the approaches are related to ours on an abstract level,
		their technical formalizations are not directly comparable to ours as they refer to specific elements of the respective formalism (such as bridge rules of MCSs and the Abox of ontologies),
		which do not exist in general logic programs.


		Most closely related to our work is a decision criterion on the inconsistency of programs~\cite{r2017-aaai-equivalence},
		which characterizes inconsistency wrt.~models and unfounded sets of the program at hand.
		In contrast, the notion of IRs from this work characterizes it in terms of the input atoms.
		However, there is a relation which we elaborate in more detail in the following.
		To this end we use the following definition and result by~\citeN{faber2005-lpnmr}:

		\begin{definition}[Unfounded Set]
			\label{def:unfoundedset}
			Given a \hex-program $P$ and an assignment $I$, let $U$
			be any set of atoms appearing in $P$. Then, 
			$U$ is an \emph{unfounded set for $P$ wrt.~$I$} if, for each rule $r \in P$ with $H(r) \cap U \not= \emptyset$,
			at least one of the following conditions holds:
			\begin{enumerate}[(i)]
				\item\label{def:unfoundedset:i} some literal of $B(r)$ is false wrt.~$I$; or
				\item\label{def:unfoundedset:ii} some literal of $B(r)$ is false wrt.~$I \setminus U$; or
				\item\label{def:unfoundedset:iii} some atom of $H(r) \setminus U$ is true wrt.~$I$.
			\end{enumerate}
		\end{definition}

		Unfounded sets have been used to characterize answer sets as follows.
		A (classical) model $M$ of a \hex-program $P$ is called \emph{unfounded-free} if it does not intersect with an unfounded set of $P$ wrt.~$M$.
		One can then show~\cite{faber2005-lpnmr}:
		
		\addProposition{prop:answerSetCharacterization}{
			A model $M$ of a \hex-program $P$ is an answer set of $P$ iff it is unfounded-free.
		}

		\addProof{prop:answerSetCharacterization}{
			See~\citeN{faber2005-lpnmr}.
		}

		We make use of the idea to restrict the sets of head and body atoms which may occur in a program component as introduced by \citeN{DBLP:journals/tplp/Woltran08}.
		For sets of atoms $\mathcal{H}$ and $\mathcal{B}$,
		let $\mathcal{P}^e_{\langle \mathcal{H}, \mathcal{B} \rangle}$ denote the set of all \hex-programs whose head atoms come only from $\mathcal{H}$
		and whose body atoms and external atom input atoms come only from $\mathcal{B}$.
		We then use the following result by~\citeN{r2017-aaai-equivalence}:

		\addProposition{prop:ufsInconsistency}{
			Let $P$ be a \hex-program. Then $P \cup R$ is inconsistent for all $R \in \mathcal{P}^e_{\langle \mathcal{H}, \mathcal{B} \rangle}$
			iff for each classical model $I$ of $P$ there is a nonempty unfounded set $U$ of $P$ wrt.~$I$ such that $U \cap I \not= \emptyset$ and $U \cap \mathcal{H} = \emptyset$.
		}

		\addProof{prop:ufsInconsistency}{
			See~\citeN{r2017-aaai-equivalence}.
		}

		Intuitively, the result states that a program is inconsistent, iff each (classical) model is dismissed as
		answer set because it is not unfounded-free.
		Our concept of IRs is related to this result as follows.
		A pair $(R^{+}, R^{-})$ of disjoint sets of atoms $R^{+} \subseteq D$ and $R^{-} \subseteq D$ is an IR of program $P$ wrt.~$D$
		iff $P$ is inconsistent for all additions of facts that contain all atoms from $R^{+}$ but none from $R^{-}$;
		this can be expressed by applying the previous result for $\mathcal{B} = \emptyset$ and an appropriate selection of $\mathcal{H}$.

		\addLemma{lem:irCorrespondence}{
			Let $P$ be a \hex-program and $D$ be a domain of atoms.
			A pair $(R^{+}, R^{-})$ of sets of atoms $R^{+} \subseteq D$ and $R^{-} \subseteq D$ with $R^{+} \cap R^{-} = \emptyset$
			is an IR of a \hex-program $P$ wrt.~$D$
			iff
			$P \cup \toFacts{R^{+}} \cup R$ is inconsistent for all $R \in \mathcal{P}^e_{\langle D \setminus R^{-}, \emptyset \rangle}$.
		}
		
		\addProof{lem:irCorrespondence}{
			The claim follows basically from the observation that the sets of programs allowed to be added on both sides are the same.
			In more detail:

			($\Rightarrow$)
				Let $(R^{+}, R^{-})$ be an IR of $P$ with $R^{+} \subseteq D$ and $R^{-} \subseteq D$ with $R^{+} \cap R^{-} = \emptyset$.
				Let $R \in \mathcal{P}^e_{\langle D \setminus R^{-}, \emptyset \rangle}$. We have to show that $P \cup \toFacts{R^{+}} \cup R$ is inconsistent.
				Take $F = R^{+} \cup \{ a \mid a \leftarrow \in R \}$; then $R^{+} \subseteq F$ and $R^{-} \cap F = \emptyset$ and thus by our precondition that $(R^{+}, R^{-})$ is an IR
				we have that $P \cup \toFacts{F}$, which is equivalent to $P \cup \toFacts{R^{+}} \cup R$, is inconsistent.
				
			($\Leftarrow$)
				Suppose $P \cup \toFacts{R^{+}} \cup R$ is inconsistent for all $R \in \mathcal{P}^e_{\langle D \setminus R^{-}, \emptyset \rangle}$.
				Let $F \subseteq D$ such that $R^{+} \subseteq F$ and $R^{-} \cap F = \emptyset$. We have to show that $P \cup \toFacts{F}$ is inconsistent.
				Take $R = \{ a \leftarrow \mid a \in F \setminus R^{+} \}$ and observe that $R \in \mathcal{P}^e_{\langle D \setminus R^{-}, \emptyset \rangle}$.
				By our precondition we have that $P \cup \toFacts{R^{+}} \cup R$ is inconsistent. The observation that the latter is equivalent to $P \cup \toFacts{F}$ proves the claim.
		}
		
		Now Proposition~\ref{prop:ufsInconsistency} and Lemma~\ref{lem:irCorrespondence}
		in combination allow for establishing a relation of IRs in terms input atoms on the one hand, and a characterization in terms of of models and unfounded sets on the other hand.

		\addProposition{prop:irCharacterization}{
			Let $P$ be a ground \hex-program and $D$ be a domain.
			Then a pair of sets of atoms $(R^{+}, R^{-})$ with $R^{+} \subseteq D$, $R^{-} \subseteq D$ and $R^{+} \cap R^{-} = \emptyset$ is an IR of $P$
			iff for all classical models $M$ of $P$ either
			(i) $R^{+} \not\subseteq M$ or
			(ii) there is a nonempty unfounded set $U$ of $P$ wrt.~$M$ such that $U \cap M \not= \emptyset$ and $U \cap (D \setminus R^{-}) = \emptyset$.
		}
		
		\addProof{prop:irCharacterization}{
			($\Rightarrow$)
				Let $(R^{+}, R^{-})$ be an IR of $P$ wrt.~$D$.
				Consider a classical model $M$ of $P$. We show that one of (i) or (ii) is satisfied.

				If $M$ is not a classical model of $P \cup \toFacts{R^{+}}$, then, since $M$ is a model of $P$, we have that $R^{+} \not\subseteq M$ and thus Condition~(i) is satisfied.
				
				In case that $M$ is a classical model of $P \cup \toFacts{R^{+}}$, first observe that,
				since $(R^{+}, R^{-})$ is an IR,
				by Lemma~\ref{lem:irCorrespondence} $P \cup \toFacts{R^{+}} \cup R$ is inconsistent for all $R \in \mathcal{P}^e_{\langle D \setminus R^{-}, \emptyset \rangle}$.
				By Proposition~\ref{prop:ufsInconsistency}, this is further the case iff for each classical model $M'$ of $P \cup \toFacts{R^{+}}$
				there is a nonempty unfounded set $U$ such that $U \cap M' \not= \emptyset$ and $U \cap (D \setminus R^{-}) = \emptyset$.
				Since $M$ is a model of $P \cup \toFacts{R^{+}}$ it follows that an unfounded set as required by Condition~(ii) exists.
			
			($\Leftarrow$)
				Let $(R^{+}, R^{-})$ be a pair of sets of atoms such that for all classical models $M$ of $P$ either (i) or (ii) holds.
				We have to show that it is an IR of $P$, i.e., $P \cup \toFacts{F}$ is inconsistent for all $F \subseteq D$ with $R^{+} \subseteq F$ and $R^{-} \cap F = \emptyset$.
				
				Assignments which are no classical models of $P \cup \toFacts{F}$ cannot be answer sets, thus it suffices to show for all classical models of $P \cup \toFacts{F}$ that they are no answer sets.
				Consider an arbitrary but fixed $F \subseteq D$ with $R^{+} \subseteq F$ and $R^{-} \cap F = \emptyset$
				and let $M$ be an arbitrary classical model of $P \cup \toFacts{F}$.
				Then $M$ is also a classical model of $P$ and thus, by our precondition, either (i) or (ii) holds.
				
				If (i) holds, then there is an $a \in R^{+}$ such that $a \not\in M$. But since $R^{+} \subseteq F$ we have $a \leftarrow \in \toFacts{F}$ and thus $M$ cannot be a classical model
				and therefore no answer set of $P \cup \toFacts{F}$.
				If (ii) holds, then there is a nonempty unfounded set $U$ of $P$ wrt.~$M$ such that $U \cap M \not= \emptyset$ and $U \cap (D \setminus R^{-}) = \emptyset$, i.e.,
				$U$ does not contain elements from $D \setminus R^{-}$.
				Then $U$ is also an unfounded set of $P \cup \toFacts{R^{+}}$ wrt.~$M$.
				Then by Proposition~\ref{prop:ufsInconsistency} we have that $P \cup \toFacts{R^{+}} \cup R$
				is inconsistent for all $R \in \mathcal{P}^e_{\langle D \setminus R^{-}, \emptyset \rangle}$.
				The latter is, by Lemma~\ref{lem:irCorrespondence}, the case iff $(R^{+}, R^{-})$ be an IR of $P$ wrt.~$D$.
		}

		Intuitively, each classical model $M$ must be excluded from being an answer set of $P \cup \toFacts{F}$ for any $F \subseteq D$
		with $R^{+} \subseteq F$ and $R^{-} \cap F = \emptyset$.
		This can either be done by ensuring that $M$ does not satisfy $R^{+}$ or that some atom $a \in M$ is unfounded if
		$F$ is added because $a \not\in D \setminus R^{-}$.

		While the previous result characterizes inconsistency of a program precisely, it is computationally expensive to apply
		because one does not only need to check a condition for all models of the program at hand, but also for all unfounded sets of all models.
		We therefore present a second, simpler condition, which refers only to the program's models but not the the unfounded sets wrt.~these models.

		\addProposition{prop:irCharacterization2}{
			Let $P$ be a ground \hex-program and $D$ be a domain such that $H(P) \cap D = \emptyset$.
			For a pair of sets of atoms $(R^{+}, R^{-})$ with $R^{+} \subseteq D$ and $R^{-} \subseteq D$,
			if for all classical models $M$ of $P$ we either have (i) $R^{+} \not\subseteq M$ or (ii) $R^{-} \cap M \not= \emptyset$,
			then $(R^{+}, R^{-})$ is an IR of $P$.
		}
		
		\addProof{prop:irCharacterization2}{
			Let $(R^{+}, R^{-})$ be a pair of sets of atoms such that for all classical models $M$ of $P$ we either have $R^{+} \not\subseteq M$ or $R^{-} \cap M \not= \emptyset$.
			We have to show that it is an IR of $P$, i.e., $P \cup \toFacts{F}$ is inconsistent for all $F \subseteq D$ with $R^{+} \subseteq F$ and $R^{-} \cap F = \emptyset$.
			
			Assignments which are no classical models of $P \cup \toFacts{F}$ cannot be answer sets, thus it suffices to show for all classical models that they are no answer sets.
			Let $M$ be an arbitrary classical model of $P \cup \toFacts{F}$.
			Then $M$ is also a classical model of $P$ and thus, by our precondition, either (i) or (ii) holds.
			
			If (i) holds, then there is an $a \in R^{+}$ such that $a \not\in M$. But since $R^{+} \subseteq F$ we have $a \leftarrow \in \toFacts{F}$ and thus $M$ cannot be a classical model
			and therefore no answer set of $P \cup \toFacts{F}$.
			If (ii) holds, then there is an $a \in R^{-} \cap M$.
			Since $R^{-} \cap F = \emptyset$ we have that $a \leftarrow \not\in \toFacts{F}$. Moreover, we have that $a \not\in H(P)$ by our precondition that $H(P) \cap D = \emptyset$.
			But then $\{ a \}$ is an unfounded set of $P \cup \toFacts{F}$ wrt.~$M$ such that $\{ a \} \cap M \not= \emptyset$ and thus, by Proposition~\ref{prop:answerSetCharacterization},
			$M$ is not an answer set of $P$.
		}
		
		However, in contrast to Proposition~\ref{prop:irCharacterization},
		note that Proposition~\ref{prop:irCharacterization2} does \emph{not} hold in inverse direction.
		That is, the proposition does not characterize inconsistency exactly but provides a practical means for identifying inconsistency in some cases.
		This is demonstrated by the following example.
		
		\begin{example}
			Let $D = \{ x \}$ and $P = \{ \leftarrow \naf a \}$. Then $R = (\emptyset, \emptyset)$ is an IR of $P$ because $P$ is inconsistent and no addition of any atoms as facts is allowed.
			However, the classical model $M = \{ a \}$ contains $R^{+}$ but does not intersect with $R^{-}$, hence the precondition of Proposition~\ref{prop:irCharacterization2} is not satisfied.
		\end{example}

	\section{Conclusion and Outlook}
	\label{sec:conclusion}

		\leanparagraph{Summary}
		In this work have have studied inconsistency of \hex-programs in two steps.
		First, we focused on fixed programs and decide inconsistency of a subprogram within a meta-program.
		This was in view of restrictions of the saturation modeling technique,
		which allows for exploiting disjunctions for solving \conp-hard problems
		that involve checking a property all objects in a given domain.
		The use of default-negation in saturation encodings turns out to be problematic and a rewriting is not always straightforward.
		On the other hand, complexity results imply that any \conp-hard problem can be reduced to brave reasoning over disjunctive ASP.
		Based on our encoding for consistency checking for normal programs, we realized query answering over subprograms,
		which paves the way for intuitive encodings of checks which involve default-negation.

		Second, we identified classes of inconsistent program instances wrt.~their input.
		To this end, we have introduced the novel notion of \emph{inconsistency reasons} for explaining why an ASP- or \hex-program is inconsistent in terms of input facts.
		In contrast to previous notions from the area of answer set debugging, which usually explain why a certain interpretation is not an answer set
		of a concrete program, we consider programs with fixed proper rules, which are instantiated with various data parts.
		Moreover, while debugging approaches aim at explanations which support human users, ours was developed with respect to
		optimizations in evaluation algorithms for \hex-programs. More precisely, \hex-programs are typically split into multiple components for the sake of efficient grounding.
		This, however, harms conflict-driven learning techniques as it introduces barriers for propagation.
		Inconsistency reasons are used to realize conflict-driven learning techniques
		for such programs and propagate learned nogoods over multiple components.
		Our experiments show that the technique is promising and leads to an up to exponential speedup.
		
		\leanparagraph{Outlook}
		For the query answering part, future work includes the application of the extension to non-ground queries.
		Currently, a separate copy of the subprogram is created for every query atom. However, it might be possible in some cases, to answer multiple queries simultaneously.
		Another possible starting point for future work is the application of our encoding for a more efficient implementation of nested \hex-programs.
		Currently, nested \hex-programs are evaluated by separate instances of the reasoner for the calling and the called program. While this approach is strictly more expressive (and thus the evaluation also more expensive)
		due to the possibility to nest programs up to an arbitrary depth, it is possible in some cases to apply the technique from the paper as an evaluation technique (e.g.~if the called program is normal and does not contain further nested calls).
		Another issue is that if the subprogram is satisfiable, then the meta-program has \emph{multiple} answer sets, each of which representing an answer set of the subprogram.
		If only consistency resp.~inconsistency of the subprogram is relevant for the further reasoning in the meta-program, this leads to the repetition of solutions.
		In an implementation, this problem can be tackled using projected solution enumeration~\cite{DBLP:conf/cpaior/GebserKS09}.
		
		For the new learning technique, an interesting starting point for future work is the generalization of nogood propagation to general,
		not necessarily inconsistent program components.
		In this work, typical solver optimizations have been disabled for tu-propagation as they can harm soundness of the algorithm in general;
		although the other approaches used them and tu-propagation was still the fastest,
		an important extension is the identification of specific solver optimizations which are
		compatible with inconsistency analysis and might lead to an even greater improvement.
		Moreover, exploiting the structure of the program might allow for finding inconsistency reasons in more cases.

	\ifinlineref

	\else
		\bibliographystyle{acmtrans}
		\bibliography{inconsistency}
	\fi

	\newpage
\begin{appendix}

\section{Proofs}
\label{sec:proofs}

\proofs{}

\end{appendix}


\end{document}